\documentclass[10pt,twocolumn,letterpaper]{article}
\PassOptionsToPackage{hyphens}{url}
\usepackage{latexml}

\newif\ifGeoGrainReview
\GeoGrainReviewfalse
\iflatexml
\usepackage{times,xspace}
\usepackage[dvipsnames]{xcolor}
\usepackage[numbers,sort&compress]{natbib}
\usepackage[format=plain,labelformat=simple,labelsep=period,font=small,compatibility=false]{caption}
\usepackage[font=footnotesize,skip=3pt,subrefformat=parens]{subcaption}
\usepackage[shortlabels,inline]{enumitem}
\setlist[itemize]{noitemsep,leftmargin=*,topsep=0em}
\setlist[enumerate]{noitemsep,leftmargin=*,topsep=0em}
\else
\usepackage[review]{cvpr}
\fi
\iflatexml\else
\ifGeoGrainReview
\else
  \toggletrue{cvprfinal}
  
  \ClearShipoutPicture
\fi
\fi

\usepackage{url}
\usepackage{graphicx}

\usepackage{booktabs}       
\usepackage{amsfonts}       
\usepackage{nicefrac}       
\usepackage{xcolor}         

\usepackage{amsmath}
\usepackage{amssymb}
\usepackage{graphicx}
\usepackage{multirow}
\usepackage{bm}
\usepackage{array}
\usepackage{makecell}
\iflatexml\else\usepackage{adjustbox}\fi
\usepackage[table]{xcolor}
\iflatexml\usepackage{colortbl}\fi
\usepackage{pifont}
\usepackage{algorithm}
\usepackage{algpseudocode}
\usepackage{wrapfig}
\iflatexml\else\usepackage{cuted}\fi
\iflatexml\else\usepackage{capt-of}\fi

\newcommand*{\email}[1]{\texttt{#1}}
\usepackage{tabularx}
\newcolumntype{Y}{>{\raggedright\arraybackslash}X}
\iflatexml
\newcolumntype{Z}{c}
\else
\newcolumntype{Z}{>{\centering\arraybackslash}X}
\fi
\newcommand{\GeoGrainDeferFloat}[2]{%
  \iflatexml
    \expandafter\gdef\csname GeoGrainFloat#1\endcsname{#2}%
  \else
    #2%
  \fi
}
\newcommand{\GeoGrainShowFloat}[1]{%
  \iflatexml\csname GeoGrainFloat#1\endcsname\fi
}

\newcommand{\GeoGrainTableFont}[1]{%
  \iflatexml\small\sffamily\else#1\fi
}

\newcommand{\GeoGrainUnderline}[1]{%
  \iflatexml
    \ifmmode\underline{#1}%
    \else\lxWithClass{ltx_framed_underline}{\underline{#1}}\fi
  \else\underline{#1}\fi%
}

\iflatexml\else
\makeatletter
\let\GG@addtocurcol@kernel\@addtocurcol
\usepackage{flafter}
\let\GG@addtocurcol@flafter\@addtocurcol
\let\@addtocurcol\GG@addtocurcol@kernel
\makeatother
\fi

\definecolor{cvprblue}{rgb}{0.21,0.49,0.74}
\usepackage[pagebackref,breaklinks,colorlinks,allcolors=cvprblue]{hyperref}
\iflatexml
\usepackage[capitalize]{cleveref}
\crefname{section}{Sec.}{Secs.}
\Crefname{section}{Section}{Sections}
\Crefname{table}{Table}{Tables}
\crefname{table}{Tab.}{Tabs.}
\fi

\iflatexml\else
\makeatletter
\ExplSyntaxOn
\tl_new:N \l_gg_bib_file_tl
\tl_new:N \l_gg_bib_header_tl
\seq_new:N \l_gg_bib_chunks_seq
\seq_new:N \l_gg_bib_order_seq
\prop_new:N \l_gg_bib_items_prop
\clist_const:Nn \c_gg_main_bib_keys_clist {kid2018,laplacian1983,foveated2026,geodiffusion2024,soda2023,sd32024,fid2017,cfg2022,laytrol2026,vae2014,flux2klein2026,gligen2023,retinanet2017,flowmatching2023,rectifiedflow2023,adamw2019,lot2026,dit2023,ldm2022,terradit2026,rope2024,scenefid2021,fourier2020,aerogen2025,ficgen2025,instancediffusion2024,terraditomega2026,pma2025,dota2018,dcfl2023,aitodr2026,qwen32025,h2rbox2023,ofdiff2026,point2rbox2024,creatilayout2025,controlnet2023,ccdiff2024,ccdiffpp2025,layout2im2019,layoutdiffusion2023,migc2024,mmrotate2022}
\cs_new_protected:Npn \gg_bib_store:w [#1]#2#3\q_stop
  {
    \prop_put:Nnn \l_gg_bib_items_prop {#2} {\bibitem[#1]{#2}#3}
    \seq_put_right:Nn \l_gg_bib_order_seq {#2}
  }
\cs_new_protected:Npn \gg_bib_emit:n #1
  {
    \prop_get:NnNT \l_gg_bib_items_prop {#1} \l_tmpa_tl
      {\prop_remove:Nn \l_gg_bib_items_prop {#1} \tl_use:N \l_tmpa_tl \par}
  }
\cs_new_protected:Npn \gg_bib_strip_end:w #1\end#2#3\q_stop {\tl_set:Nn \l_gg_bib_file_tl {#1}}
\cs_new_eq:NN \gg_bib_original_input:n \@input@
\cs_new_protected:Npn \gg_bib_input:n #1
  {
    \str_if_eq:eeTF {#1} {\jobname.bbl}
      {
        \file_get:nnNTF {#1} {} \l_gg_bib_file_tl
          {
            \exp_after:wN \gg_bib_strip_end:w \l_gg_bib_file_tl \q_stop
            \seq_set_split:NnV \l_gg_bib_chunks_seq {\bibitem} \l_gg_bib_file_tl
            \seq_pop_left:NN \l_gg_bib_chunks_seq \l_gg_bib_header_tl
            \seq_clear:N \l_gg_bib_order_seq
            \prop_clear:N \l_gg_bib_items_prop
            \seq_map_inline:Nn \l_gg_bib_chunks_seq {\gg_bib_store:w ##1\q_stop}
            \tl_use:N \l_gg_bib_header_tl
            \clist_map_function:NN \c_gg_main_bib_keys_clist \gg_bib_emit:n
            \seq_map_function:NN \l_gg_bib_order_seq \gg_bib_emit:n
            \end{thebibliography}
          }
          {\gg_bib_original_input:n {#1}}
      }
      {\gg_bib_original_input:n {#1}}
  }
\cs_new_eq:NN \gg_bib_original_bibliography:n \bibliography
\protected\def\bibliography#1
  {
    \group_begin:
    \cs_set_eq:NN \@input@ \gg_bib_input:n
    \gg_bib_original_bibliography:n {#1}
    \group_end:
  }
\ExplSyntaxOff
\makeatother

\fi

\def\confName{CVPR}
\def\confYear{2027}

\iflatexml
\newcommand{\thetitle}{GeoGrain: Granularity-Aware Layout-to-Image Generation \\for Small Objects in Remote Sensing Imagery}
\providecommand{\eg}{e.g.\xspace}
\fi

\title{GeoGrain: Granularity-Aware Layout-to-Image Generation 
\\for Small Objects in Remote Sensing Imagery}

\iflatexml
\newcommand{\GeoGrainHTMLmark}{\textsuperscript{\textdagger}}
\newcommand{\GeoGrainHTMLstrut}{\vphantom{\textsuperscript{\textdagger}}}
\author{%
Taehwan Kim\GeoGrainHTMLstrut\\
KAIST\\
\href{mailto:taehwankim@kaist.ac.kr}{\texttt{taehwankim@kaist.ac.kr}}
\and
Jeonghyeok Do\GeoGrainHTMLmark\\
KAIST\\
\href{mailto:ehwjdgur0913@kaist.ac.kr}{\texttt{ehwjdgur0913@kaist.ac.kr}}
\and
Munchurl Kim\GeoGrainHTMLmark\\
KAIST\\
\href{mailto:mkimee@kaist.ac.kr}{\texttt{mkimee@kaist.ac.kr}}
}
\else
\author{%
  \makebox[0pt][c]{%
    \raisebox{-7pt}[0pt][0pt]{%
      \begin{tabular}[c]{@{}c@{}c@{}c@{}}
        \makebox[0.333333\textwidth][c]{Taehwan Kim} &
        \makebox[0.333333\textwidth][c]{Jeonghyeok Do\textsuperscript{\normalfont\footnotesize\dag}} &
        \makebox[0.333333\textwidth][c]{Munchurl Kim\textsuperscript{\normalfont\footnotesize\dag}}\\[5pt]
        KAIST & KAIST & KAIST\\[1pt]
        {\small\email{taehwankim@kaist.ac.kr}} &
        {\small\email{ehwjdgur0913@kaist.ac.kr}} &
        {\small\email{mkimee@kaist.ac.kr}}\\[1pt]
        \multicolumn{3}{c}{{\small\url{https://kaist-viclab.github.io/GeoGrain_site}}}
      \end{tabular}%
    }%
  }%
  \strut
}
\fi

\begin{document}
\maketitle
\iflatexml
\begin{center}
\small
\url{https://kaist-viclab.github.io/GeoGrain_site}\newline
{\footnotesize\textsuperscript{\textdagger} Co-corresponding authors (equal advising).}
\end{center}
\else
\makeatletter
\ifGeoGrainReview\else
\AddToShipoutPicture*{%
  \AtTextLowerLeft{%
    \put(0,-25){\makebox(0,0)[lb]{%
      \vbox{%
        \hsize\columnwidth
        \normalfont\footnotesize
        \footnoterule
        \def\@thefnmark{\dag}%
        \@makefntext{\rule{0pt}{\footnotesep}Co-corresponding authors (equal advising).}%
      }%
    }}%
  }%
}
\fi
\makeatother
\fi

\iflatexml
\iflatexml
\begin{minipage}{2\textwidth}
\section*{\lxAddClass{ltx_title_abstract}Abstract}
\lxAddClass{ltx_abstract}
\else
\begin{abstract}
\fi
Layout-to-image (L2I) generation for remote sensing imagery requires realistic small objects aligned with prescribed oriented bounding boxes (OBBs). However, global box descriptions in prior L2I methods leave local geometry within image-token regions implicit. Coarse, uniform latent grids may also represent small objects with only one or a few image tokens. Together, these constraints hinder the generation of small objects with accurate shapes and orientations. To address these limitations, we present \textbf{GeoGrain}, a granularity-aware L2I framework for precise small-object control in remote sensing imagery. GeoGrain introduces (i) Granularity-Aware Geometry Encoding (GAGE), which constructs layout tokens from global OBB descriptions with local geometric conditions at image-token positions, and (ii) Granularity-Aware Token Expansion (GATE), which subdivides image-token grid cells overlapping the input OBBs, producing finer image tokens while retaining coarse tokens elsewhere. GAGE encodes local OBB geometry on the resulting mixed-resolution grid, aligning each layout token with its corresponding image token. Experiments on the oriented small-object detection datasets SODA-A and AI-TOD-R demonstrate state-of-the-art layout fidelity alongside high generation quality. On these datasets, GeoGrain improves AP over the strongest baseline by 18.6 and 6.3 points, respectively, reaching AP comparable to that measured on real images. Data augmentation with our generated images also improves downstream detection and outperforms augmentation with images from the baselines.
\iflatexml
\end{minipage}
\else
\end{abstract}
\fi

\fi

\iflatexml\begin{figure*}\else\begin{strip}\fi
    \centering
    \iflatexml\else\renewcommand{\HyperRaiseLinkDefault}{24pt}\fi\captionsetup{type=figure}
    \vspace{-0.6cm}
    \iflatexml
    \includegraphics[width=2\textwidth,alt={Small-object generation using the same SODA-A layout, with enlarged comparison regions on the left and normalized generation-quality and layout-fidelity radar charts for SODA-A and AI-TOD-R on the right.}]{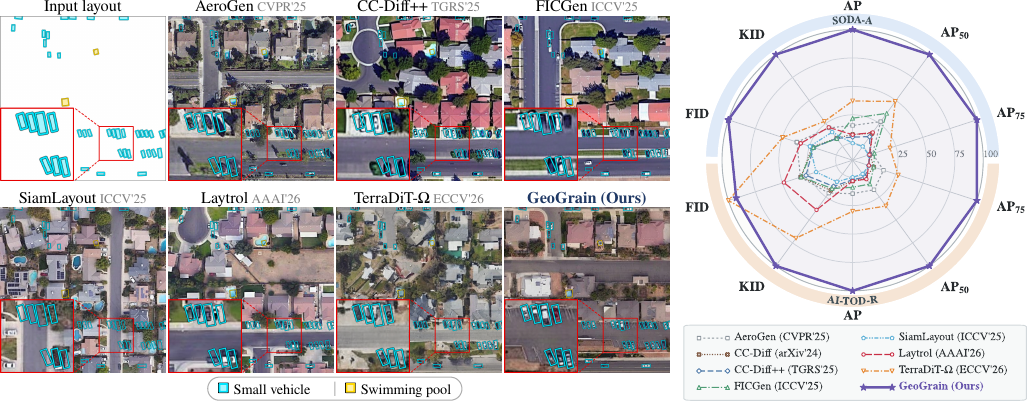}
    \else
    \includegraphics[width=\textwidth]{figures/teaser.pdf}
    \fi
    \vspace{-0.6cm}
    \caption{%
        \textbf{Precise small-object control with GeoGrain.}
        \textit{Left:} Qualitative comparison on the
        SODA-A~\cite{soda2023} dataset using the same input layout.
        Insets show enlarged red-boxed regions, highlighting
        object detail and OBB alignment.
        \textit{Right:} Generation quality and Layout fidelity
        on the SODA-A and AI-TOD-R~\cite{aitodr2026} datasets.
        Higher normalized scores indicate better performance.
    }
    \label{fig:teaser}
\iflatexml\end{figure*}\else\end{strip}\fi

\iflatexml\else
\iflatexml
\begin{minipage}{2\textwidth}
\section*{\lxAddClass{ltx_title_abstract}Abstract}
\lxAddClass{ltx_abstract}
\else
\begin{abstract}
\fi
Layout-to-image (L2I) generation for remote sensing imagery requires realistic small objects aligned with prescribed oriented bounding boxes (OBBs). However, global box descriptions in prior L2I methods leave local geometry within image-token regions implicit. Coarse, uniform latent grids may also represent small objects with only one or a few image tokens. Together, these constraints hinder the generation of small objects with accurate shapes and orientations. To address these limitations, we present \textbf{GeoGrain}, a granularity-aware L2I framework for precise small-object control in remote sensing imagery. GeoGrain introduces (i) Granularity-Aware Geometry Encoding (GAGE), which constructs layout tokens from global OBB descriptions with local geometric conditions at image-token positions, and (ii) Granularity-Aware Token Expansion (GATE), which subdivides image-token grid cells overlapping the input OBBs, producing finer image tokens while retaining coarse tokens elsewhere. GAGE encodes local OBB geometry on the resulting mixed-resolution grid, aligning each layout token with its corresponding image token. Experiments on the oriented small-object detection datasets SODA-A and AI-TOD-R demonstrate state-of-the-art layout fidelity alongside high generation quality. On these datasets, GeoGrain improves AP over the strongest baseline by 18.6 and 6.3 points, respectively, reaching AP comparable to that measured on real images. Data augmentation with our generated images also improves downstream detection and outperforms augmentation with images from the baselines.
\iflatexml
\end{minipage}
\else
\end{abstract}
\fi

\fi    
\section{Introduction}
\label{sec:introduction}
Layout-to-image (L2I) generation synthesizes realistic images from spatial layouts, typically specified by category-labeled bounding boxes~\cite{gligen2023,layout2im2019,layoutdiffusion2023}. Users can explicitly control object categories, locations, and sizes, while the model generates object appearance and a coherent surrounding scene. In remote sensing (aerial or satellite) imagery, layout control must accommodate objects that are often small, densely arranged, and arbitrarily oriented~\cite{soda2023,dota2018,aitodr2026}. Oriented bounding boxes (OBBs) capture these orientations along with object locations and sizes, but obtaining accurate annotations is labor-intensive, especially when object boundaries are difficult to discern~\cite{aitodr2026,h2rbox2023,point2rbox2024}. L2I generation can therefore augment datasets with synthetic images paired with their input OBB layouts~\cite{aerogen2025}. For reliable detector training, the generated objects must faithfully match the prescribed classes and OBBs.

Existing diffusion-based remote sensing L2I methods have extended layout control to oriented objects~\cite{aerogen2025} and modeled relationships between objects and their surroundings~\cite{ficgen2025,ccdiff2024,ccdiffpp2025}. Despite improvements in scene composition and contextual consistency, generating realistic small objects that accurately match their prescribed OBBs remains challenging, as illustrated in Fig.~\ref{fig:teaser}. Recent advances in diffusion transformer (DiT)-based~\cite{dit2023} L2I generation~\cite{laytrol2026,creatilayout2025} offer a promising foundation for remote sensing L2I generation. However, simply extending their horizontal bounding box (HBB) inputs to OBBs retains object-level geometric conditioning and a coarse, uniform latent grid, both of which limit precise control over small objects.

To address these limitations, we introduce \textbf{GeoGrain}, a granularity-aware L2I framework for precise small-object control in remote sensing imagery. Built on a pretrained DiT~\cite{flux2klein2026}, GeoGrain integrates two complementary modules: \textit{Granularity-Aware Geometry Encoding (GAGE)} and \textit{Granularity-Aware Token Expansion (GATE)}.

\begin{figure}[t]
    \centering
    \iflatexml
    \includegraphics[width=2\textwidth,alt={Comparison of prior DiT-based layout-to-image methods with GeoGrain: local oriented-box geometry enriches layout tokens, and finer image tokens are allocated only to cells overlapping input objects.}]{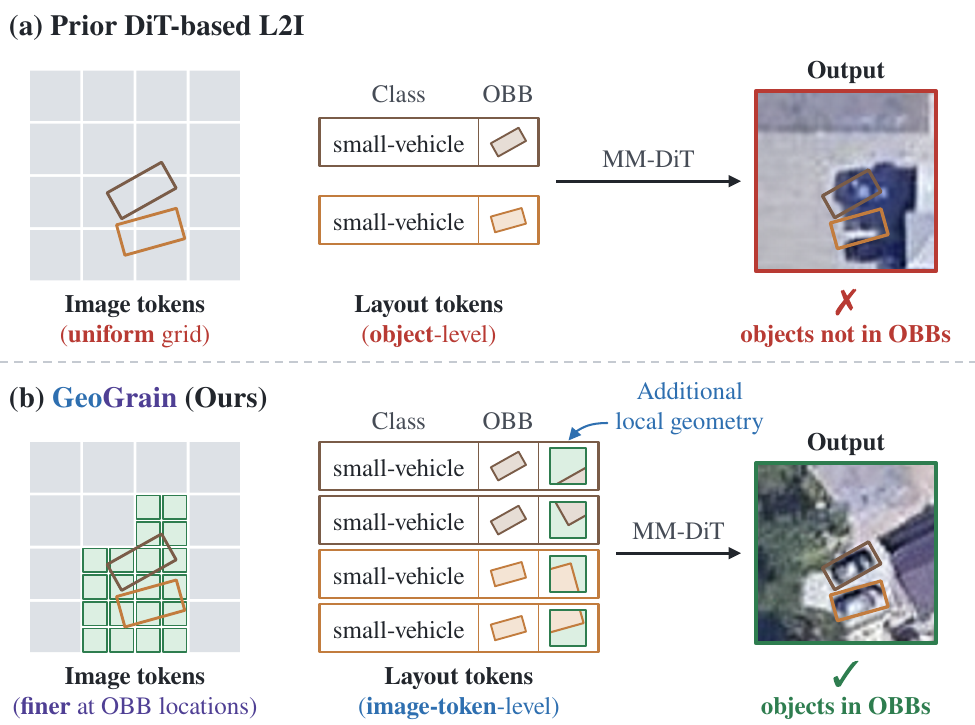}
    \else
    \includegraphics[width=\linewidth]{figures/motivation.pdf}
    \fi
    \caption{
    \textbf{Granularity-aware conditioning and synthesis.}
    (a) Prior DiT-based L2I methods~\cite{laytrol2026,creatilayout2025} use object-level box conditions and a coarse, uniform image-token grid, leaving local OBB geometry implicit and only a few tokens available for small objects. (b) GeoGrain adds  local OBB geometry to layout tokens at corresponding image-token positions and subdivides only the image-token grid cells overlapping the input OBBs.}
    \label{fig:motivation}
\end{figure}

\emph{(i) Granularity of geometric conditioning.} As shown in Fig.~\ref{fig:motivation}(a), prior methods describe each object using its class and global box coordinates. The model must infer how each box aligns with individual image-token regions, hindering precise OBB alignment. In contrast, \textbf{GAGE} in Fig.~\ref{fig:motivation}(b) introduces additional local geometric conditions alongside the object-level description. For each OBB, it constructs a layout token for every overlapping image-token region, so a single object can have multiple layout tokens describing different portions of its box. When multiple objects share the same image-token region, each retains its own layout token, keeping their geometric conditions distinct.

\emph{(ii) Granularity of latent image tokens.} As depicted in Fig.~\ref{fig:motivation}(a), small objects may have only one or a few image tokens available on a coarse, uniform latent grid, limiting fine-grained control over their shapes and orientations. In Fig.~\ref{fig:motivation}(b), \textbf{GATE} subdivides image-token grid cells overlapping the input OBBs, providing finer image tokens for object synthesis while preserving the original grid elsewhere. GAGE constructs its layout tokens on the resulting mixed-resolution grid, matching local geometric conditions to the corresponding image-token positions. Fine and coarse image tokens are jointly denoised, allowing object regions to interact with the surrounding scene context.

We evaluate GeoGrain on two remote sensing datasets specifically designed for oriented small-object detection, SODA-A~\cite{soda2023} and AI-TOD-R~\cite{aitodr2026}. Quantitative and qualitative results demonstrate its effectiveness in generating realistic small objects aligned with their prescribed OBBs.

Our contributions are summarized as follows:
\begin{itemize}
    \item
    We introduce \textbf{GeoGrain}, a granularity-aware framework for remote sensing L2I generation. To the best of our knowledge, this is the first L2I study on remote sensing benchmarks specifically designed for oriented small-object detection.

    \item
    We propose \textbf{GAGE}, which introduces instance-specific local OBB conditions at individual image-token positions, and \textbf{GATE}, which subdivides image-token grid cells overlapping the input OBBs. Their shared grid aligns local geometric guidance with the image tokens used for synthesis.

    \item
    We achieve state-of-the-art layout fidelity alongside high generation quality, surpassing the strongest baseline in AP by 18.6 points on SODA-A and 6.3 on AI-TOD-R, reaching AP comparable to that on real images. Augmentation with our generated images also improves oriented small-object detection, outperforming all baselines.
\end{itemize}

\section{Related Work}
\label{sec:related_work}

\subsection{Layout-to-Image Generation}
Earlier diffusion-based L2I approaches use trainable attention modules to incorporate object descriptions and bounding boxes into pretrained U-Net models~\cite{gligen2023,instancediffusion2024,migc2024}. Recent dual-branch L2I approaches based on multimodal diffusion transformers (MM-DiTs)~\cite{sd32024} separate image--text and image--layout interactions. SiamLayout~\cite{creatilayout2025} uses this separation to reduce modality competition, while Laytrol~\cite{laytrol2026} emphasizes pretrained knowledge preservation through parameter copying and zero-initialized output projections. However, their layout encodings do not explicitly describe how each box intersects individual image-token regions. This leaves the denoising network to infer the local geometric relationships needed for precise spatial control. GeoGrain introduces additional layout conditions that explicitly describe local OBB geometry at image-token positions.

\subsection{L2I Generation for Remote Sensing}
Recent remote sensing L2I methods focus on improving layout adherence, object fidelity~\cite{ofdiff2026}, and contextual coherence. AeroGen~\cite{aerogen2025} supports horizontal and oriented boxes through layout embeddings and masked attention. CC-Diff~\cite{ccdiff2024} and CC-Diff++~\cite{ccdiffpp2025} model foreground--background dependencies, while FICGen~\cite{ficgen2025} uses frequency-aware guidance and an instance coherence map for contextual disentanglement. Among DiT-based methods, TerraDiT~\cite{terradit2026} uses adaptive local attention for point conditioning, while TerraDiT-$\Omega$~\cite{terraditomega2026} extends control to boxes, polygons, and polylines through geometry-aware local attention. However, these methods retain coarse, uniform latent grids. Small objects may therefore occupy only a few spatial elements, making precise synthesis difficult. GeoGrain addresses this limitation by providing finer image tokens in grid cells overlapping the input OBBs.

\subsection{Spatially Adaptive Diffusion Models}
Spatially adaptive diffusion methods vary token density to balance computational cost and visual detail. Phase-Aligned Mixed-Resolution Attention (PMA)~\cite{pma2025} aligns Rotary Position Embedding (RoPE)~\cite{rope2024} phases across resolutions for frozen DiTs without training. Foveated Diffusion~\cite{foveated2026} reduces the token count by representing peripheral regions with coarser latent tokens, while Level-of-Token (LoT) Diffusion~\cite{lot2026} uses patch-wise asymmetric flow to preserve full-resolution latent flow prediction with fewer tokens. While these methods emphasize efficiency by using coarser tokens in background regions, GeoGrain subdivides image-token grid cells overlapping the input OBBs for precise small-object synthesis. It encodes instance-specific local OBB geometry on the resulting grid, matching geometric conditioning to the image-token granularity.

\GeoGrainDeferFloat{methodfigure}{%
\begin{figure*}[t]
\centering
\iflatexml
\includegraphics[width=2\textwidth,alt={GeoGrain architecture with frozen image-text and trainable image-layout branches, geometry-aware GAGE layout tokens, GATE expansion of object-overlapping image tokens, and blended coarse- and fine-scale reconstruction.}]{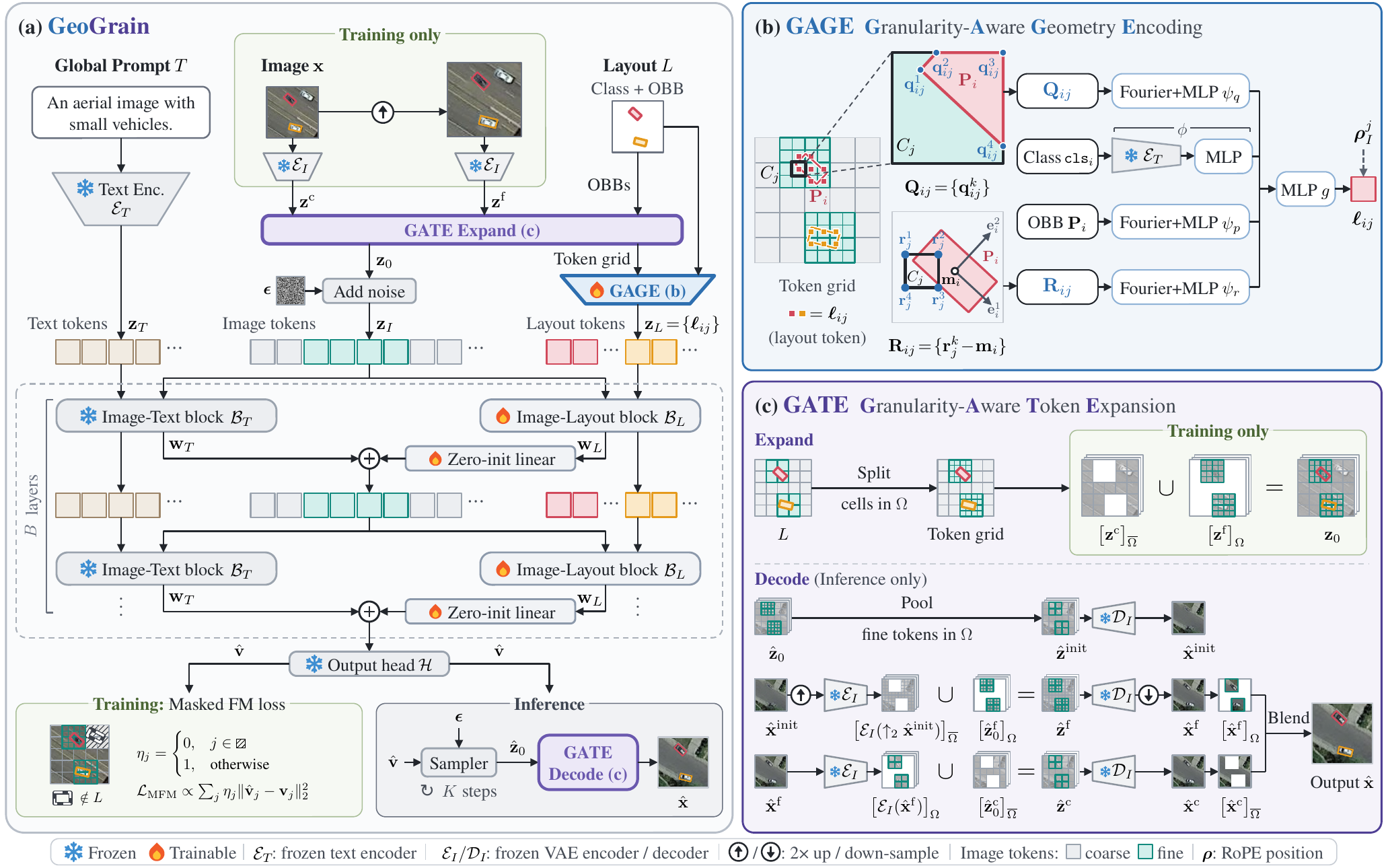}
\else
\includegraphics[width=\textwidth]{figures/method.pdf}
\fi
\caption{
\textbf{Overview of GeoGrain.}
(a) Frozen image--text and trainable image--layout branches
jointly denoise mixed-resolution image tokens.
(b) On the mixed-resolution grid, GAGE combines class labels,
global OBB coordinates, local intersection geometry
\(\mathbf{Q}_{ij}\), and object-relative coordinates
\(\mathbf{R}_{ij}\) into one layout token per intersecting
OBB--cell pair.
(c) GATE replaces each coarse token whose grid cell overlaps
an input OBB with four fine tokens, retaining coarse tokens
elsewhere.
For reconstruction, uniform latent grids are assembled
at coarse and fine scales, and the decoded images are blended.
The local descriptors in (b) are schematic;
full definitions appear in Sec.~\ref{sec:gage}.
}
\label{fig:overview}
\end{figure*}
}
\section{Method}

Given a global text prompt \(T\) and an object layout \(L\), GeoGrain aims to synthesize a realistic remote sensing image \(\mathbf{x}\in\mathbb{R}^{H\times W\times3}\) of height \(H\) and width \(W\) that adheres to both conditions.
The layout \(L=\{(\mathbf{P}_i,\mathtt{cls}_i)\}_{i=1}^{M}\) specifies \(M\) objects, where \(\mathtt{cls}_i\) denotes the class label of object \(i\), and \(\mathbf{P}_i=\{(x_{ik},y_{ik})\}_{k=1}^{4}\in\mathbb{R}^8\) contains its four ordered oriented bounding box (OBB) vertices.
Each vertex \((x_{ik},y_{ik})\) is expressed in normalized image coordinates, with horizontal and vertical coordinates divided by \(W\) and \(H\), respectively.

As illustrated in Fig.~\ref{fig:overview}, GeoGrain adopts an MM-DiT architecture with image--text and image--layout branches (Sec.~\ref{sec:geograin}). GAGE encodes local OBB geometry into layout tokens (Sec.~\ref{sec:gage}), while GATE subdivides image-token grid cells overlapping the input OBBs (Sec.~\ref{sec:gate}). GAGE constructs its layout tokens on the resulting mixed-resolution grid, matching local geometric conditions to the corresponding image-token positions. The training objective and inference procedure are described in Sec.~\ref{sec:training}.

\GeoGrainShowFloat{methodfigure}

\subsection{GeoGrain}
\label{sec:geograin}

As illustrated in Fig.~\ref{fig:overview}(a), GeoGrain builds on a pretrained text-to-image diffusion transformer~\cite{flux2klein2026} operating on latent image tokens from a variational autoencoder (VAE)~\cite{vae2014,ldm2022}. Following prior work on dual-branch L2I generation~\cite{laytrol2026}, we initialize a trainable image--layout branch from the pretrained transformer weights while keeping the image--text branch, text encoder, and VAE frozen.

The frozen text encoder \(\mathcal{E}_T\) maps the global prompt \(T\) to text tokens \(\mathbf{z}_T=\mathcal{E}_T(T)\).
During training, we construct clean mixed-resolution image tokens \(\mathbf{z}_0\) from the input image \(\mathbf{x}\) and layout \(L\) with the frozen VAE encoder \(\mathcal{E}_I\) and GATE before adding noise to obtain noisy image tokens \(\mathbf{z}_I\).
On the same mixed-resolution grid, GAGE encodes \(L\) into layout tokens \(\mathbf{z}_L\).
At each layer \(n=1,\ldots,B\), where \(B\) is the number of
transformer layers, the image--text block \(\mathcal{B}_T^n\)
and image--layout block \(\mathcal{B}_L^n\) process shared
image features with their respective conditioning streams:
\iflatexml\begin{minipage}{2\textwidth}\fi%
\begin{equation}
\begin{aligned}
(\mathbf{z}_{I}^{0},\mathbf{z}_{T}^{0},\mathbf{z}_{L}^{0})
&=
(\mathbf{z}_{I},\mathbf{z}_{T},\mathbf{z}_{L}),\\
(\mathbf{w}_{T}^{n},\mathbf{z}_{T}^{n})
&=
\mathcal{B}_{T}^{n}
(\mathbf{z}_{I}^{n-1},\mathbf{z}_{T}^{n-1};\tau),\\
(\mathbf{w}_{L}^{n},\mathbf{z}_{L}^{n})
&=
\mathcal{B}_{L}^{n}
(\mathbf{z}_{I}^{n-1},\mathbf{z}_{L}^{n-1};\tau),\\
\mathbf{z}_{I}^{n}
&=
\mathbf{w}_{T}^{n}
+
\operatorname{linear}(\mathbf{w}_{L}^{n}),
\end{aligned}
\label{eq:block}
\end{equation}
\iflatexml\end{minipage}\fi%
where \(\tau\) is the timestep, \(\mathbf{w}_{T}^{n}\) and \(\mathbf{w}_{L}^{n}\) are the branch-specific image features, and \(\operatorname{linear}\) is a layer-specific trainable linear projection with zero-initialized weights and bias~\cite{laytrol2026,controlnet2023}.
The pretrained output head \(\mathcal{H}\) remains frozen
and maps the final image features to the predicted velocity
\(\hat{\mathbf{v}}=\mathcal{H}(\mathbf{z}_I^{B};\tau)\).

\subsection{Granularity-Aware Geometry Encoding}
\label{sec:gage}

As depicted in Fig.~\ref{fig:overview}(b), GAGE constructs the layout-token sequence \(\mathbf{z}_L\) by explicitly encoding the geometric relationship between each OBB \(\mathbf{P}_i\) and the image-token regions it intersects.
Let \(C_j\) denote the square grid cell associated with the \(j\)-th image token, expressed in output-image pixel coordinates.
For each object \(i\) whose OBB overlaps \(C_j\) with positive area, we create a separate layout token \(\boldsymbol{\ell}_{ij}\).
Each token combines the object-level class label \(\mathtt{cls}_i\) and global OBB coordinates \(\mathbf{P}_i\) with two complementary local descriptors: (i) \(\mathbf{Q}_{ij}\), which describes the portion of the OBB \(\mathbf{P}_i\) within \(C_j\), and (ii) \(\mathbf{R}_{ij}\), which expresses the position of \(C_j\) relative to object $i$.

\noindent\textbf{Local intersection geometry.}
We rescale the vertices in \(\mathbf{P}_i\) to output-image
pixel coordinates and intersect the resulting OBB with \(C_j\).
Each intersection vertex \(\mathbf{q}_{ij}^{k}\) is normalized to cell coordinates as
\iflatexml\begin{minipage}{2\textwidth}\fi%
\begin{equation}
\widetilde{\mathbf{q}}_{ij}^{k}
=
(\mathbf{q}_{ij}^{k}-\mathbf{a}_j)/s_j
\in[0,1]^2,
\label{eq:equation_2}
\end{equation}
\iflatexml\end{minipage}\fi%
where \(\mathbf{a}_j\) and \(s_j\) denote the top-left corner and side length of \(C_j\), respectively.
Since the intersection polygon has at most eight vertices, we arrange the normalized vertices in a consistent order, pad the sequence to eight slots, and concatenate their coordinates to form
\(\mathbf{Q}_{ij}=\{\widetilde{\mathbf{q}}_{ij}^{k}\}_{k=1}^{8}\in\mathbb{R}^{16}\).

\noindent\textbf{Object-relative coordinates.}
To describe the position of \(C_j\) relative to object \(i\),
we express the four corners of \(C_j\) in an object coordinate
system. \(\mathbf{m}_i\) denotes the OBB center, and
\(\mathbf{e}_i^{1}\) and \(\mathbf{e}_i^{2}\) are unit vectors along
the long and short OBB axes, respectively.
The corresponding side lengths are \(w_i\) and \(h_i\),
with \(w_i\geq h_i\).
For each corner \(\mathbf{r}_j^k\), \(k=1,\ldots,4\), of \(C_j\),
we compute normalized object-relative coordinates as
\iflatexml\begin{minipage}{2\textwidth}\fi%
\begin{equation}
\widetilde{\mathbf{r}}_{ij}^{k}
=
\begin{bmatrix}
(\mathbf{e}_i^{1})^{\top}/\max(w_i,s_j)\\
(\mathbf{e}_i^{2})^{\top}/\max(h_i,s_j)
\end{bmatrix}
(\mathbf{r}_j^k-\mathbf{m}_i)
+
0.5\begin{bmatrix}1\\1\end{bmatrix},
\end{equation}
\iflatexml\end{minipage}\fi%
where the normalization maps the object center to
\((0.5,0.5)\) and the OBB into \([0,1]^2\).
Using the cell side length \(s_j\) as a minimum denominator
avoids excessively large relative coordinates.
We then concatenate the four coordinate pairs in a fixed corner order
to form
\(\mathbf{R}_{ij}
=\{\widetilde{\mathbf{r}}_{ij}^{k}\}_{k=1}^{4}
\in\mathbb{R}^{8}\).

\noindent\textbf{Layout-token construction.}
We combine the four encoded components to construct each layout token:
\iflatexml\begin{minipage}{2\textwidth}\fi%
\begin{equation}
\boldsymbol{\ell}_{ij}
=
g\!\left(
\left[
\phi(\mathtt{cls}_i)
\mid\psi_p(\mathbf{P}_i)
\mid\psi_q(\mathbf{Q}_{ij})
\mid\psi_r(\mathbf{R}_{ij})
\right]
\right),
\label{eq:equation_4}
\end{equation}
\iflatexml\end{minipage}\fi%
where \(\mid\) denotes concatenation and \(g\) is a multilayer
perceptron (MLP) projecting the concatenated features to the
transformer hidden dimension \(d\).
The class encoder \(\phi\) encodes each class name with the
frozen text encoder followed by an MLP, while the geometric
encoders \(\psi_p\), \(\psi_q\), and \(\psi_r\) encode the
corresponding geometric descriptors with Fourier
encoding~\cite{fourier2020} followed by separate MLPs.
For positional alignment, each layout token shares RoPE
position indices with its associated image token \(j\),
\(\boldsymbol{\rho}_L^{ij}
=\boldsymbol{\rho}_I^{j}\).
The resulting tokens initialize the layout stream as
\(\mathbf{z}_L=\{\boldsymbol{\ell}_{ij}\}\).
All geometric descriptors are derived solely from the input
OBBs and token grid, without additional annotations.

\subsection{Granularity-Aware Token Expansion}
\label{sec:gate}

As shown in Fig.~\ref{fig:overview}(c), GATE selects coarse
grid cells that overlap any input OBB with positive area
and subdivides each selected cell into a \(2\times2\) grid,
replacing its coarse token with four fine tokens.
Let \(\Omega\) denote the image region covered by the selected
cells and \(\overline{\Omega}\) its complement within the image.
We use fine tokens in \(\Omega\) and retain coarse tokens in
\(\overline{\Omega}\), forming a mixed-resolution grid that remains
fixed throughout denoising.
Coarse and fine tokens take RoPE position indices in a shared
coordinate system defined at the fine-token resolution.
On the resulting grid, GAGE computes local geometric
conditions to construct layout tokens at the corresponding
image-token positions.

\noindent\textbf{Mixed-resolution latent construction.}
The frozen VAE operates on uniform spatial grids.
Inspired by prior work on mixed-resolution
tokenization~\cite{foveated2026}, we therefore construct
the clean mixed-resolution image-token sequence \(\mathbf{z}_0\)
from separate encodings of the training image \(\mathbf{x}\)
and its \(2\times\) upsampled version as
\iflatexml\begin{minipage}{2\textwidth}\fi%
\begin{equation}
\begin{aligned}
&\mathbf{z}^{\mathrm{c}}=\mathcal{E}_I(\mathbf{x}),\quad
\mathbf{z}^{\mathrm{f}}=\mathcal{E}_I(\uparrow_2\mathbf{x}),\\
&\mathbf{z}_0
=\operatorname{merge}\bigl(
\bigl[\mathbf{z}^{\mathrm{c}}\bigr]_{\overline{\Omega}},
\bigl[\mathbf{z}^{\mathrm{f}}\bigr]_{\Omega}
\bigr),
\end{aligned}
\label{eq:equation_5}
\end{equation}
\iflatexml\end{minipage}\fi%
where \(\mathbf{z}^{\mathrm{c}}\) and \(\mathbf{z}^{\mathrm{f}}\)
denote the coarse and fine grids of image tokens, respectively,
and \(\uparrow_2\) doubles both image dimensions.
The coarse and fine grids are derived from the same training
image, requiring no additional high-resolution images.
To form \(\mathbf{z}_0\), the operators
\([\cdot]_{\vphantom{\overline{\Omega}}\Omega}\)
and \([\cdot]_{\overline{\Omega}}\)
select tokens from the corresponding regions of each grid,
and \(\operatorname{merge}\) assembles the tokens according
to their spatial positions.

\noindent\textbf{Decoding at inference.}
To decode $\hat{\mathbf{z}}_0$ with the frozen VAE decoder
$\mathcal{D}_I$ after mixed-resolution denoising,
we first complete a coarse grid by pooling
the fine tokens in $\Omega$:
\iflatexml\begin{minipage}{2\textwidth}\fi%
\begin{equation}
\hat{\mathbf{z}}^{\mathrm{init}}
=
\operatorname{merge}\bigl(
\bigl[\hat{\mathbf{z}}^{\mathrm{c}}_0\bigr]_{\overline{\Omega}},
\operatorname{pool}\bigl(
\bigl[\hat{\mathbf{z}}^{\mathrm{f}}_0\bigr]_{\Omega}
\bigr)
\bigr),
\end{equation}
\iflatexml\end{minipage}\fi%
where $\hat{\mathbf{z}}^{\mathrm{c}}_0$ and $\hat{\mathbf{z}}^{\mathrm{f}}_0$ are the coarse and fine tokens of $\hat{\mathbf{z}}_0$,
$\operatorname{pool}$ is spatial average pooling,
and decoding $\hat{\mathbf{z}}^{\mathrm{init}}$ gives 
the initial reconstruction
${\hat{\mathbf{x}}^{\mathrm{init}}
=\mathcal{D}_I(\hat{\mathbf{z}}^{\mathrm{init}})}$.
Since pooling can suppress fine spatial detail in $\Omega$,
we complete the fine grid by retaining the denoised fine
tokens in $\Omega$ and filling $\overline{\Omega}$ with tokens
obtained by re-encoding the $2\times$ upsampled
$\hat{\mathbf{x}}^{\mathrm{init}}$:
\iflatexml\begin{minipage}{2\textwidth}\fi%
\begin{equation}
\hat{\mathbf{z}}^{\mathrm{f}}
=
\operatorname{merge}\bigl(
\bigl[
\mathcal{E}_I(\uparrow_2\hat{\mathbf{x}}^{\mathrm{init}})
\bigr]_{\overline{\Omega}},
\bigl[\hat{\mathbf{z}}^{\mathrm{f}}_0\bigr]_{\Omega}
\bigr).
\end{equation}
\iflatexml\end{minipage}\fi%
We decode $\hat{\mathbf{z}}^{\mathrm{f}}$ and downsample
the resulting image by $2\times$ to obtain
${\hat{\mathbf{x}}^{\mathrm{f}}
={\downarrow_2}\,\mathcal{D}_I(\hat{\mathbf{z}}^{\mathrm{f}})}$
at the original $H\times W$ resolution.
However, the $\overline{\Omega}$ region of
$\hat{\mathbf{x}}^{\mathrm{f}}$ is obtained through
an additional VAE encoding--decoding cycle, which can
introduce reconstruction error.
We therefore also obtain a coarse reconstruction for
$\overline{\Omega}$.
To form the coarse grid, we retain the denoised coarse
tokens in $\overline{\Omega}$ and replace the initially pooled
tokens in $\Omega$ with tokens obtained by re-encoding
$\hat{\mathbf{x}}^{\mathrm{f}}$:
\iflatexml\begin{minipage}{2\textwidth}\fi%
\begin{equation}
\hat{\mathbf{z}}^{\mathrm{c}}
=
\operatorname{merge}\bigl(
\bigl[\hat{\mathbf{z}}^{\mathrm{c}}_0\bigr]_{\overline{\Omega}},
\bigl[
\mathcal{E}_I(\hat{\mathbf{x}}^{\mathrm{f}})
\bigr]_{\Omega}
\bigr).
\end{equation}
\iflatexml\end{minipage}\fi%
Finally, we decode $\hat{\mathbf{z}}^{\mathrm{c}}$ to obtain
${\hat{\mathbf{x}}^{\mathrm{c}}
=\mathcal{D}_I(\hat{\mathbf{z}}^{\mathrm{c}})}$
and combine $\hat{\mathbf{x}}^{\mathrm{f}}$ in $\Omega$
with $\hat{\mathbf{x}}^{\mathrm{c}}$ in $\overline{\Omega}$
using Laplacian-pyramid blending~\cite{laplacian1983},
yielding the final image $\hat{\mathbf{x}}$.

\GeoGrainDeferFloat{datasets}{%
\begin{table}[t]
\caption{\textbf{Dataset statistics.}
Object size \(\sqrt{wh}\) is reported as mean \(\pm\)
standard deviation, where \(w,h\) are OBB side lengths.}
\label{tab:datasets}
\GeoGrainTableFont{\scriptsize}
\centering
\setlength{\tabcolsep}{6pt}
\renewcommand{\arraystretch}{1.15}
\resizebox{1.0\columnwidth}{!}{%
\begin{tabular}{lllll}
\toprule
\textbf{Dataset} & \multicolumn{2}{c}{\textbf{SODA-A}~\cite{soda2023}} & \multicolumn{2}{c}{\textbf{AI-TOD-R}~\cite{aitodr2026}} \\
\midrule
Classes & \multicolumn{2}{c}{9} & \multicolumn{2}{c}{8} \\
Images (train / test) & \multicolumn{2}{c}{23,032 / 15,989} & \multicolumn{2}{c}{7,752 / 9,705} \\
Objects (train / test) & \multicolumn{2}{c}{347,884 / 300,158} & \multicolumn{2}{c}{131,274 / 154,083} \\
Object size (px) & \multicolumn{2}{c}{16.2\,$\pm$\,7.4} & \multicolumn{2}{c}{10.8\,$\pm$\,4.8} \\
\midrule
\multirow{4}{*}{Size range (px$^2$)} & \textit{eS} & 0-144 & \textit{vt} & 4-64 \\
 & \textit{rS} & 144-400 & \textit{t} & 64-256 \\
 & \textit{gS} & 400-1,024 & \textit{s} & 256-1,024 \\
 & \textit{N} & 1,024-2,000 & \textit{m} & 1,024-4,096 \\
\bottomrule
\end{tabular}}
\end{table}
}

\GeoGrainDeferFloat{mainresults}{%
\begin{table*}[t]
\caption{\textbf{Quantitative results for Generation quality and Layout fidelity.}
KID and SceneKID are scaled by $10^3$.
The best and second-best results are \textbf{bold}
and \GeoGrainUnderline{underlined}, respectively.}
\label{tab:main}
\GeoGrainTableFont{\scriptsize}
\centering
\setlength{\tabcolsep}{6pt}
\renewcommand{\arraystretch}{1.15}
\resizebox{1.0\textwidth}{!}{%
\begin{tabular}{llccccccccccc}
\toprule
\multirow{2}{*}{\textbf{Method}} & \multirow{2}{*}{\textbf{Venue}} & \multicolumn{4}{c}{\textbf{Generation quality} $\downarrow$} & \multicolumn{7}{c}{\textbf{Layout fidelity} $\uparrow$} \\
\cmidrule(lr){3-6}\cmidrule(l){7-13}
 & & FID & KID & SceneFID & SceneKID & AP & AP$_{50}$ & AP$_{75}$ & \multicolumn{4}{c}{AP by object size} \\
\midrule
\rowcolor{gray!5}\multicolumn{2}{l}{\textbf{SODA-A} Dataset} & & & & & & & & \textit{eS} & \textit{rS} & \textit{gS} & \textit{N} \\
Real images &  & -- & -- & -- & -- & 29.0 & 67.4 & 18.3 & 6.6 & 22.5 & 38.9 & 36.4 \\
AeroGen~\cite{aerogen2025} & CVPR'25 & 28.00 & 13.40 & 35.03 & 18.31 & 4.6 & 19.2 & 0.6 & 0.3 & 1.9 & 7.3 & 10.8 \\
CC-Diff~\cite{ccdiff2024} & arXiv'24 & 45.37 & 29.97 & 48.71 & 27.77 & 2.3 & 10.0 & 0.2 & 0.0 & 0.4 & 3.4 & 7.0 \\
CC-Diff++~\cite{ccdiffpp2025} & TGRS'25 & 48.14 & 31.29 & 61.66 & 39.39 & 1.6 & 7.5 & 0.2 & 0.0 & 0.3 & 2.3 & 4.7 \\
FICGen~\cite{ficgen2025} & ICCV'25 & 47.97 & 31.24 & 41.90 & 25.24 & 6.5 & 25.6 & 0.9 & 0.2 & 1.9 & 10.3 & 16.8 \\
SiamLayout~\cite{creatilayout2025} & ICCV'25 & 41.74 & 17.37 & 87.70 & 54.42 & 0.0 & 0.0 & 0.0 & 0.0 & 0.0 & 0.0 & 0.0 \\
Laytrol~\cite{laytrol2026} & AAAI'26 & 30.97 & 13.05 & 24.94 & 11.45 & 2.2 & 10.8 & 0.1 & 0.2 & 1.1 & 3.5 & 3.2 \\
TerraDiT-$\Omega$~\cite{terraditomega2026} & ECCV'26 & \GeoGrainUnderline{20.59} & \GeoGrainUnderline{9.95} & \GeoGrainUnderline{16.01} & \GeoGrainUnderline{8.37} & \GeoGrainUnderline{11.0} & \GeoGrainUnderline{34.8} & \GeoGrainUnderline{3.4} & \GeoGrainUnderline{0.8} & \GeoGrainUnderline{6.7} & \GeoGrainUnderline{18.1} & \GeoGrainUnderline{19.4} \\
\rowcolor{RoyalPurple!5} \textbf{GeoGrain (Ours)} & -- & \textbf{10.24} & \textbf{2.71} & \textbf{8.85} & \textbf{5.52} & \textbf{29.6} & \textbf{71.1} & \textbf{17.3} & \textbf{6.2} & \textbf{22.5} & \textbf{39.7} & \textbf{39.3} \\
\midrule
\rowcolor{gray!5}\multicolumn{2}{l}{\textbf{AI-TOD-R} Dataset} & & & & & & & & \textit{vt} & \textit{t} & \textit{s} & \textit{m} \\
Real images &  & -- & -- & -- & -- & 9.1 & 29.9 & 2.4 & 3.1 & 8.3 & 14.1 & 17.5 \\
AeroGen~\cite{aerogen2025} & CVPR'25 & 54.32 & 22.72 & 40.24 & 31.38 & 1.3 & 5.2 & 0.3 & 0.1 & 0.6 & 2.3 & 5.4 \\
CC-Diff~\cite{ccdiff2024} & arXiv'24 & 55.27 & 26.31 & 41.83 & 22.78 & 0.3 & 1.3 & 0.1 & 0.0 & 0.1 & 0.2 & 2.6 \\
CC-Diff++~\cite{ccdiffpp2025} & TGRS'25 & 64.36 & 34.89 & 42.70 & 25.45 & 0.2 & 1.0 & 0.0 & 0.0 & 0.0 & 0.2 & 2.2 \\
FICGen~\cite{ficgen2025} & ICCV'25 & 55.40 & 29.88 & 33.74 & 19.26 & 0.8 & 3.3 & 0.1 & 0.1 & 0.2 & 1.3 & 7.5 \\
SiamLayout~\cite{creatilayout2025} & ICCV'25 & 96.60 & 55.82 & 50.72 & 34.17 & 0.0 & 0.0 & 0.0 & 0.0 & 0.0 & 0.0 & 0.0 \\
Laytrol~\cite{laytrol2026} & AAAI'26 & 37.28 & 12.08 & 18.10 & \GeoGrainUnderline{10.35} & 0.3 & 1.7 & 0.0 & 0.1 & 0.2 & 0.5 & 1.6 \\
TerraDiT-$\Omega$~\cite{terraditomega2026} & ECCV'26 & \textbf{18.12} & \GeoGrainUnderline{6.76} & \GeoGrainUnderline{14.94} & 11.67 & \GeoGrainUnderline{2.7} & \GeoGrainUnderline{10.6} & \GeoGrainUnderline{0.6} & \GeoGrainUnderline{0.3} & \GeoGrainUnderline{1.7} & \GeoGrainUnderline{6.1} & \GeoGrainUnderline{11.5} \\
\rowcolor{RoyalPurple!5} \textbf{GeoGrain (Ours)} & -- & \GeoGrainUnderline{19.54} & \textbf{4.72} & \textbf{4.02} & \textbf{1.81} & \textbf{9.0} & \textbf{30.4} & \textbf{2.2} & \textbf{2.1} & \textbf{7.2} & \textbf{15.5} & \textbf{24.3} \\
\bottomrule
\end{tabular}}
\end{table*}
}

\subsection{Training and Inference}
\label{sec:training}

We train GAGE, the image--layout branch $\mathcal{B}_L$,
and the fusion projections.
For each clean image-token sequence $\mathbf{z}_0$, we sample
$\boldsymbol{\epsilon}\sim\mathcal{N}(\mathbf{0},\mathbf{I})$
and a single timestep $\tau\in[0,1]$ for both coarse and
fine tokens.
We then define the noisy sequence $\mathbf{z}_I$
and target velocity $\mathbf{v}$~\cite{flowmatching2023,rectifiedflow2023} as
\iflatexml\begin{minipage}{2\textwidth}\fi%
\begin{equation}
\mathbf{z}_I=\mathbf{z}_{\tau}
=(1-\tau)\mathbf{z}_0+\tau\boldsymbol{\epsilon},
\qquad
\mathbf{v}=\boldsymbol{\epsilon}-\mathbf{z}_0.
\end{equation}
\iflatexml\end{minipage}\fi%

Prior L2I methods set a maximum number of objects in the
input layout $L$ during
training~\cite{gligen2023,ficgen2025,ccdiff2024,migc2024},
reducing computational costs.
However, the denoising loss applies to image regions
occupied by objects not included in $L$,
potentially encouraging the generation of additional
objects not specified by $L$.
We therefore define a binary loss mask $\eta_j$ as $0$
when grid cell $C_j$ has positive-area overlap only with
annotated OBBs not included in $L$, and otherwise $1$.
Using $\eta_j$, we minimize the masked flow-matching (FM) loss:
\iflatexml\begin{minipage}{2\textwidth}\fi%
\begin{equation}
\mathcal{L}_{\mathrm{MFM}}
=
\mathbb{E}\!\left[
\frac{1}{N}
\sum\nolimits_{j=1}^{N}
\eta_j
\lVert
\hat{\mathbf{v}}_j-\mathbf{v}_j
\rVert_2^2
\right],
\end{equation}
\iflatexml\end{minipage}\fi%
where $N$ is the number of image tokens and the expectation
is over training samples, noise, and timesteps.

At inference, we initialize image tokens with Gaussian
noise on the GATE grid, denoise them from $\tau=1$ to $0$
over $K$ steps under the text and GAGE conditions,
and reconstruct $\hat{\mathbf{x}}$ as described in
Sec.~\ref{sec:gate}.

\section{Experiments}
\label{sec:experiment}
\subsection{Experimental Setup}
\label{sec:setup}
\noindent\textbf{Datasets.}
We use SODA-A~\cite{soda2023} and AI-TOD-R~\cite{aitodr2026},
two remote sensing datasets for oriented small-object detection.
All experiments use $512\times512$ crops from each dataset's
official training and test splits for image generation and
evaluation.
Table~\ref{tab:datasets} reports statistics of the cropped
datasets and the official object-size ranges used for evaluation.

\GeoGrainShowFloat{datasets}

\noindent\textbf{Baselines.}
We compare GeoGrain with the U-Net-based L2I methods
AeroGen~\cite{aerogen2025}, CC-Diff~\cite{ccdiff2024},
CC-Diff++~\cite{ccdiffpp2025}, and FICGen~\cite{ficgen2025}.
We also include the DiT-based methods
SiamLayout~\cite{creatilayout2025}, Laytrol~\cite{laytrol2026},
and TerraDiT-$\Omega$~\cite{terraditomega2026}.

\noindent\textbf{Implementation details.}
GeoGrain builds on FLUX.2 [klein] 4B Base~\cite{flux2klein2026}
and its Qwen3~\cite{qwen32025} text encoder.
We train the model for 40K iterations
with an effective batch size of 16, using AdamW~\cite{adamw2019}
at a learning rate of \mbox{5e-5} under a cosine schedule
with 80 warm-up steps.
Following prior work~\cite{ficgen2025}, we set the maximum
number of objects used for training to 15 per image.
At inference, we use 50 Euler sampling steps with a text
classifier-free guidance~\cite{cfg2022} scale of 4.0,
without limiting the number of objects.
Further details are provided in the \textit{Appendix}.

\GeoGrainDeferFloat{qualitative}{%
\begin{figure*}[t]
\centering
\iflatexml
\includegraphics[width=2\textwidth,alt={Qualitative comparison on AI-TOD-R using the same input layout. Oriented bounding boxes are overlaid on generated images, with enlarged regions for comparing small-object detail and alignment.}]{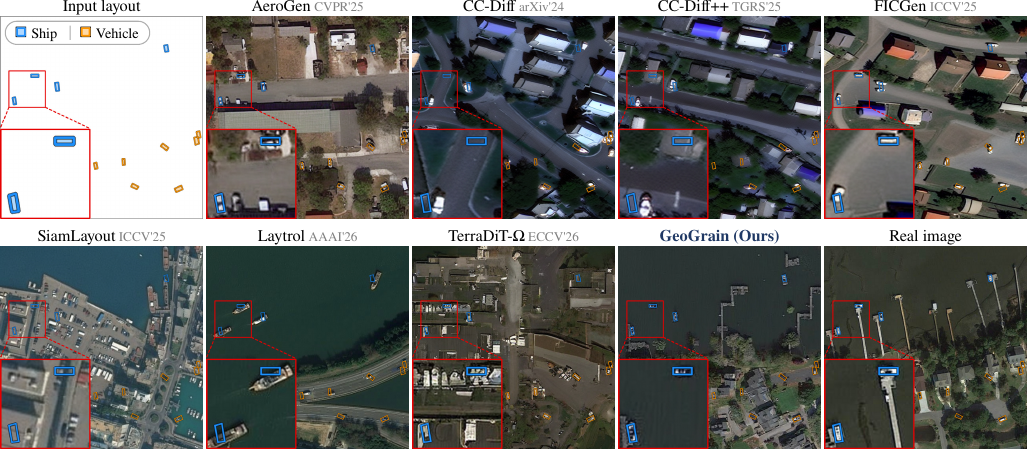}
\else
\includegraphics[width=\linewidth]{figures/visual.pdf}
\fi
\caption{\textbf{Qualitative comparison on the AI-TOD-R dataset.}
All methods use the same input layout.
Input OBBs are overlaid on the generated images,
and red-boxed regions are enlarged in the insets.}
\label{fig:qualitative}
\end{figure*}
}

\noindent\textbf{Evaluation metrics.}
(i) \textit{Generation quality:}
FID~\cite{fid2017} and KID~\cite{kid2018} assess global
image quality, while SceneFID~\cite{scenefid2021} and
SceneKID measure object-level quality using resized
object crops from real and generated images.
(ii) \textit{Layout fidelity:}
Images generated from test layouts are evaluated using
a fixed detector trained on real training images,
with input OBBs as annotations~\cite{geodiffusion2024}.
(iii) \textit{Trainability:}
Following~\cite{geodiffusion2024}, we augment layouts
from the training set with random horizontal flips and
translations before image generation.
Unless otherwise stated, we train detectors on all real
training images with an equal number of synthetic images
and evaluate them on real test images.
Both (ii) and (iii) use DCFL~\cite{dcfl2023},
designed for oriented tiny-object detection, with
Rotated RetinaNet~\cite{retinanet2017}.
All detectors follow the official AI-TOD-R
configuration~\cite{aitodr2026} using an MMRotate-based
implementation~\cite{mmrotate2022} with a standard $1\times$ schedule (12 epochs).
We report average precision (AP) under the official
evaluation protocol for each dataset, averaging over
classes and IoU thresholds from 0.50 to 0.95 in steps of 0.05.

\subsection{Quantitative Evaluation}

\noindent\textbf{Generation quality.}
Table~\ref{tab:main} compares the generation quality
of GeoGrain and the baselines.
GeoGrain achieves the best FID and KID on SODA-A
and the best KID on AI-TOD-R.
It also outperforms all baselines in the object-level
metrics SceneFID and SceneKID on both datasets.
Notably, its SceneFID on AI-TOD-R is 4.02,
compared with 14.94 for the strongest baseline,
TerraDiT-$\Omega$.
Together, these results suggest improved small-object
appearance alongside strong overall image quality.

\GeoGrainShowFloat{mainresults}

\noindent\textbf{Layout fidelity.}
GeoGrain substantially outperforms all baselines in layout
fidelity on both datasets, as summarized in Table~\ref{tab:main}.
It improves AP over the strongest baseline,
TerraDiT-$\Omega$, by 18.6 points on SODA-A and
6.3 points on AI-TOD-R, achieving AP comparable to
that measured on real images in both datasets.
The gains span all object-size ranges, including
the smallest objects.
At the stricter IoU threshold of 0.75, GeoGrain achieves an AP$_{75}$ of 17.3 on SODA-A,
compared with 3.4 for TerraDiT-$\Omega$,
indicating more precise object alignment with the input OBBs.

\GeoGrainDeferFloat{trainability}{%
\begin{table}[t]
\caption{\textbf{Quantitative results for Trainability.}
$\Delta$AP is relative to real-only training.
The best and second-best results are \textbf{bold}
and \GeoGrainUnderline{underlined}, respectively.}
\label{tab:trainability}
\GeoGrainTableFont{\scriptsize}
\centering
\setlength{\tabcolsep}{3pt}
\renewcommand{\arraystretch}{1.15}
\resizebox{1.0\columnwidth}{!}{%
\begin{tabular}{l cccc cccc}
\toprule
\multirow{2}{*}{\textbf{Method}} & \multicolumn{8}{c}{\textbf{Trainability} $\uparrow$} \\
\cmidrule(l){2-9}
 & AP & $\Delta$AP & AP$_{50}$ & AP$_{75}$ & \multicolumn{4}{c}{AP by object size} \\
\midrule
\rowcolor{gray!5}\multicolumn{2}{l}{\textbf{SODA-A} Dataset} & & & & \textit{eS} & \textit{rS} & \textit{gS} & \textit{N} \\
Real only & 29.0 & -- & 67.4 & 18.3 & 6.6 & 22.5 & 38.9 & 36.4 \\
AeroGen~\cite{aerogen2025} & \GeoGrainUnderline{29.2} & \GeoGrainUnderline{+0.2} & \GeoGrainUnderline{67.5} & \GeoGrainUnderline{18.8} & \GeoGrainUnderline{6.5} & \GeoGrainUnderline{22.6} & 39.4 & \GeoGrainUnderline{37.1} \\
CC-Diff~\cite{ccdiff2024} & 28.4 & \textminus{}0.6 & 66.4 & 17.6 & 6.2 & 21.7 & 38.0 & 35.7 \\
CC-Diff++~\cite{ccdiffpp2025} & 28.4 & \textminus{}0.6 & 66.8 & 17.3 & 6.1 & 21.9 & 38.4 & 36.6 \\
FICGen~\cite{ficgen2025} & 29.1 & +0.1 & 67.4 & 18.4 & 6.4 & 22.4 & \GeoGrainUnderline{39.9} & 36.8 \\
SiamLayout~\cite{creatilayout2025} & 27.9 & \textminus{}1.1 & 65.9 & 17.0 & 6.0 & 21.4 & 36.5 & 35.0 \\
Laytrol~\cite{laytrol2026} & 28.3 & \textminus{}0.7 & 66.5 & 17.2 & 6.3 & 21.5 & 38.1 & 36.6 \\
TerraDiT-$\Omega$~\cite{terraditomega2026} & 28.5 & \textminus{}0.5 & 66.5 & 18.0 & 6.0 & 21.7 & 38.1 & 36.0 \\
\rowcolor{RoyalPurple!5} \textbf{GeoGrain (Ours)} & \textbf{30.8} & \textbf{+1.8} & \textbf{69.4} & \textbf{20.9} & \textbf{7.4} & \textbf{23.8} & \textbf{40.5} & \textbf{37.8} \\
\midrule
\rowcolor{gray!5}\multicolumn{2}{l}{\textbf{AI-TOD-R} Dataset} & & & & \textit{vt} & \textit{t} & \textit{s} & \textit{m} \\
Real only & 9.1 & -- & 29.9 & 2.4 & 3.1 & 8.3 & 14.1 & 17.5 \\
AeroGen~\cite{aerogen2025} & 9.3 & +0.2 & \GeoGrainUnderline{31.9} & 2.3 & 2.8 & 8.2 & 13.5 & 19.3 \\
CC-Diff~\cite{ccdiff2024} & 8.5 & \textminus{}0.6 & 28.7 & 2.4 & 2.6 & 7.8 & 12.3 & 18.8 \\
CC-Diff++~\cite{ccdiffpp2025} & 8.8 & \textminus{}0.3 & 28.6 & \GeoGrainUnderline{2.6} & 2.5 & 7.9 & 12.5 & 18.5 \\
FICGen~\cite{ficgen2025} & 8.9 & \textminus{}0.2 & 29.4 & 2.5 & 2.9 & 8.2 & 12.7 & \GeoGrainUnderline{20.1} \\
SiamLayout~\cite{creatilayout2025} & 6.7 & \textminus{}2.4 & 22.5 & 1.7 & 2.5 & 7.1 & 9.5 & 16.1 \\
Laytrol~\cite{laytrol2026} & 8.8 & \textminus{}0.3 & 29.9 & 2.2 & 3.0 & 8.0 & 12.4 & 16.4 \\
TerraDiT-$\Omega$~\cite{terraditomega2026} & \GeoGrainUnderline{9.5} & \GeoGrainUnderline{+0.4} & 31.0 & 2.4 & \GeoGrainUnderline{3.4} & \GeoGrainUnderline{8.4} & \GeoGrainUnderline{14.2} & 19.4 \\
\rowcolor{RoyalPurple!5} \textbf{GeoGrain (Ours)} & \textbf{11.2} & \textbf{+2.1} & \textbf{35.8} & \textbf{3.3} & \textbf{4.2} & \textbf{10.2} & \textbf{16.1} & \textbf{21.7} \\
\bottomrule
\end{tabular}}
\end{table}
}

\noindent\textbf{Trainability.}
The trainability results reported in
Table~\ref{tab:trainability} show that using images
generated by GeoGrain for data augmentation improves
detector AP over real-only training by 1.8 points on
SODA-A and 2.1 points on AI-TOD-R.
Under the same setting, images from competing generators
yield AP gains of at most 0.2 and 0.4 points, respectively.
Detectors trained with images generated by GeoGrain
also achieve the highest AP in every object-size range
and the highest AP$_{75}$ on real test images from both datasets,
demonstrating the effectiveness of GeoGrain for data
augmentation in downstream oriented small-object detection.

\GeoGrainShowFloat{trainability}

\subsection{Qualitative Evaluation}
\label{sec:qualitative}

On SODA-A, GeoGrain generates distinct, recognizable
small vehicles whose positions and orientations closely
match the input OBBs, as illustrated by the enlarged views
in the left panel of Fig.~\ref{fig:teaser}.
In comparison, the remote sensing methods
TerraDiT-$\Omega$, AeroGen, CC-Diff++, and FICGen
omit some small vehicles specified by the layout.
This comparison highlights the ability of GeoGrain to
control individual small objects even when they are
closely spaced.

Fig.~\ref{fig:qualitative} presents the qualitative
comparison on \mbox{AI-TOD-R}.
GeoGrain generates recognizable small ships that more
closely match the prescribed positions, sizes, and
orientations than those produced by competing methods.
The generated ships also fit naturally within the
surrounding waterfront scene.
In contrast, although SiamLayout and Laytrol adopt a similar
dual-branch design and generate in-domain scenes, they do not
place objects precisely at the input OBBs, as also reflected by
their near-zero AP in Table~\ref{tab:main}.

\GeoGrainShowFloat{qualitative}

\subsection{Ablation Studies}
\label{sec:ablations}

\GeoGrainDeferFloat{ablationmain}{%
\begin{table}[t]
\caption{\textbf{Ablation results on GAGE and GATE.}
The best and second-best results are \textbf{bold}
and \GeoGrainUnderline{underlined}, respectively.}
\label{tab:ablation_main}
\GeoGrainTableFont{\scriptsize}
\centering
\setlength{\tabcolsep}{7pt}
\renewcommand{\arraystretch}{1.15}
\resizebox{1.0\columnwidth}{!}{%
\begin{tabular}{cccc ccc}
\toprule
\multicolumn{3}{c}{\textbf{GAGE}} & \multirow{2}{*}{\textbf{GATE}} & \multicolumn{3}{c}{\textbf{Layout fidelity} $\uparrow$} \\
\cmidrule(r){1-3}\cmidrule(l){5-7}

RoPE ($\boldsymbol{\rho}_L^{ij}$) & $\mathbf{Q}_{ij}$ & $\mathbf{R}_{ij}$ & & AP & AP$_{50}$ & AP$_{75}$ \\
\midrule
\ding{55} & \ding{55} & \ding{55} & \ding{55} & 0.5 & 2.4 & 0.0 \\
\ding{51} & \ding{55} & \ding{55} & \ding{55} & 2.9 & 13.8 & 0.2 \\
\ding{51} & \ding{51} & \ding{55} & \ding{55} & 18.9 & 54.2 & 7.3 \\
\ding{51} & \ding{55} & \ding{51} & \ding{55} & 18.9 & 54.3 & 7.4 \\
\ding{51} & \ding{51} & \ding{51} & \ding{55} & \GeoGrainUnderline{21.7} & \GeoGrainUnderline{59.5} & \GeoGrainUnderline{9.0} \\
\ding{51} & \ding{55} & \ding{55} & \ding{51} & 14.1 & 47.4 & 3.3 \\
\rowcolor{RoyalPurple!5} \ding{51} & \ding{51} & \ding{51} & \ding{51} & \textbf{29.6} & \textbf{71.1} & \textbf{17.3} \\
\bottomrule
\end{tabular}}
\end{table}
}

\GeoGrainDeferFloat{ablationmask}{%
\begin{table}[t]
\caption{\textbf{Ablation results on Masked FM loss.}}
\label{tab:ablation_maskedloss}
\GeoGrainTableFont{\scriptsize}
\centering
\setlength{\tabcolsep}{5pt}
\renewcommand{\arraystretch}{1.15}
\resizebox{1.0\columnwidth}{!}{%
\begin{tabular}{c ccc ccc}
\toprule
\multirow{2}{*}{\makecell{\textbf{Masked FM loss}\\ $\mathcal{L}_{\mathrm{MFM}}$}} & \multicolumn{3}{c}{\textbf{Layout fidelity} $\uparrow$} & \multicolumn{3}{c}{\textbf{Trainability} $\uparrow$} \\
\cmidrule(lr){2-4}\cmidrule(l){5-7}
 & AP & AP$_{50}$ & AP$_{75}$ & AP & AP$_{50}$ & AP$_{75}$ \\
\midrule
\ding{55} & 24.8 & 59.9 & 14.1 & 30.3 & 68.6 & 20.1 \\
\rowcolor{RoyalPurple!5} \ding{51} & \textbf{29.6} & \textbf{71.1} & \textbf{17.3} & \textbf{30.8} & \textbf{69.4} & \textbf{20.9} \\
\bottomrule
\end{tabular}}
\vspace{-2mm}

\end{table}
}

We conduct ablations on SODA-A to examine the contributions
of individual components in GeoGrain and the effect of
synthetic data quantity.

\noindent\textbf{GAGE and GATE.}
As shown in Table~\ref{tab:ablation_main}, the dual-branch
baseline without GAGE and GATE achieves an AP of only 0.5,
similar to the 0.0 and 2.2 of the dual-branch methods SiamLayout and Laytrol in Table~\ref{tab:main}.
This indicates that a dual-branch design alone is insufficient for precise small-object control.
Sharing RoPE position indices between associated layout
and image tokens, i.e., $\boldsymbol{\rho}_L^{ij}=\boldsymbol{\rho}_I^j$,
modestly raises AP.
With this alignment, adding either local descriptor
$\mathbf{Q}_{ij}$ or $\mathbf{R}_{ij}$ substantially
improves AP.
Using both descriptors further improves AP over either
alone, suggesting that they provide complementary
geometric information.
GATE also improves AP without the local descriptors,
supporting the benefit of finer image tokens in object regions.
Adding GATE increases AP$_{75}$ by 3.1 points without
the local descriptors and by 8.3 points with both.
This larger gain under the stricter localization criterion
suggests that combining local geometric guidance with
finer image tokens is particularly effective for precise
OBB alignment.

\GeoGrainShowFloat{ablationmain}

\begin{figure}[t]
\centering
\iflatexml
\includegraphics[width=2\textwidth,alt={Effect of Masked FM loss. Green boxes indicate objects specified in the input layout; red dashed boxes highlight additional generated objects.}]{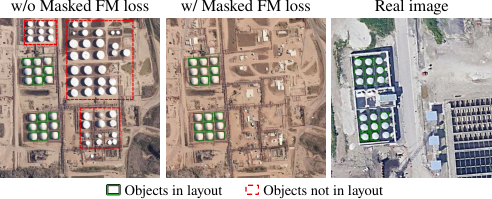}
\else
\includegraphics{figures/maskedloss.pdf}
\fi
\caption{\textbf{Qualitative results on Masked FM loss.}
Green boxes mark objects specified by the input layout,
while red dashed boxes highlight additional objects
generated without $\mathcal{L}_{\mathrm{MFM}}$.}
\label{fig:masked_loss}
\vspace{-2mm}
\end{figure}

\noindent\textbf{Masked FM loss.}
Table~\ref{tab:ablation_maskedloss} shows that using
$\mathcal{L}_{\mathrm{MFM}}$ improves AP by 4.8 points
for layout fidelity and 0.5 points for trainability.
The qualitative comparison in Fig.~\ref{fig:masked_loss}
suggests a possible explanation for these gains:
without loss masking, the model generates additional
storage tanks not specified by the input layout,
whereas the masked FM loss helps avoid generating
these extra objects.
Such unannotated objects can reduce layout fidelity
and may introduce label noise when treated as background
during detector training.

\GeoGrainShowFloat{ablationmask}

\GeoGrainDeferFloat{augmentation}{%
\begin{table}[t]
\caption{\textbf{Ablation results on Synthetic data quantity.}
$\Delta$AP is relative to real-only training.
The best and second-best results are \textbf{bold}
and \GeoGrainUnderline{underlined}, respectively.}
\label{tab:aug_amount}
\GeoGrainTableFont{\scriptsize}
\centering
\setlength{\tabcolsep}{5pt}
\renewcommand{\arraystretch}{1.15}
\resizebox{1.0\columnwidth}{!}{%
\begin{tabular}{lccccc}
\toprule
\multirow{2}{*}{\textbf{Training data}} &
\multirow{2}{*}{\textbf{\# Images}} &
\multicolumn{4}{c}{\textbf{Trainability} $\uparrow$} \\
\cmidrule(l){3-6}
 & & AP & $\Delta$AP & AP$_{50}$ & AP$_{75}$ \\
\midrule
\rowcolor{gray!5} Real 100\% & 23,032 & 29.0 & -- & 67.4 & 18.3 \\
\, + Synthetic 25\% & +5,758 & 29.2 & +0.2 & 67.8 & 18.3 \\
\, + Synthetic 50\% & +11,516 & 29.7 & +0.7 & 68.3 & 19.1 \\
\, + Synthetic 75\% & +17,274 & 30.1 & +1.1 & 69.1 & 19.3 \\
\, + Synthetic 100\% & +23,032 & \GeoGrainUnderline{30.8} & \GeoGrainUnderline{+1.8} & \GeoGrainUnderline{69.4} & \GeoGrainUnderline{20.9} \\
\, + Synthetic 200\% & +46,064 & \textbf{31.5} & \textbf{+2.5} & \textbf{69.6} & \textbf{22.3} \\
\bottomrule
\end{tabular}}
\end{table}
}

\noindent\textbf{Synthetic data quantity.}
To assess how synthetic data quantity affects the
effectiveness of data augmentation, we add different
numbers of images generated by GeoGrain to the full
real training set, keeping the $1\times$ schedule.
Table~\ref{tab:aug_amount} shows that adding 100\%
synthetic data improves AP by 1.8 points over real-only
training.
With 200\% synthetic data, the AP gain increases to
2.5 points, and AP$_{75}$ improves by 4.0 points.
These results indicate that using more synthetic images
can improve detection on real test
images, even when more precise localization is required.

\GeoGrainShowFloat{augmentation}
\section{Conclusion}
\label{sec:conclusion}

We presented GeoGrain, a granularity-aware L2I framework
for precise small-object control in remote sensing imagery.
GAGE encodes instance-specific local OBB geometry,
while GATE subdivides image-token grid cells that
overlap the input OBBs.
On the resulting mixed-resolution grid, GAGE constructs
layout tokens aligned with their corresponding image tokens.
Experiments on SODA-A and AI-TOD-R demonstrate substantial
gains in layout fidelity and improved small-object appearance
alongside strong overall image quality.
Data augmentation with our generated images also improves
oriented small-object detection, outperforming all baselines. 

\clearpage

\iflatexml\else\ifGeoGrainReview\else\nolinenumbers\fi\fi
\clearpage
\iflatexml\else
\makeatletter
\global\let\@addtocurcol\GG@addtocurcol@flafter
\global\@dblfpsep\dblfloatsep
\makeatother
\fi
\hyphenation{Geo-Grain FICGen}
\crefname{appendix}{Sec.}{Secs.}
\Crefname{appendix}{Section}{Sections}
\appendix
\newpage
\iflatexml
\part*{Appendix}
\else
\twocolumn[
 \centering
 \Large
 \textbf{\thetitle}\\
 \vspace{0.5em}Appendix \\
 \vspace{1.0em}
]
\fi

\iflatexml\else\setlength{\stripsep}{14.5pt plus 1pt minus 1pt}\fi
\iflatexml\begin{table*}\else\begin{strip}\fi
\vspace*{-35.5pt}
\centering
\captionof{table}{\textbf{Overview of the appendix.}}
\label{tab:supp_overview}
\vspace{8.5pt}
\GeoGrainTableFont{\fontsize{8.0}{9.5}\selectfont}
\setlength{\tabcolsep}{6pt}
\renewcommand{\arraystretch}{1.10}
\begin{tabularx}{\textwidth}{>{\raggedright\arraybackslash}p{0.30\textwidth}Y}
\toprule
\rowcolor{RoyalPurple!5}
\textbf{Section} & \textbf{Contents} \\
\midrule
\textbf{\ref{supp:datasets}. Datasets} &
Dataset Descriptions and Preprocessing. \\
\addlinespace[0.9pt]
\textbf{\ref{supp:implementation}. Implementation Details} &
\ref{supp:geograin_impl}~GeoGrain;
\ref{supp:baselines}~Baselines. \\
\addlinespace[0.9pt]
\textbf{\ref{supp:quantitative}. Additional Quantitative Results} &
\ref{supp:detectors}~Detector and Backbone Generalization;
\ref{supp:perclass}~Class-wise Evaluation;\newline
\ref{supp:dior}~Evaluation on DIOR-R;
\ref{supp:runtime}~Inference Efficiency. \\
\addlinespace[0.9pt]
\textbf{\ref{supp:ablations}. Additional Ablation Studies} &
\ref{supp:architecture}~Dual-branch Architecture Compatibility;
\ref{supp:granularity}~Token Expansion;\newline
\ref{supp:position}~Mixed-resolution Positional Encoding;
\ref{supp:decoding}~Mixed-resolution Decoding;\newline
\ref{supp:synthetic}--\ref{supp:updates}~Trainability at Fixed Image Count or Iteration Count. \\
\addlinespace[0.9pt]
\textbf{\ref{supp:qualitative}. Additional Qualitative Results} &
Visual Comparisons across Datasets. \\
\addlinespace[0.9pt]
\textbf{\ref{supp:limitations}. Limitations and Future Work} &
Remaining Challenges and Future Directions. \\
\bottomrule
\end{tabularx}
\iflatexml\end{table*}\else\end{strip}\fi

\noindent
This appendix provides dataset and implementation details, additional quantitative and qualitative analyses, and ablation studies. We also discuss limitations and future directions.
\Cref{tab:supp_overview} summarizes the contents.

\section{Datasets}
\label{sec:supp_data}
\label{supp:datasets}

We evaluate GeoGrain on two small-object detection benchmarks, SODA-A~\cite{soda2023} and AI-TOD-R~\cite{aitodr2026}, and additionally on DIOR-R~\cite{cheng2022diorr}, a general remote sensing object detection benchmark. All experiments use $512\times512$ images with OBB annotations derived from the official training and test splits. These processed splits are used consistently for image generation and evaluation. We report the mean object size in the processed images, measured as $\sqrt{wh}$, where $w$ and $h$ denote the OBB side lengths.

\noindent\textbf{SODA-A.} SODA-A is a large-scale benchmark for oriented small-object detection in aerial images. It consists of 2,513 high-resolution Google Earth images with 872,069 OBB annotations across nine categories, divided into 1,067 training, 576 validation, and 870 test images. We divide each training and test image into $512\times512$ crops at its original resolution and assign an object to a crop only if at least 70\% of its area lies within it. Crops that contain no objects or that overlap regions marked as \textit{ignore} in the annotations are discarded. The resulting training and test sets contain 23,032 and 15,989 crops, with 347,884 and 300,158 objects, respectively. The mean object size is 16.2~px.\looseness=1

\noindent\textbf{AI-TOD-R.} AI-TOD-R is dedicated to oriented tiny-object detection. It consists of 28,036 images of $800\times800$~px collected from five remote sensing datasets~\cite{wang2021aitod}: xView~\cite{lam2018xview}, Airbus Ship~\cite{airbus2018}, DOTA-v1.5~\cite{dota2018}, DIOR~\cite{dior2020}, and VisDrone2018-Det~\cite{zhu2022visdrone}. The dataset contains 752,460 OBB annotations across eight categories and is divided into 11,214 training, 2,804 validation, and 14,018 test images. We extract a single $512\times512$ crop from each training and test image, centered on the valid image region at the original resolution and excluding black padding borders. We apply the same 70\% object-assignment rule as for SODA-A and discard crops that contain no objects or that include invalid image regions. The resulting training and test sets contain 7,752 and 9,705 images, with 131,274 and 154,083 objects, respectively. The mean object size is 10.8~px.

\noindent\textbf{DIOR-R.} To evaluate GeoGrain beyond datasets focused on small objects, we additionally use DIOR-R, a general remote sensing object detection benchmark. It contains 23,463 Google Earth images of $800\times800$~px with 192,518 OBB annotations across 20 categories. The dataset is split into 5,862 training, 5,863 validation, and 11,738 test images. Following~\cite{aerogen2025}, we resize all training and test images to $512\times512$. The resulting training and test sets contain 5,862 and 11,738 images, with 32,632 and 124,397 objects, respectively. The mean object size is 35.0~px.

\section{Implementation Details}
\label{supp:implementation}

\subsection{GeoGrain}
\label{supp:geograin_impl}
\label{sec:supp_impl}

\noindent\textbf{Model.}
GeoGrain is built on FLUX.2 [klein] 4B Base~\cite{flux2klein2026}. Its Qwen3~\cite{qwen32025} text encoder maps the global prompt $T$ to text tokens for the image--text branch $\mathcal{B}_T$. GAGE reuses this encoder to embed object-class prompts, followed by a small MLP, while separate MLPs process Fourier-encoded~\cite{fourier2020} OBB geometry. We freeze the pretrained image--text branch, text encoder, and VAE~\cite{vae2014,ldm2022}, and train the GAGE modules, image--layout branch $\mathcal{B}_L$, and fusion projections. We initialize $\mathcal{B}_L$ from pretrained transformer blocks and set the fusion projection weights and biases to zero~\cite{laytrol2026,controlnet2023}. GATE assigns RoPE~\cite{rope2024} position indices defined on the fine-token grid, with each coarse token taking the position of the top-left fine cell in its region.

\noindent\textbf{Text prompts.}
For SODA-A and DIOR-R, the global prompt $T$ lists the object classes present in the layout (\eg, ``An aerial image with airplanes, small vehicles.''). For AI-TOD-R, we replace the generic ``aerial image'' description with the imagery type and source name to reflect differences in image appearance across sources: ``satellite'' for xView and Airbus, ``Google Earth'' for DOTA and DIOR, and ``drone-view'' for VisDrone. We prepend ``dark'' when the per-image mean luminance is below 45 and ``bright'' when it exceeds 77 on a 0--255 scale; otherwise, no brightness descriptor is added. For xView images, we additionally use ``grainy'' to indicate sensor noise (\eg, ``A dark grainy satellite image from xView with vehicles.''). Across all datasets, the local prompt consists of the corresponding object class name (\eg, ``small vehicle'').

\noindent\textbf{Training.}
We train GeoGrain for 40K iterations with an effective batch size of 16, using AdamW~\cite{adamw2019} with a weight decay of 0.01 and gradient clipping at 1.0. Training is performed with bfloat16 mixed precision on two NVIDIA B200 GPUs. The learning rate is linearly warmed up to \mbox{5e-5} over 80 iterations, followed by cosine decay~\cite{loshchilov2017sgdr}. Following prior work~\cite{ficgen2025}, we limit the number of objects per training image to 15. We apply text-conditioning dropout with a probability of 0.1 by replacing the global prompt with an empty prompt while retaining the layout condition, encouraging the model to rely on layout information~\cite{laytrol2026}. We sample a single timestep uniformly for both coarse and fine tokens and apply the resolution-dependent timestep shift~\cite{sd32024} used in FLUX.2, computed from the mixed-resolution token sequence length. To construct the fine grid for GATE, we encode training images after $2\times$ Lanczos upsampling~\cite{duchon1979lanczos}, without requiring higher-resolution training data.

\noindent\textbf{Inference.}
We use 50 Euler sampling steps with a text classifier-free guidance~\cite{cfg2022} scale of 4.0. For the unconditional prediction, we use an empty global prompt while retaining the layout condition. Unlike training, we do not impose an object-count limit at inference. GATE is applied to all objects in SODA-A and AI-TOD-R. For DIOR-R, which contains many large objects, we apply GATE only to coarse cells overlapping small objects during both training and inference to reduce computational cost. An object is considered small if its minimum-area OBB has side lengths $w$ and $h$ satisfying $wh\leq32^2$~px$^2$, following the MS COCO area threshold~\cite{lin2014microsoft} adopted by SODA-A~\cite{soda2023}. This criterion is applied consistently during training and inference. We use Lanczos interpolation for both $2\times$ upsampling and downsampling, followed by five-level Laplacian-pyramid blending~\cite{laplacian1983} to combine the fine and coarse reconstructions.

\subsection{Baselines}
\label{sec:supp_baselines}
\label{supp:baselines}

We compare GeoGrain with four U-Net-based methods (AeroGen~\cite{aerogen2025}, CC-Diff~\cite{ccdiff2024}, CC-Diff++~\cite{ccdiffpp2025}, and FICGen~\cite{ficgen2025}) and three DiT-based~\cite{dit2023} methods (Siam\-Layout~\cite{creatilayout2025}, Laytrol~\cite{laytrol2026}, and TerraDiT-$\Omega$~\cite{terraditomega2026}). We follow their official implementations except for the adaptations described below.\looseness=-1

\noindent\textbf{Backbones.}
The U-Net-based methods retain the backbones of their official implementations. Among the DiT-based methods, SiamLayout and Laytrol adopt a dual-branch design similar to GeoGrain, with a dedicated branch for layout conditioning. To control for backbone differences, we implement both methods using the same FLUX.2 [klein] 4B Base backbone as GeoGrain and use the same number of sampling steps and guidance scale, applying layout conditioning at every step. For TerraDiT-$\Omega$, we fine-tune its released pretrained model on each dataset at $512\times512$.

\noindent\textbf{Layout conditioning.}
AeroGen, CC-Diff, and CC-Diff++ natively support OBBs, whereas FICGen, SiamLayout, and Laytrol originally use HBB conditioning and are adapted to support OBBs. Following AeroGen~\cite{aerogen2025}, we represent each OBB using its four vertices. For the adapted methods, we extend the box embeddings from four HBB coordinates to eight OBB vertex coordinates, using the same vertex ordering and normalization rules as GeoGrain during both training and inference. Thus, GeoGrain and the three adapted methods use identical OBB geometry. TerraDiT-$\Omega$ defines its bounding-box primitive as axis-aligned, so we provide each OBB as a four-vertex polygon primitive.

\noindent\textbf{Text prompts.}
For AeroGen, CC-Diff, CC-Diff++, and FICGen, we construct text prompts following their official procedures for remote sensing images. SiamLayout and Laytrol are designed for natural images, whereas \mbox{TerraDiT-$\Omega$} uses captions from Git-10M~\cite{liu2025text2earth} without specifying how to construct prompts for other datasets. We therefore use GeoGrain's global and local prompts (Sec.~\ref{sec:supp_impl}) for all three methods.

\noindent\textbf{Training and evaluation.}
We train each baseline separately on each dataset using the same training split as GeoGrain and evaluate all methods on the corresponding test split. We match the training budget across all methods, including GeoGrain, so that each processes 640K training samples in total.
\GeoGrainDeferFloat{detectorfidelity}{\begin{table*}[t]
\centering
\caption{\textbf{Quantitative results for Layout fidelity across detectors and backbones.} The best and second-best results are \textbf{bold} and \GeoGrainUnderline{underlined}, respectively.}
\label{tab:supp_fidelity_detectors}
\label{tab:supp_detectors}
\GeoGrainTableFont{\fontsize{8.0}{9.0}\selectfont}
\setlength{\tabcolsep}{2.6pt}
\renewcommand{\arraystretch}{1.10}
\begin{tabularx}{\textwidth}{l*{14}{Z}}
\toprule
\multirow{3}{*}{\textbf{Method}} & \multicolumn{7}{c}{\textbf{SODA-A}} & \multicolumn{7}{c}{\textbf{AI-TOD-R}} \\
\cmidrule(lr){2-8}\cmidrule(l){9-15}
 & \multirow{2}{*}{AP} & \multirow{2}{*}{AP$_{50}$} & \multirow{2}{*}{AP$_{75}$} & \multicolumn{4}{c}{AP by object size} & \multirow{2}{*}{AP} & \multirow{2}{*}{AP$_{50}$} & \multirow{2}{*}{AP$_{75}$} & \multicolumn{4}{c}{AP by object size} \\
\cmidrule(lr){5-8}\cmidrule(l){12-15}
 &  &  &  & \textit{eS} & \textit{rS} & \textit{gS} & \textit{N} &  &  &  & \textit{vt} & \textit{t} & \textit{s} & \textit{m} \\
\midrule
\rowcolor{gray!5}\multicolumn{15}{l}{\textbf{DCFL w/ Oriented R-CNN (ResNet-50)}} \\
Real images & 35.2 & 75.2 & 26.7 & 11.4 & 28.6 & 43.9 & 40.1 & 12.5 & 40.7 & 3.4 & 4.6 & 12.1 & 16.7 & 21.3 \\
AeroGen~\cite{aerogen2025} & 4.9 & 19.7 & 0.8 & 0.4 & 1.9 & 7.7 & 11.8 & 1.5 & 6.2 & 0.3 & 0.2 & 0.6 & 2.7 & 7.3 \\
CC-Diff~\cite{ccdiff2024} & 2.2 & 9.8 & 0.2 & 0.1 & 0.3 & 3.4 & 7.0 & 0.3 & 1.3 & 0.0 & 0.0 & 0.0 & 0.2 & 3.3 \\
CC-Diff++~\cite{ccdiffpp2025} & 1.5 & 7.0 & 0.1 & 0.1 & 0.3 & 2.1 & 5.0 & 0.2 & 0.9 & 0.0 & 0.0 & 0.0 & 0.1 & 2.3 \\
FICGen~\cite{ficgen2025} & 6.7 & 25.8 & 1.0 & 0.2 & 1.6 & 10.9 & 18.8 & 0.6 & 2.8 & 0.1 & 0.0 & 0.1 & 0.8 & 9.7 \\
SiamLayout~\cite{creatilayout2025} & 0.0 & 0.0 & 0.0 & 0.0 & 0.0 & 0.0 & 0.0 & 0.0 & 0.0 & 0.0 & 0.0 & 0.0 & 0.0 & 0.0 \\
Laytrol~\cite{laytrol2026} & 2.2 & 10.2 & 0.4 & 0.3 & 1.1 & 3.3 & 3.1 & 0.4 & 1.6 & 0.1 & 0.2 & 0.2 & 0.4 & 1.9 \\
TerraDiT-$\Omega$~\cite{terraditomega2026} & \GeoGrainUnderline{12.2} & \GeoGrainUnderline{36.8} & \GeoGrainUnderline{4.1} & \GeoGrainUnderline{0.9} & \GeoGrainUnderline{7.3} & \GeoGrainUnderline{19.4} & \GeoGrainUnderline{19.5} & \GeoGrainUnderline{2.8} & \GeoGrainUnderline{11.4} & \GeoGrainUnderline{0.5} & \GeoGrainUnderline{0.3} & \GeoGrainUnderline{2.0} & \GeoGrainUnderline{6.6} & \GeoGrainUnderline{11.7} \\
\rowcolor{RoyalPurple!5} \textbf{GeoGrain (Ours)} & \textbf{34.1} & \textbf{76.3} & \textbf{23.1} & \textbf{9.8} & \textbf{27.1} & \textbf{42.8} & \textbf{42.7} & \textbf{11.3} & \textbf{39.1} & \textbf{2.1} & \textbf{2.7} & \textbf{10.0} & \textbf{17.9} & \textbf{26.1} \\
\midrule
\rowcolor{gray!5}\multicolumn{15}{l}{\textbf{Oriented R-CNN (ResNet-50)}} \\
Real images & 33.7 & 69.2 & 27.6 & 6.8 & 27.3 & 44.6 & 41.2 & 10.5 & 32.4 & 3.2 & 0.4 & 9.1 & 18.4 & 22.3 \\
AeroGen~\cite{aerogen2025} & 4.9 & 19.4 & 0.8 & 0.4 & 2.0 & 7.7 & 11.5 & 1.9 & 7.4 & 0.2 & \GeoGrainUnderline{0.1} & 0.8 & 3.2 & 7.8 \\
CC-Diff~\cite{ccdiff2024} & 2.2 & 9.3 & 0.2 & 0.1 & 0.3 & 3.3 & 6.7 & 0.3 & 1.2 & 0.1 & 0.0 & 0.0 & 0.3 & 3.3 \\
CC-Diff++~\cite{ccdiffpp2025} & 1.5 & 6.9 & 0.1 & 0.1 & 0.3 & 2.2 & 4.8 & 0.2 & 1.0 & 0.0 & 0.0 & 0.0 & 0.1 & 2.2 \\
FICGen~\cite{ficgen2025} & 6.5 & 25.4 & 1.0 & 0.2 & 1.7 & 10.4 & 18.5 & 0.7 & 3.2 & 0.1 & 0.0 & 0.2 & 1.1 & 9.6 \\
SiamLayout~\cite{creatilayout2025} & 0.0 & 0.1 & 0.0 & 0.0 & 0.0 & 0.0 & 0.0 & 0.0 & 0.0 & 0.0 & 0.0 & 0.0 & 0.0 & 0.0 \\
Laytrol~\cite{laytrol2026} & 2.3 & 10.6 & 0.5 & 0.4 & 1.1 & 3.7 & 3.1 & 0.4 & 1.8 & 0.0 & \GeoGrainUnderline{0.1} & 0.2 & 0.4 & 1.7 \\
TerraDiT-$\Omega$~\cite{terraditomega2026} & \GeoGrainUnderline{11.8} & \GeoGrainUnderline{35.0} & \GeoGrainUnderline{4.2} & \GeoGrainUnderline{0.8} & \GeoGrainUnderline{6.8} & \GeoGrainUnderline{19.2} & \GeoGrainUnderline{20.1} & \GeoGrainUnderline{3.2} & \GeoGrainUnderline{12.0} & \GeoGrainUnderline{0.6} & \GeoGrainUnderline{0.1} & \GeoGrainUnderline{1.8} & \GeoGrainUnderline{7.2} & \GeoGrainUnderline{13.1} \\
\rowcolor{RoyalPurple!5} \textbf{GeoGrain (Ours)} & \textbf{33.1} & \textbf{71.1} & \textbf{24.3} & \textbf{6.2} & \textbf{26.2} & \textbf{44.0} & \textbf{43.6} & \textbf{10.7} & \textbf{33.4} & \textbf{2.7} & \textbf{0.3} & \textbf{8.4} & \textbf{19.8} & \textbf{28.6} \\
\midrule
\rowcolor{gray!5}\multicolumn{15}{l}{\textbf{Oriented R-CNN (Swin-T)}} \\
Real images & 33.8 & 70.0 & 27.3 & 8.1 & 28.0 & 46.2 & 42.2 & 10.5 & 33.0 & 3.0 & 0.6 & 8.8 & 18.7 & 23.3 \\
AeroGen~\cite{aerogen2025} & 4.9 & 20.3 & 0.7 & 0.4 & 2.0 & 7.8 & 11.8 & 1.8 & 7.3 & 0.3 & \GeoGrainUnderline{0.1} & 0.6 & 2.9 & 7.8 \\
CC-Diff~\cite{ccdiff2024} & 2.1 & 9.1 & 0.2 & 0.1 & 0.2 & 3.0 & 6.5 & 0.3 & 1.4 & 0.1 & 0.0 & 0.0 & 0.3 & 3.3 \\
CC-Diff++~\cite{ccdiffpp2025} & 1.4 & 6.5 & 0.2 & 0.1 & 0.4 & 2.1 & 4.7 & 0.2 & 1.1 & 0.0 & 0.0 & 0.1 & 0.2 & 2.2 \\
FICGen~\cite{ficgen2025} & 6.3 & 25.0 & 0.8 & 0.2 & 1.6 & 10.0 & 18.0 & 0.8 & 3.2 & 0.2 & 0.0 & 0.1 & 1.3 & 9.2 \\
SiamLayout~\cite{creatilayout2025} & 0.0 & 0.1 & 0.0 & 0.0 & 0.0 & 0.0 & 0.0 & 0.0 & 0.0 & 0.0 & 0.0 & 0.0 & 0.0 & 0.0 \\
Laytrol~\cite{laytrol2026} & 2.2 & 10.4 & 0.2 & 0.3 & 1.0 & 3.3 & 3.0 & 0.4 & 2.0 & 0.2 & \GeoGrainUnderline{0.1} & 0.2 & 0.5 & 2.0 \\
TerraDiT-$\Omega$~\cite{terraditomega2026} & \GeoGrainUnderline{11.7} & \GeoGrainUnderline{35.1} & \GeoGrainUnderline{4.1} & \GeoGrainUnderline{0.8} & \GeoGrainUnderline{6.7} & \GeoGrainUnderline{18.9} & \GeoGrainUnderline{19.0} & \GeoGrainUnderline{3.2} & \GeoGrainUnderline{12.3} & \GeoGrainUnderline{0.6} & \GeoGrainUnderline{0.1} & \GeoGrainUnderline{1.7} & \GeoGrainUnderline{6.9} & \GeoGrainUnderline{13.3} \\
\rowcolor{RoyalPurple!5} \textbf{GeoGrain (Ours)} & \textbf{33.2} & \textbf{71.9} & \textbf{24.3} & \textbf{7.4} & \textbf{27.1} & \textbf{43.2} & \textbf{43.3} & \textbf{11.2} & \textbf{35.7} & \textbf{2.9} & \textbf{0.4} & \textbf{8.5} & \textbf{20.3} & \textbf{28.9} \\
\bottomrule
\end{tabularx}
\par\vskip5.85pt\relax 
\end{table*}
}
\section{Additional Quantitative Results}
\label{supp:quantitative}

This section extends the generation quality, layout fidelity, and trainability evaluations in \Cref{tab:main,tab:trainability}. We examine layout fidelity and trainability across different detectors and backbones, analyze class-wise performance for both evaluations, and extend the evaluation of all three aspects to DIOR-R. We also compare inference time across methods. Unless otherwise specified, layout fidelity is evaluated using fixed detectors trained on real training images for 12 epochs (the standard $1\times$ schedule), whereas trainability is evaluated using detectors trained for the same schedule on the full real training set augmented with an equal number of synthetic images (Real 100\% + Synthetic 100\%). Layout fidelity is measured on images generated from the test layouts, whereas trainability is measured on real test images~\cite{geodiffusion2024}.

\subsection{Detector and Backbone Generalization}
\label{supp:detectors}

For the results reported in \Cref{tab:main,tab:trainability}, we use DCFL~\cite{dcfl2023} with Rotated RetinaNet~\cite{retinanet2017} (ResNet-50~\cite{he2016deep}), following the official AI-TOD-R configuration~\cite{aitodr2026} using an MMRotate-based implementation~\cite{mmrotate2022} with the standard $1\times$ schedule (12 epochs). To examine whether GeoGrain's performance generalizes across detector architectures and backbones, we evaluate both layout fidelity and trainability using three additional configurations, as reported in \Cref{tab:supp_fidelity_detectors,tab:supp_trainability_detectors}. These comprise DCFL with Oriented R-CNN~\cite{xie2021oriented} (ResNet-50) and standard Oriented R-CNN with ResNet-50 and Swin-T~\cite{liu2021swin} backbones. The first follows the official AI-TOD-R configuration, while the latter two follow the official MMRotate configurations. Specifically, the first \mbox{configuration} changes the detector architecture within DCFL while retaining the ResNet-50 backbone, whereas the latter two examine standard Oriented R-CNN with different backbones.

\GeoGrainDeferFloat{detectortrainability}{\begin{table*}[t]
\centering
\caption{\textbf{Quantitative results for Trainability across detectors and backbones.} $\Delta$AP is relative to real-only training. The best and second-best results are \textbf{bold} and \GeoGrainUnderline{underlined}, respectively.}
\label{tab:supp_trainability_detectors}
\GeoGrainTableFont{\fontsize{8.0}{9.0}\selectfont}
\setlength{\tabcolsep}{2.6pt}
\renewcommand{\arraystretch}{1.10}
\begin{tabularx}{\textwidth}{l*{16}{Z}}
\toprule
\multirow{3}{*}{\textbf{Method}} & \multicolumn{8}{c}{\textbf{SODA-A}} & \multicolumn{8}{c}{\textbf{AI-TOD-R}} \\
\cmidrule(lr){2-9}\cmidrule(l){10-17}
 & \multirow{2}{*}{AP} & \multirow{2}{*}{$\Delta$AP} & \multirow{2}{*}{AP$_{50}$} & \multirow{2}{*}{AP$_{75}$} & \multicolumn{4}{c}{AP by object size} & \multirow{2}{*}{AP} & \multirow{2}{*}{$\Delta$AP} & \multirow{2}{*}{AP$_{50}$} & \multirow{2}{*}{AP$_{75}$} & \multicolumn{4}{c}{AP by object size} \\
\cmidrule(lr){6-9}\cmidrule(l){14-17}
 &  &  &  &  & \textit{eS} & \textit{rS} & \textit{gS} & \textit{N} &  &  &  &  & \textit{vt} & \textit{t} & \textit{s} & \textit{m} \\
\midrule
\rowcolor{gray!5}\multicolumn{17}{l}{\textbf{DCFL w/ Oriented R-CNN (ResNet-50)}} \\
Real only & 35.2 & -- & 75.2 & 26.7 & 11.4 & 28.6 & 43.9 & 40.1 & 12.5 & -- & 40.7 & 3.4 & 4.6 & 12.1 & 16.7 & 21.3 \\
AeroGen~\cite{aerogen2025} & 34.3 & \textminus{}0.9 & 74.5 & 25.3 & 11.0 & 27.7 & 42.5 & \GeoGrainUnderline{40.1} & 12.2 & \textminus{}0.3 & 39.5 & 3.6 & 4.4 & 11.5 & 16.7 & 22.0 \\
CC-Diff~\cite{ccdiff2024} & 34.2 & \textminus{}1.0 & 74.1 & 25.2 & \GeoGrainUnderline{11.2} & 27.8 & 42.3 & 38.9 & 12.5 & 0.0 & 40.4 & 3.4 & \GeoGrainUnderline{4.7} & 11.9 & 17.3 & 21.3 \\
CC-Diff++~\cite{ccdiffpp2025} & \GeoGrainUnderline{34.5} & \GeoGrainUnderline{\textminus{}0.7} & \GeoGrainUnderline{74.6} & 25.5 & \GeoGrainUnderline{11.2} & \GeoGrainUnderline{28.1} & \GeoGrainUnderline{43.6} & 39.4 & 12.8 & +0.3 & 40.7 & 3.4 & 4.5 & 12.2 & \GeoGrainUnderline{17.5} & 21.6 \\
FICGen~\cite{ficgen2025} & 34.4 & \textminus{}0.8 & 74.4 & 25.5 & \GeoGrainUnderline{11.2} & 28.0 & \GeoGrainUnderline{43.6} & 38.8 & \GeoGrainUnderline{12.9} & \GeoGrainUnderline{+0.4} & \GeoGrainUnderline{41.6} & 3.4 & \GeoGrainUnderline{4.7} & \GeoGrainUnderline{12.3} & 17.1 & 22.1 \\
SiamLayout~\cite{creatilayout2025} & 33.1 & \textminus{}2.1 & 72.8 & 24.2 & 10.2 & 26.9 & 40.4 & 37.8 & 12.0 & \textminus{}0.5 & 38.8 & 3.3 & 4.2 & 11.8 & 16.4 & 20.5 \\
Laytrol~\cite{laytrol2026} & 34.3 & \textminus{}0.9 & 73.6 & \GeoGrainUnderline{25.8} & 11.0 & 27.7 & 43.5 & 39.9 & 11.8 & \textminus{}0.7 & 37.5 & 3.3 & 4.2 & 11.5 & 16.1 & 18.7 \\
TerraDiT-$\Omega$~\cite{terraditomega2026} & 33.6 & \textminus{}1.6 & 72.3 & 25.3 & 10.4 & 26.9 & 40.1 & 39.4 & 12.4 & \textminus{}0.1 & 40.2 & \GeoGrainUnderline{3.7} & \GeoGrainUnderline{4.7} & 11.9 & \GeoGrainUnderline{17.5} & \GeoGrainUnderline{22.2} \\
\rowcolor{RoyalPurple!5} \textbf{GeoGrain (Ours)} & \textbf{35.7} & \textbf{+0.5} & \textbf{75.8} & \textbf{27.5} & \textbf{12.4} & \textbf{29.5} & \textbf{44.0} & \textbf{40.6} & \textbf{14.0} & \textbf{+1.5} & \textbf{44.1} & \textbf{4.7} & \textbf{5.1} & \textbf{13.5} & \textbf{18.9} & \textbf{23.2} \\
\midrule
\rowcolor{gray!5}\multicolumn{17}{l}{\textbf{Oriented R-CNN (ResNet-50)}} \\
Real only & 33.7 & -- & 69.2 & 27.6 & 6.8 & 27.3 & 44.6 & 41.2 & 10.5 & -- & 32.4 & 3.2 & 0.4 & 9.1 & 18.4 & 22.3 \\
AeroGen~\cite{aerogen2025} & 33.1 & \textminus{}0.6 & 68.5 & \GeoGrainUnderline{27.5} & \GeoGrainUnderline{6.7} & 26.4 & 43.2 & \GeoGrainUnderline{41.7} & \GeoGrainUnderline{10.2} & \GeoGrainUnderline{\textminus{}0.3} & \GeoGrainUnderline{31.6} & \GeoGrainUnderline{3.3} & 0.3 & \GeoGrainUnderline{8.7} & 17.8 & 21.4 \\
CC-Diff~\cite{ccdiff2024} & \GeoGrainUnderline{33.2} & \GeoGrainUnderline{\textminus{}0.5} & \GeoGrainUnderline{69.0} & 26.5 & 6.5 & 26.9 & 44.4 & 40.9 & 9.7 & \textminus{}0.8 & 29.8 & 2.9 & \GeoGrainUnderline{0.6} & 8.2 & 17.8 & 21.3 \\
CC-Diff++~\cite{ccdiffpp2025} & \GeoGrainUnderline{33.2} & \GeoGrainUnderline{\textminus{}0.5} & 68.9 & 26.8 & 6.4 & \GeoGrainUnderline{27.0} & \GeoGrainUnderline{44.7} & 40.8 & 9.4 & \textminus{}1.1 & 29.3 & 2.7 & 0.5 & 7.9 & 17.5 & 21.0 \\
FICGen~\cite{ficgen2025} & 33.1 & \textminus{}0.6 & 68.8 & 26.4 & 6.5 & 26.7 & 44.6 & 40.9 & 9.8 & \textminus{}0.7 & 30.4 & 2.9 & 0.5 & 8.1 & 18.3 & \GeoGrainUnderline{21.7} \\
SiamLayout~\cite{creatilayout2025} & 32.3 & \textminus{}1.4 & 67.9 & 25.5 & 6.2 & 26.3 & 42.1 & 39.7 & 8.1 & \textminus{}2.4 & 24.6 & 2.6 & 0.3 & 6.8 & 15.8 & 18.9 \\
Laytrol~\cite{laytrol2026} & \GeoGrainUnderline{33.2} & \GeoGrainUnderline{\textminus{}0.5} & 68.7 & 26.5 & 6.6 & 26.6 & 44.2 & 41.5 & 9.1 & \textminus{}1.4 & 28.4 & 2.6 & 0.5 & 7.8 & 16.7 & 19.0 \\
TerraDiT-$\Omega$~\cite{terraditomega2026} & 32.0 & \textminus{}1.7 & 67.5 & 24.9 & 6.3 & 25.4 & 41.4 & 40.3 & 10.1 & \textminus{}0.4 & 31.2 & 3.0 & 0.4 & 8.6 & \GeoGrainUnderline{18.7} & 21.6 \\
\rowcolor{RoyalPurple!5} \textbf{GeoGrain (Ours)} & \textbf{34.2} & \textbf{+0.5} & \textbf{69.7} & \textbf{28.4} & \textbf{7.3} & \textbf{27.8} & \textbf{45.2} & \textbf{42.0} & \textbf{11.4} & \textbf{+0.9} & \textbf{34.2} & \textbf{3.9} & \textbf{0.8} & \textbf{9.8} & \textbf{20.1} & \textbf{24.1} \\
\midrule
\rowcolor{gray!5}\multicolumn{17}{l}{\textbf{Oriented R-CNN (Swin-T)}} \\
Real only & 33.8 & -- & 70.0 & 27.3 & 8.1 & 28.0 & 46.2 & 42.2 & 10.5 & -- & 33.0 & 3.0 & 0.6 & 8.8 & 18.7 & 23.3 \\
AeroGen~\cite{aerogen2025} & 33.8 & 0.0 & 70.4 & 26.7 & \GeoGrainUnderline{8.1} & 27.6 & 44.8 & 41.8 & 10.1 & \textminus{}0.4 & 31.0 & 2.9 & 0.4 & \GeoGrainUnderline{8.4} & 17.8 & \GeoGrainUnderline{23.2} \\
CC-Diff~\cite{ccdiff2024} & 33.5 & \textminus{}0.3 & 70.2 & 26.1 & 7.8 & 27.5 & 45.8 & 41.1 & 9.7 & \textminus{}0.8 & 29.8 & 2.7 & 0.4 & 8.0 & 17.5 & 22.5 \\
CC-Diff++~\cite{ccdiffpp2025} & 33.6 & \textminus{}0.2 & \GeoGrainUnderline{70.7} & 26.3 & 7.9 & \GeoGrainUnderline{28.0} & 45.8 & 41.0 & 9.5 & \textminus{}1.0 & 29.8 & 2.6 & 0.5 & 8.0 & 17.3 & 22.2 \\
FICGen~\cite{ficgen2025} & \GeoGrainUnderline{33.9} & \GeoGrainUnderline{+0.1} & \GeoGrainUnderline{70.7} & \GeoGrainUnderline{27.1} & 7.9 & 27.8 & \GeoGrainUnderline{46.2} & 41.4 & 9.7 & \textminus{}0.8 & 30.3 & 2.6 & \GeoGrainUnderline{0.6} & 8.0 & 17.6 & 22.4 \\
SiamLayout~\cite{creatilayout2025} & 33.6 & \textminus{}0.2 & 69.9 & 26.8 & 7.9 & 27.6 & 44.8 & 41.1 & 9.0 & \textminus{}1.5 & 28.6 & 2.4 & 0.5 & 7.5 & 16.4 & 20.4 \\
Laytrol~\cite{laytrol2026} & 33.5 & \textminus{}0.3 & 70.1 & 26.3 & 7.9 & 27.6 & 45.7 & 41.9 & 9.2 & \textminus{}1.3 & 28.7 & 2.6 & 0.4 & 7.9 & 17.1 & 20.9 \\
TerraDiT-$\Omega$~\cite{terraditomega2026} & 33.6 & \textminus{}0.2 & 70.4 & 26.4 & 7.9 & 27.7 & 43.9 & \GeoGrainUnderline{42.0} & \GeoGrainUnderline{10.2} & \GeoGrainUnderline{\textminus{}0.3} & \GeoGrainUnderline{31.7} & \GeoGrainUnderline{3.1} & 0.4 & \GeoGrainUnderline{8.4} & \GeoGrainUnderline{18.3} & 21.7 \\
\rowcolor{RoyalPurple!5} \textbf{GeoGrain (Ours)} & \textbf{34.7} & \textbf{+0.9} & \textbf{70.9} & \textbf{28.9} & \textbf{8.4} & \textbf{28.9} & \textbf{47.3} & \textbf{42.6} & \textbf{11.0} & \textbf{+0.5} & \textbf{33.2} & \textbf{3.9} & \textbf{0.7} & \textbf{9.4} & \textbf{19.3} & \textbf{24.2} \\
\bottomrule
\end{tabularx}
\par\vskip-6.4pt\relax 
\end{table*}
}
In \Cref{tab:main}, GeoGrain achieves a SODA-A layout AP of 29.6, compared with 11.0 for TerraDiT-$\Omega$, the strongest baseline. Extending this comparison to three additional detector configurations, \Cref{tab:supp_fidelity_detectors} shows that GeoGrain achieves the highest layout AP among the compared L2I methods in all six dataset--configuration combinations, exceeding TerraDiT-$\Omega$ by 21.3--21.9 AP points on SODA-A and 7.5--8.5 AP points on AI-\mbox{TOD-R.} The same advantage holds for AP$_{50}$, AP$_{75}$, and every object-size range among the compared L2I methods. Notably, GeoGrain maintains layout AP within 1.2 points of the corresponding real-image reference across diverse detector architectures and backbones on both datasets. These results demonstrate that GeoGrain's layout fidelity generalizes beyond the detector used in the main evaluation.

\GeoGrainShowFloat{detectorfidelity}

\Cref{tab:trainability} reports that GeoGrain improves trainability AP over real-only training by 1.8 points on SODA-A and 2.1 points on AI-TOD-R. To assess whether this benefit extends to other detector architectures and backbones, \Cref{tab:supp_trainability_detectors} reports results for the same three additional configurations, with each detector trained for 12 epochs using Real 100\% + Synthetic 100\%. GeoGrain improves over the corresponding real-only reference in all six dataset--configuration combinations, with gains of 0.5--0.9 points on SODA-A and 0.5--1.5 points on AI-TOD-R. It also achieves the highest AP among the compared L2I methods in every configuration. Although the magnitude of the gains varies across detectors and datasets, their consistent direction indicates that GeoGrain's synthetic images provide useful training signals across diverse detector configurations.

\GeoGrainShowFloat{detectortrainability}

\GeoGrainDeferFloat{classsoda}{\renewcommand{\dblfloatpagefraction}{0.85}
\begin{table*}[p]
\centering
\caption{\textbf{Quantitative results for class-wise Layout fidelity and Trainability on the SODA-A dataset.} The best and second-best results are \textbf{bold} and \GeoGrainUnderline{underlined}, respectively.}
\label{tab:supp_classes_soda}
\GeoGrainTableFont{\fontsize{8.0}{9.0}\selectfont}
\setlength{\tabcolsep}{2.6pt}
\renewcommand{\arraystretch}{1.08}
\begin{tabularx}{\textwidth}{l*{10}{Z}}
\toprule
\multirow{2}{*}{\textbf{Method}} & \multirow{2}{*}{AP} & \multicolumn{9}{c}{AP by class} \\
\cmidrule(l){3-11}
 & & \textit{APL} & \textit{HEL} & \textit{SV} & \textit{LV} & \textit{SH} & \textit{CT} & \textit{ST} & \textit{SP} & \textit{WM} \\
\midrule
\rowcolor{gray!5}\multicolumn{11}{l}{\textbf{Layout fidelity} $\uparrow$} \\
Real images & 29.0 & 46.3 & 27.3 & 21.1 & 17.9 & 27.0 & 30.3 & 36.1 & 34.1 & 20.8 \\
AeroGen~\cite{aerogen2025} & 4.6 & 7.9 & 6.4 & 0.7 & 2.0 & 3.3 & 4.0 & 7.9 & 6.2 & 2.7 \\
CC-Diff~\cite{ccdiff2024} & 2.3 & 1.8 & 5.1 & 0.1 & 0.5 & 0.3 & 1.5 & 0.7 & 6.7 & 3.7 \\
CC-Diff++~\cite{ccdiffpp2025} & 1.6 & 1.1 & 4.2 & 0.1 & 0.4 & 0.3 & 1.4 & 0.3 & 4.0 & 2.3 \\
FICGen~\cite{ficgen2025} & 6.5 & 7.0 & \GeoGrainUnderline{10.6} & 0.5 & \GeoGrainUnderline{3.3} & 1.6 & 4.5 & 5.9 & 17.2 & \GeoGrainUnderline{7.8} \\
SiamLayout~\cite{creatilayout2025} & 0.0 & 0.0 & 0.0 & 0.0 & 0.0 & 0.0 & 0.0 & 0.0 & 0.0 & 0.0 \\
Laytrol~\cite{laytrol2026} & 2.2 & 3.3 & 5.9 & 0.8 & 0.6 & 2.0 & 2.2 & 2.1 & 1.7 & 1.4 \\
TerraDiT-$\Omega$~\cite{terraditomega2026} & \GeoGrainUnderline{11.0} & \GeoGrainUnderline{14.3} & 5.8 & \GeoGrainUnderline{5.0} & 2.6 & \GeoGrainUnderline{11.2} & \GeoGrainUnderline{15.3} & \GeoGrainUnderline{20.4} & \GeoGrainUnderline{18.3} & 6.2 \\
\rowcolor{RoyalPurple!5} \textbf{GeoGrain (Ours)} & \textbf{29.6} & \textbf{43.8} & \textbf{28.3} & \textbf{22.0} & \textbf{19.9} & \textbf{31.8} & \textbf{30.8} & \textbf{36.8} & \textbf{33.6} & \textbf{19.8} \\
\midrule
\rowcolor{gray!5}\multicolumn{11}{l}{\textbf{Trainability} $\uparrow$} \\
Real only & 29.0 & 46.3 & 27.3 & 21.1 & 17.9 & 27.0 & 30.3 & 36.1 & 34.1 & 20.8 \\
AeroGen~\cite{aerogen2025} & \GeoGrainUnderline{29.2} & \GeoGrainUnderline{46.8} & \GeoGrainUnderline{27.6} & 21.0 & \GeoGrainUnderline{18.0} & 28.3 & 30.4 & 35.6 & 34.1 & 20.5 \\
CC-Diff~\cite{ccdiff2024} & 28.4 & 45.8 & 26.4 & 20.8 & 17.3 & 25.9 & 29.3 & 35.9 & 33.8 & \GeoGrainUnderline{20.8} \\
CC-Diff++~\cite{ccdiffpp2025} & 28.4 & 45.5 & 25.7 & 21.1 & 17.5 & 27.0 & 28.9 & 35.6 & 33.6 & \GeoGrainUnderline{20.8} \\
FICGen~\cite{ficgen2025} & 29.1 & 46.4 & 26.6 & \GeoGrainUnderline{21.4} & 17.9 & 28.1 & 29.4 & 36.1 & \GeoGrainUnderline{34.8} & \GeoGrainUnderline{20.8} \\
SiamLayout~\cite{creatilayout2025} & 27.9 & 45.2 & 24.5 & 20.4 & 16.0 & \GeoGrainUnderline{28.5} & 28.9 & 34.4 & 33.7 & 19.6 \\
Laytrol~\cite{laytrol2026} & 28.3 & 45.7 & 26.6 & 20.4 & 17.6 & 25.7 & 29.5 & 35.6 & 33.4 & 20.5 \\
TerraDiT-$\Omega$~\cite{terraditomega2026} & 28.5 & 45.8 & 25.9 & 20.9 & 16.7 & 26.3 & \GeoGrainUnderline{31.0} & \GeoGrainUnderline{36.4} & 34.3 & 19.7 \\
\rowcolor{RoyalPurple!5} \textbf{GeoGrain (Ours)} & \textbf{30.8} & \textbf{48.2} & \textbf{28.0} & \textbf{23.1} & \textbf{20.6} & \textbf{29.6} & \textbf{33.6} & \textbf{37.6} & \textbf{35.4} & \textbf{21.2} \\
\bottomrule
\end{tabularx}
\par\smallskip\raggedright\scriptsize
APL: airplane; HEL: helicopter; SV/LV: small/large vehicle; SH: ship; CT: container; ST: storage tank; SP: swimming pool; WM: windmill.
\end{table*}
}
\GeoGrainDeferFloat{classaitod}{\begin{table*}[p]
\centering
\caption{\textbf{Quantitative results for class-wise Layout fidelity and Trainability on the AI-TOD-R dataset.} The best and second-best results are \textbf{bold} and \GeoGrainUnderline{underlined}, respectively.}
\label{tab:supp_classes_aitod}
\GeoGrainTableFont{\fontsize{8.0}{9.0}\selectfont}
\setlength{\tabcolsep}{2.6pt}
\renewcommand{\arraystretch}{1.08}
\begin{tabularx}{\textwidth}{l*{9}{Z}}
\toprule
\multirow{2}{*}{\textbf{Method}} & \multirow{2}{*}{AP} & \multicolumn{8}{c}{AP by class} \\
\cmidrule(l){3-10}
 & & \textit{APL} & \textit{BR} & \textit{ST} & \textit{SH} & \textit{SP} & \textit{VE} & \textit{PE} & \textit{WM} \\
\midrule
\rowcolor{gray!5}\multicolumn{10}{l}{\textbf{Layout fidelity} $\uparrow$} \\
Real images & 9.1 & 9.9 & 4.3 & 12.9 & 17.5 & 11.2 & 12.8 & 3.8 & 0.2 \\
AeroGen~\cite{aerogen2025} & 1.3 & \GeoGrainUnderline{3.7} & \GeoGrainUnderline{0.9} & 1.3 & 1.3 & 2.1 & 0.8 & 0.2 & 0.0 \\
CC-Diff~\cite{ccdiff2024} & 0.3 & 1.7 & 0.0 & 0.1 & 0.2 & 0.0 & 0.1 & 0.1 & 0.0 \\
CC-Diff++~\cite{ccdiffpp2025} & 0.2 & 1.0 & 0.0 & 0.1 & 0.1 & 0.0 & 0.1 & 0.4 & 0.0 \\
FICGen~\cite{ficgen2025} & 0.8 & 3.5 & 0.0 & 1.0 & 0.7 & 0.1 & 0.6 & 0.2 & 0.0 \\
SiamLayout~\cite{creatilayout2025} & 0.0 & 0.0 & 0.0 & 0.0 & 0.0 & 0.0 & 0.0 & 0.0 & 0.0 \\
Laytrol~\cite{laytrol2026} & 0.3 & 1.0 & 0.1 & 1.0 & 0.1 & 0.2 & 0.3 & 0.1 & 0.0 \\
TerraDiT-$\Omega$~\cite{terraditomega2026} & \GeoGrainUnderline{2.7} & 2.3 & \GeoGrainUnderline{0.9} & \GeoGrainUnderline{6.4} & \GeoGrainUnderline{3.5} & \GeoGrainUnderline{5.9} & \GeoGrainUnderline{2.3} & \GeoGrainUnderline{0.7} & 0.0 \\
\rowcolor{RoyalPurple!5} \textbf{GeoGrain (Ours)} & \textbf{9.0} & \textbf{11.6} & \textbf{3.9} & \textbf{13.9} & \textbf{13.5} & \textbf{12.7} & \textbf{11.6} & \textbf{4.3} & \textbf{0.3} \\
\midrule
\rowcolor{gray!5}\multicolumn{10}{l}{\textbf{Trainability} $\uparrow$} \\
Real only & 9.1 & 9.9 & 4.3 & 12.9 & 17.5 & 11.2 & 12.8 & 3.8 & 0.2 \\
AeroGen~\cite{aerogen2025} & 9.3 & 11.1 & 5.2 & 11.7 & 16.3 & 11.7 & 12.0 & 3.5 & \GeoGrainUnderline{2.8} \\
CC-Diff~\cite{ccdiff2024} & 8.5 & 9.5 & \GeoGrainUnderline{5.6} & 10.4 & 15.7 & 11.3 & 11.8 & 3.6 & 0.3 \\
CC-Diff++~\cite{ccdiffpp2025} & 8.8 & 9.9 & 5.3 & 11.5 & 15.9 & 10.7 & 12.1 & 3.7 & 1.0 \\
FICGen~\cite{ficgen2025} & 8.9 & 9.3 & \textbf{5.7} & 12.2 & 15.8 & 11.7 & 12.0 & \GeoGrainUnderline{3.8} & 0.9 \\
SiamLayout~\cite{creatilayout2025} & 6.7 & 1.3 & \GeoGrainUnderline{5.6} & 9.1 & 15.4 & 7.3 & 11.9 & 2.9 & 0.2 \\
Laytrol~\cite{laytrol2026} & 8.8 & 10.6 & 4.2 & 11.5 & 15.2 & 11.1 & 11.6 & 3.4 & 2.6 \\
TerraDiT-$\Omega$~\cite{terraditomega2026} & \GeoGrainUnderline{9.5} & \GeoGrainUnderline{11.2} & 4.4 & \GeoGrainUnderline{13.6} & \GeoGrainUnderline{16.7} & \GeoGrainUnderline{12.5} & \GeoGrainUnderline{12.7} & 3.5 & 1.2 \\
\rowcolor{RoyalPurple!5} \textbf{GeoGrain (Ours)} & \textbf{11.2} & \textbf{13.8} & \textbf{5.7} & \textbf{15.1} & \textbf{18.6} & \textbf{14.5} & \textbf{14.2} & \textbf{4.5} & \textbf{3.6} \\
\bottomrule
\end{tabularx}
\par\smallskip\raggedright\scriptsize
APL: airplane; BR: bridge; ST: storage tank; SH: ship; SP: swimming pool; VE: vehicle; PE: person; WM: windmill.
\end{table*}
}
\subsection{Class-wise Evaluation}
\label{supp:perclass}

We analyze class-wise layout fidelity and trainability on SODA-A and AI-TOD-R using DCFL with Rotated \mbox{RetinaNet} (ResNet-50).

On SODA-A, \Cref{tab:supp_classes_soda} shows that GeoGrain achieves the highest layout AP among the compared L2I methods in all nine categories. Its class-wise layout AP reaches 43.8 for airplanes, 31.8 for ships, and 36.8 for storage tanks, with an overall layout AP of 29.6, compared with 11.0 for TerraDiT-$\Omega$. For trainability, GeoGrain improves over real-only training in all nine categories, raising the mean AP from 29.0 to 30.8. The gains are particularly notable for containers (+3.3 points), large vehicles (+2.7), and ships (+2.6), showing that the benefit extends across different object categories.

\GeoGrainShowFloat{classsoda}

\GeoGrainDeferFloat{diorfidelity}{\begin{table*}[t]
\centering
\caption{\textbf{Quantitative results for Generation quality and Layout fidelity on the DIOR-R dataset.} KID and SceneKID are scaled by $10^3$. The best and second-best results are \textbf{bold} and \GeoGrainUnderline{underlined}, respectively.}
\label{tab:supp_dior_fidelity}
\label{tab:supp_dior}
\GeoGrainTableFont{\fontsize{8.0}{9.0}\selectfont}
\setlength{\tabcolsep}{1.6pt}
\renewcommand{\arraystretch}{1.10}
\begin{tabularx}{\textwidth}{l*{16}{Z}}
\toprule
\multirow{3}{*}{\textbf{Method}} & \multicolumn{4}{c}{\textbf{Generation quality} $\downarrow$} & \multicolumn{12}{c}{\textbf{Layout fidelity} $\uparrow$} \\
\cmidrule(lr){2-5}\cmidrule(l){6-17}
 & \multirow{2}{*}{FID} & \multirow{2}{*}{KID} & \multirow{2}{*}{\makecell{Scene\\FID}} & \multirow{2}{*}{\makecell{Scene\\KID}} & \multicolumn{3}{c}{\makecell{DCFL w/ Rotated\\RetinaNet\\(ResNet-50)}} & \multicolumn{3}{c}{\makecell{DCFL w/ Oriented\\R-CNN\\(ResNet-50)}} & \multicolumn{3}{c}{\makecell{Oriented R-CNN\\(ResNet-50)}} & \multicolumn{3}{c}{\makecell{Oriented R-CNN\\(Swin-T)}} \\
\cmidrule(lr){6-8}\cmidrule(lr){9-11}\cmidrule(lr){12-14}\cmidrule(l){15-17}
 & & & & & AP & AP$_{50}$ & AP$_{75}$ & AP & AP$_{50}$ & AP$_{75}$ & AP & AP$_{50}$ & AP$_{75}$ & AP & AP$_{50}$ & AP$_{75}$ \\
\midrule
Real images & -- & -- & -- & -- & 28.04 & 50.77 & 26.38 & 30.42 & 52.22 & 31.34 & 31.86 & 52.32 & 32.80 & 33.49 & 55.06 & 34.77 \\
AeroGen~\cite{aerogen2025} & 26.77 & 12.99 & 25.56 & 10.04 & 16.33 & 33.15 & 14.42 & 16.49 & 32.11 & 15.34 & 17.71 & 33.69 & 15.94 & 18.20 & 34.37 & 18.12 \\
CC-Diff~\cite{ccdiff2024} & 47.87 & 28.31 & 56.78 & 30.53 & 19.86 & 38.10 & 17.95 & 20.45 & 37.90 & 19.52 & 21.75 & 39.76 & 21.12 & 22.45 & 39.97 & 22.15 \\
CC-Diff++~\cite{ccdiffpp2025} & 47.73 & 27.87 & 60.66 & 33.44 & 20.19 & 38.43 & 18.79 & 20.45 & 38.04 & 19.87 & 22.00 & 39.73 & 22.14 & 22.86 & 40.74 & 23.19 \\
FICGen~\cite{ficgen2025} & 45.61 & 24.91 & 44.27 & 20.19 & \GeoGrainUnderline{24.93} & \GeoGrainUnderline{45.35} & \GeoGrainUnderline{24.42} & \GeoGrainUnderline{25.68} & 44.75 & \GeoGrainUnderline{25.89} & \GeoGrainUnderline{27.59} & \GeoGrainUnderline{46.74} & \GeoGrainUnderline{28.75} & \GeoGrainUnderline{29.16} & \GeoGrainUnderline{49.22} & \GeoGrainUnderline{30.77} \\
SiamLayout~\cite{creatilayout2025} & 54.84 & 18.52 & 83.89 & 48.27 & 0.44 & 1.59 & 0.17 & 0.46 & 1.55 & 0.19 & 0.64 & 2.47 & 0.08 & 0.37 & 1.56 & 0.09 \\
Laytrol~\cite{laytrol2026} & 27.55 & 9.44 & 25.65 & 10.06 & 18.63 & 42.46 & 14.47 & 18.54 & 42.09 & 13.83 & 18.93 & 42.71 & 14.45 & 19.81 & 44.34 & 15.54 \\
TerraDiT-$\Omega$~\cite{terraditomega2026} & \GeoGrainUnderline{17.46} & \GeoGrainUnderline{8.03} & \GeoGrainUnderline{16.83} & \GeoGrainUnderline{6.47} & 23.60 & 45.01 & 21.59 & 25.01 & \GeoGrainUnderline{45.17} & 23.33 & 26.15 & 45.60 & 25.91 & 26.92 & 46.79 & 27.20 \\
\rowcolor{RoyalPurple!5} \textbf{GeoGrain (Ours)} & \textbf{11.82} & \textbf{2.26} & \textbf{6.96} & \textbf{1.80} & \textbf{29.66} & \textbf{54.69} & \textbf{27.76} & \textbf{32.27} & \textbf{55.84} & \textbf{32.30} & \textbf{33.48} & \textbf{55.70} & \textbf{34.26} & \textbf{34.45} & \textbf{58.15} & \textbf{35.28} \\
\bottomrule
\end{tabularx}
\end{table*}
}
\GeoGrainDeferFloat{diortrainability}{\begin{table*}[t]
\centering
\caption{\textbf{Quantitative results for Trainability on the DIOR-R dataset.} $\Delta$AP is relative to real-only training. The best and second-best results are \textbf{bold} and \GeoGrainUnderline{underlined}, respectively.}
\label{tab:supp_dior_trainability}
\GeoGrainTableFont{\fontsize{8.0}{9.0}\selectfont}
\setlength{\tabcolsep}{1.6pt}
\renewcommand{\arraystretch}{1.10}
\begin{tabularx}{\textwidth}{l*{16}{Z}}
\toprule
\multirow{2}{*}{\textbf{Method}} & \multicolumn{4}{c}{\makecell{\textbf{DCFL w/ Rotated RetinaNet}\\\textbf{(ResNet-50)}}} & \multicolumn{4}{c}{\makecell{\textbf{DCFL w/ Oriented R-CNN}\\\textbf{(ResNet-50)}}} & \multicolumn{4}{c}{\makecell{\textbf{Oriented R-CNN}\\\textbf{(ResNet-50)}}} & \multicolumn{4}{c}{\makecell{\textbf{Oriented R-CNN}\\\textbf{(Swin-T)}}} \\
\cmidrule(lr){2-5}\cmidrule(lr){6-9}\cmidrule(lr){10-13}\cmidrule(l){14-17}
 & AP & $\Delta$AP & AP$_{50}$ & AP$_{75}$ & AP & $\Delta$AP & AP$_{50}$ & AP$_{75}$ & AP & $\Delta$AP & AP$_{50}$ & AP$_{75}$ & AP & $\Delta$AP & AP$_{50}$ & AP$_{75}$ \\
\midrule
Real only & 28.04 & -- & 50.77 & 26.38 & 30.42 & -- & 52.22 & 31.34 & 31.86 & -- & 52.32 & 32.80 & 33.49 & -- & 55.06 & 34.77 \\
AeroGen~\cite{aerogen2025} & 29.35 & +1.31 & 52.44 & 28.03 & 32.11 & +1.69 & 53.76 & 32.48 & 32.46 & +0.60 & 51.70 & 33.64 & 33.96 & +0.47 & 54.53 & 35.70 \\
CC-Diff~\cite{ccdiff2024} & 29.40 & +1.36 & 52.01 & 28.25 & 32.15 & +1.73 & \GeoGrainUnderline{54.30} & 33.15 & 33.02 & +1.16 & 53.19 & 34.29 & 33.74 & +0.25 & 54.58 & 34.99 \\
CC-Diff++~\cite{ccdiffpp2025} & 29.08 & +1.04 & 51.88 & 27.78 & 32.05 & +1.63 & 53.87 & 32.48 & 32.64 & +0.78 & 52.50 & 33.96 & 33.48 & \textminus{}0.01 & 54.19 & 34.53 \\
FICGen~\cite{ficgen2025} & 29.63 & +1.59 & 51.87 & 28.27 & 32.23 & +1.81 & 53.71 & 33.24 & 33.12 & +1.26 & 52.52 & 34.33 & 34.07 & +0.58 & 54.19 & 35.48 \\
SiamLayout~\cite{creatilayout2025} & 26.33 & \textminus{}1.71 & 49.04 & 24.52 & 29.03 & \textminus{}1.39 & 50.68 & 29.69 & 29.88 & \textminus{}1.98 & 49.70 & 30.58 & 32.03 & \textminus{}1.46 & 52.79 & 32.92 \\
Laytrol~\cite{laytrol2026} & 28.82 & +0.78 & 51.85 & 27.22 & 31.57 & +1.15 & 53.59 & 32.33 & 32.81 & +0.95 & 53.01 & 33.65 & 33.80 & +0.31 & 54.99 & 34.88 \\
TerraDiT-$\Omega$~\cite{terraditomega2026} & \GeoGrainUnderline{30.14} & \GeoGrainUnderline{+2.10} & \GeoGrainUnderline{53.29} & \GeoGrainUnderline{29.71} & \GeoGrainUnderline{32.57} & \GeoGrainUnderline{+2.15} & \GeoGrainUnderline{54.30} & \GeoGrainUnderline{33.30} & \GeoGrainUnderline{33.68} & \GeoGrainUnderline{+1.82} & \GeoGrainUnderline{53.96} & \GeoGrainUnderline{34.80} & \GeoGrainUnderline{34.40} & \GeoGrainUnderline{+0.91} & \GeoGrainUnderline{55.19} & \GeoGrainUnderline{35.80} \\
\rowcolor{RoyalPurple!5} \textbf{GeoGrain (Ours)} & \textbf{31.56} & \textbf{+3.52} & \textbf{54.94} & \textbf{31.08} & \textbf{34.15} & \textbf{+3.73} & \textbf{56.75} & \textbf{34.65} & \textbf{34.96} & \textbf{+3.10} & \textbf{55.35} & \textbf{36.08} & \textbf{36.00} & \textbf{+2.51} & \textbf{58.04} & \textbf{37.81} \\
\bottomrule
\end{tabularx}
\par\vskip-6.5pt\relax 
\end{table*}
}
A similar pattern is observed on AI-TOD-R. As shown in \Cref{tab:supp_classes_aitod}, GeoGrain achieves the highest layout AP among the compared L2I methods in all eight categories, with an overall layout AP of 9.0, compared with 2.7 for TerraDiT-$\Omega$. Its class-wise layout AP reaches 13.9 for storage tanks, 13.5 for ships, and 12.7 for swimming pools. For trainability, GeoGrain improves over real-only training in all eight categories, increasing the mean AP from 9.1 to 11.2. The largest gains are observed for airplanes (+3.9 points), windmills (+3.4), and swimming pools (+3.3).

\GeoGrainShowFloat{classaitod}

Taken together, the results on both datasets show that GeoGrain provides broad class-wise improvements in layout fidelity and consistently benefits detector training.

\subsection{Evaluation on DIOR-R}
\label{supp:dior}

To assess GeoGrain beyond the small-object benchmarks, we further
evaluate it on DIOR-R, a general remote sensing object detection
benchmark covering a broader range of object categories and sizes.
\Cref{tab:supp_dior_fidelity} reports \mbox{generation} quality and
layout fidelity, with the latter evaluated \mbox{using} the four
detector configurations of Sec.~\ref{supp:detectors} under the
official MMRotate evaluation protocol for \mbox{DIOR-R}
(\mbox{11-point} AP of PASCAL VOC 2007~\cite{everingham2010pascal}).
\Cref{tab:supp_dior_trainability} evaluates trainability under the
same configurations and protocol.

For generation quality, \Cref{tab:supp_dior_fidelity} shows that GeoGrain achieves the best results across all four metrics~\cite{kid2018,fid2017,scenefid2021}. Compared with TerraDiT-$\Omega$, it reduces FID from 17.46 to 11.82, KID from 8.03 to 2.26, SceneFID from 16.83 to 6.96, and SceneKID from 6.47 to 1.80. For layout fidelity, GeoGrain also achieves the highest AP,
AP$_{50}$, and AP$_{75}$ among the compared L2I methods across all four detector configurations, which differ in
architecture and backbone.
Together, these results demonstrate GeoGrain's ability to generate
high-quality images while maintaining accurate layout control on
a general remote sensing dataset with diverse object categories
and sizes.

\GeoGrainShowFloat{diorfidelity}

For trainability, \Cref{tab:supp_dior_trainability} shows that GeoGrain consistently improves over real-only training across all four detector configurations, with AP gains ranging from 2.51 to 3.73 points. It also achieves the highest AP, AP$_{50}$, and AP$_{75}$ among the compared L2I methods in every configuration. These results further demonstrate the utility of GeoGrain's generated images for detector training across diverse remote sensing datasets.

\GeoGrainShowFloat{diortrainability}

\subsection{Inference Efficiency}
\label{supp:runtime}

\GeoGrainDeferFloat{runtime}{\renewcommand{\dblfloatpagefraction}{0.99}
\begin{table}[t]
\centering
\caption{\textbf{Inference time comparison across methods.} The best and second-best results are \textbf{bold} and \GeoGrainUnderline{underlined}, respectively.}
\label{tab:supp_runtime}
\label{tab:speed}
\GeoGrainTableFont{\fontsize{8.0}{9.5}\selectfont}
\setlength{\tabcolsep}{2.5pt}
\renewcommand{\arraystretch}{1.15}
\begin{tabularx}{\columnwidth}{lcZZZ}
\toprule
\multirow{2}{*}{\textbf{Method}} & \multirow{2}{*}{\textbf{Steps}} & \multicolumn{3}{c}{\textbf{Time (s/image)} $\downarrow$} \\
\cmidrule(l){3-5}
 & & SODA-A & AI-TOD-R & DIOR-R \\
\midrule
\rowcolor{gray!5}\multicolumn{5}{l}{\textbf{U-Net-based}} \\
AeroGen~\cite{aerogen2025} & 50 & 3.05 & 2.95 & 2.81 \\
CC-Diff~\cite{ccdiff2024} & 50 & \textbf{2.83} & \textbf{2.76} & \textbf{2.66} \\
CC-Diff++~\cite{ccdiffpp2025} & 50 & \GeoGrainUnderline{2.87} & \GeoGrainUnderline{2.78} & \GeoGrainUnderline{2.70} \\
FICGen~\cite{ficgen2025} & 50 & 3.79 & 3.66 & 3.47 \\
\midrule
\rowcolor{gray!5}\multicolumn{5}{l}{\textbf{DiT-based}} \\
SiamLayout~\cite{creatilayout2025} & 50 & 4.10 & 3.95 & 3.79 \\
Laytrol~\cite{laytrol2026} & 50 & 2.98 & 2.95 & 2.91 \\
TerraDiT-$\Omega$~\cite{terraditomega2026} & 100 & 9.83 & 9.49 & 9.38 \\
\rowcolor{RoyalPurple!5} \textbf{GeoGrain (Ours)} & 50 & 5.18 & 4.97 & 4.68 \\
\bottomrule
\end{tabularx}
\end{table}
}
We compare the inference time of GeoGrain and the baselines on a single NVIDIA B200 GPU with a batch size of 1. \mbox{\Cref{tab:supp_runtime}} reports
the mean inference time per image, including text and layout encoding,
denoising, and VAE decoding. Each method is evaluated using its respective
sampler and guidance settings.

For each dataset, we sort the test images by object count, divide them into
200 groups, and randomly select one image from each group. All methods are
evaluated on the same 200 images per dataset, and we report the mean
inference time across these images.

Under this evaluation protocol, GeoGrain takes 5.18, 4.97, and 4.68 seconds per image on SODA-A, AI-TOD-R, and DIOR-R, respectively. Although most baselines require less inference time,
GeoGrain achieves substantially better generation quality
and layout fidelity across the datasets, as shown in
\Cref{tab:main,tab:supp_dior_fidelity}.

\GeoGrainShowFloat{runtime}

\GeoGrainDeferFloat{architecture}{\begin{table*}[t]
\centering
\caption{\textbf{Ablation results on Dual-branch architectures.} KID and SceneKID are scaled by $10^3$. The best and second-best results are \textbf{bold} and \GeoGrainUnderline{underlined}, respectively.}
\label{tab:supp_architecture}
\label{tab:supp_siamlayout_style}
\GeoGrainTableFont{\fontsize{8.0}{9.5}\selectfont}
\setlength{\tabcolsep}{2pt}
\renewcommand{\arraystretch}{1.15}
\begin{tabularx}{\textwidth}{l*{16}{Z}}
\toprule
\multirow{3}{*}{\textbf{Method}} & \multicolumn{4}{c}{\textbf{Generation quality} $\downarrow$} & \multicolumn{6}{c}{\textbf{Layout fidelity} $\uparrow$} & \multicolumn{6}{c}{\textbf{Trainability} $\uparrow$} \\
\cmidrule(lr){2-5}\cmidrule(lr){6-11}\cmidrule(l){12-17}
 & \multirow{2}{*}{FID} & \multirow{2}{*}{KID} & \multirow{2}{*}{\makecell{Scene\\FID}} & \multirow{2}{*}{\makecell{Scene\\KID}} & \multicolumn{3}{c}{DCFL} & \multicolumn{3}{c}{Oriented R-CNN} & \multicolumn{3}{c}{DCFL} & \multicolumn{3}{c}{Oriented R-CNN} \\
\cmidrule(lr){6-8}\cmidrule(lr){9-11}\cmidrule(lr){12-14}\cmidrule(l){15-17}
 & & & & & AP & AP$_{50}$ & AP$_{75}$ & AP & AP$_{50}$ & AP$_{75}$ & AP & AP$_{50}$ & AP$_{75}$ & AP & AP$_{50}$ & AP$_{75}$ \\
\midrule
SiamLayout~\cite{creatilayout2025} & 41.74 & 17.37 & 87.70 & 54.42 & 0.0 & 0.0 & 0.0 & 0.0 & 0.1 & 0.0 & 27.9 & 65.9 & 17.0 & 33.6 & 69.9 & 26.8 \\
\rowcolor{SkyBlue!5} GeoGrain (SiamLayout-style) & \GeoGrainUnderline{11.12} & \GeoGrainUnderline{2.89} & \GeoGrainUnderline{9.21} & \textbf{5.50} & \GeoGrainUnderline{28.1} & \GeoGrainUnderline{69.2} & \GeoGrainUnderline{15.6} & \GeoGrainUnderline{31.4} & \GeoGrainUnderline{69.7} & \GeoGrainUnderline{21.7} & \GeoGrainUnderline{30.7} & \GeoGrainUnderline{69.3} & \GeoGrainUnderline{20.8} & \GeoGrainUnderline{34.6} & \GeoGrainUnderline{70.8} & \GeoGrainUnderline{28.4} \\
\rowcolor{RoyalPurple!5} \textbf{GeoGrain (Laytrol-style)} & \textbf{10.24} & \textbf{2.71} & \textbf{8.85} & \GeoGrainUnderline{5.52} & \textbf{29.6} & \textbf{71.1} & \textbf{17.3} & \textbf{33.2} & \textbf{71.9} & \textbf{24.3} & \textbf{30.8} & \textbf{69.4} & \textbf{20.9} & \textbf{34.7} & \textbf{70.9} & \textbf{28.9} \\
\bottomrule
\end{tabularx}
\par\vskip6.0pt\relax 
\end{table*}
}
\section{Additional Ablation Studies}
\label{supp:ablations}
This section extends the ablations in \cref{sec:ablations}. We examine GeoGrain's compatibility with an alternative dual-branch DiT architecture. We also analyze token expansion, positional encoding, and decoding for mixed-resolution generation. Finally, we evaluate the trainability of GeoGrain-generated images under two controlled settings: a fixed image count and a fixed iteration count. Throughout this section, DCFL and Oriented R-CNN refer to DCFL with Rotated RetinaNet (ResNet-50) and Oriented R-CNN (Swin-T), respectively (\cref{supp:detectors}).

\GeoGrainDeferFloat{siamfigure}{%
\begin{figure}[t]
\centering
\iflatexml
\includegraphics[width=2\textwidth,alt={SiamLayout and SiamLayout-style GeoGrain outputs for the same SODA-A layout, with input oriented boxes and enlarged red-boxed regions.}]{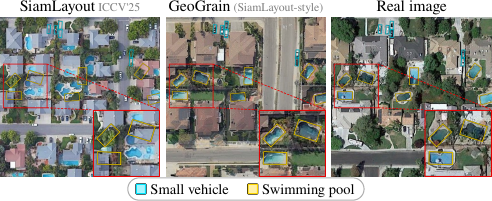}
\else
\includegraphics[width=\linewidth]{figures_supp/qualitative_siam.pdf}
\fi
\caption{\textbf{Qualitative comparison between SiamLayout and SiamLayout-style GeoGrain on the SODA-A dataset.} Both methods use the same input layout. Input OBBs are overlaid, and red-boxed regions are enlarged in the insets.}
\label{fig:supp_siamlayout}
\par\vskip3.65pt\relax 
\end{figure}
}

\subsection{Dual-branch Architecture Compatibility}
\label{supp:architecture}

The ablation in \cref{sec:ablations} shows that a dual-branch
DiT architecture alone does not ensure precise small-object control:
the baseline without GAGE and GATE achieves only 0.5 AP. This motivates
examining whether GeoGrain's granularity-aware components remain
effective when paired with an alternative dual-branch architecture.\looseness=-1

\GeoGrainDeferFloat{tokenexpansion}{\begin{table*}[t]
\centering
\caption{\textbf{Ablation results on Token expansion.} KID and SceneKID are scaled by $10^3$. The best and second-best results are \textbf{bold} and \GeoGrainUnderline{underlined}, respectively.}
\label{tab:supp_gate_grid}
\label{tab:supp_granularity}
\GeoGrainTableFont{\fontsize{8.0}{9.5}\selectfont}
\setlength{\tabcolsep}{2pt}
\renewcommand{\arraystretch}{1.15}
\renewcommand{\tabularxcolumn}[1]{m{#1}}
\begin{tabularx}{\textwidth}{l*{2}{Z}*{2}{>{\hsize=1.35\hsize\linewidth=\hsize}Z}*{6}{>{\hsize=.85\hsize\linewidth=\hsize}Z}*{2}{>{\hsize=1.1\hsize\linewidth=\hsize}Z}}
\toprule
\multirow{3}{*}{\textbf{Variant}} & \multicolumn{4}{c}{\textbf{Generation quality} $\downarrow$} & \multicolumn{6}{c}{\textbf{Layout fidelity} $\uparrow$} & \multicolumn{2}{c}{\textbf{Inference cost} $\downarrow$} \\
\cmidrule(lr){2-5}\cmidrule(lr){6-11}\cmidrule(l){12-13}
 & \multirow{2}{*}{FID} & \multirow{2}{*}{KID} & \multirow{2}{*}{SceneFID} & \multirow{2}{*}{SceneKID} & \multicolumn{3}{c}{DCFL} & \multicolumn{3}{c}{Oriented R-CNN} & \multirow{2}{*}{\# Tokens} & \multirow{2}{*}{\makecell{Time\\(s/image)}} \\
\cmidrule(lr){6-8}\cmidrule(lr){9-11}
 & & & & & AP & AP$_{50}$ & AP$_{75}$ & AP & AP$_{50}$ & AP$_{75}$ & & \\
\midrule
w/o GATE (uniform $1\times$) & \GeoGrainUnderline{10.31} & \textbf{2.41} & \GeoGrainUnderline{12.18} & \GeoGrainUnderline{6.20} & 21.7 & 59.5 & 9.0 & 23.1 & 59.3 & 11.6 & \textbf{1,024} & \textbf{3.89} \\
w/o GATE (uniform $2\times$) & 17.15 & 5.97 & 17.08 & 10.73 & \GeoGrainUnderline{25.5} & \GeoGrainUnderline{64.7} & \GeoGrainUnderline{12.9} & \GeoGrainUnderline{28.3} & \GeoGrainUnderline{65.7} & \GeoGrainUnderline{18.3} & 4,096 & 12.29 \\
\rowcolor{RoyalPurple!5}\textbf{w/ GATE (mixed resolution)} & \textbf{10.24} & \GeoGrainUnderline{2.71} & \textbf{8.85} & \textbf{5.52} & \textbf{29.6} & \textbf{71.1} & \textbf{17.3} & \textbf{33.2} & \textbf{71.9} & \textbf{24.3} & \GeoGrainUnderline{1,189} & \GeoGrainUnderline{5.18} \\
\bottomrule
\end{tabularx}
\par\vskip7.2pt\relax 
\end{table*}
}
To this end, we construct a SiamLayout-style GeoGrain variant by
adopting SiamLayout's~\cite{creatilayout2025} dual-branch architecture
while keeping the remaining GeoGrain components, including GAGE
and GATE, unchanged, and evaluate it on SODA-A. \Cref{tab:supp_architecture}
reports generation quality, layout fidelity, and trainability for
the SiamLayout baseline and the two GeoGrain variants. The
SiamLayout-style variant achieves 28.1 AP with DCFL and 31.4 AP
with Oriented R-CNN, whereas the SiamLayout baseline obtains
0.0 AP with both detectors, indicating that the baseline does not
reliably follow the input OBBs.

\GeoGrainShowFloat{architecture}

The SiamLayout-style variant also achieves trainability close to
that of the base Laytrol-style GeoGrain model, with AP scores of
30.7 and 34.6 for DCFL and Oriented R-CNN, respectively, compared
with 30.8 and 34.7 for the base model. The qualitative comparison
in \cref{fig:supp_siamlayout} further illustrates this contrast:
under the same input layouts, objects in images generated by the
SiamLayout baseline are visibly misaligned with their specified
OBBs, whereas the SiamLayout-style variant places the small vehicles
and swimming pools accordingly. The enlarged
insets further highlight these placement differences. Overall,
these results demonstrate that granularity-aware
components of GeoGrain can be successfully combined with an alternative dual-branch 
architecture while retaining effective layout control and
trainability.

\GeoGrainShowFloat{siamfigure}

\subsection{Token Expansion}
\label{supp:granularity}
To assess the benefit of GATE's token expansion, we compare it with uniform image-token grids on SODA-A, retaining GAGE in all variants. Without GATE, the uniform $1\times$ and $2\times$ variants are trained and sampled at $512\times512$ and $1024\times1024$, respectively. Uniform $2\times$ outputs are Lanczos-downsampled to $512\times512$ for evaluation.

\GeoGrainShowFloat{tokenexpansion}

\Cref{tab:supp_gate_grid} shows that uniform $2\times$ improves all AP metrics over uniform $1\times$, whereas all four generation quality metrics favor uniform $1\times$. GATE maintains global image quality comparable to uniform $1\times$ while improving SceneFID, SceneKID, and all AP metrics over both uniform variants. Relative to uniform $2\times$, it also improves FID and KID and raises AP by 4.1 points with DCFL and 4.9 points with Oriented R-CNN. These results support GATE's mixed-resolution design, which matches the fine-token resolution of uniform $2\times$ in grid cells overlapping input OBBs while retaining coarse tokens \mbox{elsewhere.}\looseness=1

To quantify GATE's inference cost, we report the mean image-token count (\#~Tokens) and measure inference time following \cref{supp:runtime}. GATE uses more image tokens and takes longer than uniform $1\times$, but reduces the image-token count by 71.0\% and inference time by 57.9\% relative to uniform $2\times$. GATE achieves better generation quality and layout fidelity than uniform $2\times$ at lower inference cost, demonstrating the efficiency of its mixed-resolution design.

\GeoGrainDeferFloat{position}{\begin{table}[t]
\centering
\caption{\textbf{Ablation results on Mixed-resolution positional encoding.} KID and SceneKID are scaled by $10^3$.}
\label{tab:supp_pma_ablation}
\label{tab:supp_position}
\GeoGrainTableFont{\fontsize{8.0}{9.5}\selectfont}
\setlength{\tabcolsep}{1.5pt}
\renewcommand{\arraystretch}{1.1}
\renewcommand{\tabularxcolumn}[1]{m{#1}}
\begin{tabularx}{\columnwidth}{l<{\hspace{5pt}}*{2}{>{\hsize=.75\hsize\linewidth=\hsize}Z}*{2}{>{\hsize=1.25\hsize\linewidth=\hsize}Z}>{\hsize=.8\hsize\linewidth=\hsize}Z>{\hsize=1.2\hsize\linewidth=\hsize}Z}
\toprule
\multirow{2}{*}{\textbf{Variant}} & \multicolumn{4}{c}{\textbf{Generation quality} $\downarrow$} & \multicolumn{2}{c}{\makecell{\textbf{Layout fidelity}\\(AP) $\uparrow$}} \\
\cmidrule(lr){2-5}\cmidrule(l){6-7}
 & FID & KID & \makecell{Scene\\FID} & \makecell{Scene\\KID} & DCFL & \makecell{Oriented\\R-CNN} \\
\midrule
\rowcolor{gray!5}\multicolumn{7}{l}{\textbf{SODA-A}} \\
w/ PMA & 10.38 & \textbf{2.56} & 11.90 & 7.79 & 28.4 & 32.2 \\
\rowcolor{RoyalPurple!5}\textbf{w/o PMA (Ours)} & \textbf{10.24} & 2.71 & \textbf{8.85} & \textbf{5.52} & \textbf{29.6} & \textbf{33.2} \\
\midrule
\rowcolor{gray!5}\multicolumn{7}{l}{\textbf{AI-TOD-R}} \\
w/ PMA & 20.30 & 5.24 & 8.50 & 4.61 & \textbf{9.0} & 10.9 \\
\rowcolor{RoyalPurple!5}\textbf{w/o PMA (Ours)} & \textbf{19.54} & \textbf{4.72} & \textbf{4.02} & \textbf{1.81} & \textbf{9.0} & \textbf{11.2} \\
\bottomrule
\end{tabularx}
\par\vskip-12.2pt\relax 
\end{table}
}
\GeoGrainDeferFloat{decoding}{\begin{table*}[t]
\centering
\caption{\textbf{Ablation results on Mixed-resolution decoding.} KID and SceneKID are scaled by $10^3$. The best and second-best results are \textbf{bold} and \GeoGrainUnderline{underlined}, respectively.}
\label{tab:supp_decoding_ablation}
\label{tab:supp_decoding}
\GeoGrainTableFont{\fontsize{8.0}{9.5}\selectfont}
\setlength{\tabcolsep}{3pt}
\renewcommand{\arraystretch}{1.15}
\begin{tabularx}{\textwidth}{l*{2}{>{\hsize=.8\hsize\linewidth=\hsize}Z}*{2}{>{\hsize=1.2\hsize\linewidth=\hsize}Z}*{3}{>{\hsize=.8\hsize\linewidth=\hsize}Z}*{2}{>{\hsize=1.3\hsize\linewidth=\hsize}Z}}
\toprule
\multirow{2}{*}{\textbf{Output}} & \multicolumn{4}{c}{\textbf{Generation quality} $\downarrow$} & \multicolumn{3}{c}{\textbf{Layout fidelity} $\uparrow$} & \multicolumn{2}{c}{\textbf{Time (s/image)} $\downarrow$} \\
\cmidrule(lr){2-5}\cmidrule(lr){6-8}\cmidrule(l){9-10}
 & FID & KID & SceneFID & SceneKID & AP & AP$_{50}$ & AP$_{75}$ & Decoding & \makecell{Overall\\inference} \\
\midrule
\rowcolor{gray!5}\multicolumn{10}{l}{\textbf{SODA-A}} \\
Initial reconstruction $\hat{\mathbf{x}}^{\mathrm{init}}$ & 12.20 & 3.98 & 54.62 & 41.54 & \GeoGrainUnderline{12.8} & 33.2 & 6.7 & \textbf{0.02} & \textbf{4.97} \\
Fine reconstruction $\hat{\mathbf{x}}^{\mathrm{f}}$ & \GeoGrainUnderline{10.51} & \GeoGrainUnderline{2.90} & \GeoGrainUnderline{9.18} & \GeoGrainUnderline{5.89} & \textbf{33.2} & \textbf{72.2} & \GeoGrainUnderline{24.2} & \GeoGrainUnderline{0.19} & \GeoGrainUnderline{5.13} \\
\rowcolor{RoyalPurple!5} \textbf{Final image} $\hat{\mathbf{x}}$ & \textbf{10.24} & \textbf{2.71} & \textbf{8.85} & \textbf{5.52} & \textbf{33.2} & \GeoGrainUnderline{71.9} & \textbf{24.3} & 0.23 & 5.18 \\
\midrule
\rowcolor{gray!5}\multicolumn{10}{l}{\textbf{AI-TOD-R}} \\
Initial reconstruction $\hat{\mathbf{x}}^{\mathrm{init}}$ & 22.46 & 7.10 & \GeoGrainUnderline{52.13} & 39.07 & 4.8 & 16.2 & 0.9 & \textbf{0.02} & \textbf{4.76} \\
Fine reconstruction $\hat{\mathbf{x}}^{\mathrm{f}}$ & \GeoGrainUnderline{20.80} & \GeoGrainUnderline{5.02} & \textbf{4.02} & \GeoGrainUnderline{1.82} & \GeoGrainUnderline{11.0} & \GeoGrainUnderline{34.9} & \textbf{3.0} & \GeoGrainUnderline{0.19} & \GeoGrainUnderline{4.93} \\
\rowcolor{RoyalPurple!5} \textbf{Final image} $\hat{\mathbf{x}}$ & \textbf{19.54} & \textbf{4.72} & \textbf{4.02} & \textbf{1.81} & \textbf{11.2} & \textbf{35.7} & \GeoGrainUnderline{2.9} & 0.23 & 4.97 \\
\bottomrule
\end{tabularx}
\par\vskip-4.6pt\relax 
\end{table*}}
\subsection{Mixed-resolution Positional Encoding}
\label{supp:position}
On GATE's mixed-resolution grid, GeoGrain uses RoPE position indices defined on the fine-token grid for both coarse and fine image tokens (\cref{supp:geograin_impl}). Each coarse token takes the position of the top-left fine cell in its region. Each layout token uses the same RoPE position indices as its associated image token (\cref{sec:gage}). An alternative is Phase-Aligned Mixed-Resolution Attention (PMA)~\cite{pma2025}, which rescales key positions to the native resolution of each query.

We train GeoGrain on SODA-A and AI-TOD-R with and without applying PMA to all attention layers of the image--layout branch $\mathcal{B}_L$ and the image--text branch $\mathcal{B}_T$. \Cref{tab:supp_pma_ablation} shows that GeoGrain without PMA achieves lower SceneFID and SceneKID on both datasets, with equal or higher AP for both detectors. SceneFID decreases from 11.90 to 8.85 on SODA-A and from 8.50 to 4.02 on AI-TOD-R, while global image quality remains comparable. These results support using fine-grid RoPE positions for GeoGrain's mixed-resolution generation under this training setup.

\GeoGrainShowFloat{position}

\subsection{Mixed-resolution Decoding}
\label{supp:decoding}
To evaluate the mixed-resolution decoding procedure in \cref{sec:gate}, we compare the initial reconstruction $\hat{\mathbf{x}}^{\mathrm{init}}$, fine reconstruction $\hat{\mathbf{x}}^{\mathrm{f}}$, and final image $\hat{\mathbf{x}}$ on SODA-A and AI-TOD-R, with Oriented R-CNN. All three outputs are obtained from the same denoised tokens $\hat{\mathbf{z}}_0$.

\GeoGrainShowFloat{decoding}

The initial reconstruction uses a coarse grid with pooled fine tokens, whereas the fine reconstruction retains the original denoised fine tokens. \Cref{tab:supp_decoding_ablation} shows that the fine reconstruction substantially improves SceneFID, SceneKID, and all AP metrics over the initial reconstruction, with AP gains of 20.4 and 6.2 points on SODA-A and AI-\mbox{TOD-R,} respectively. The fine reconstruction thus accounts for most of the improvements in object-level generation quality and layout fidelity. The final image $\hat{\mathbf{x}}$ combines the fine reconstruction $\hat{\mathbf{x}}^{\mathrm{f}}$ in the grid cells selected by GATE and the coarse \mbox{reconstruction} $\hat{\mathbf{x}}^{\mathrm{c}}$ elsewhere using Laplacian-pyramid blending. This further improves global image quality, reducing FID and KID on both datasets while maintaining comparable SceneFID, SceneKID, and layout fidelity.

To assess the cost of these reconstruction steps, we measure inference time following \cref{supp:runtime}. Producing the final image adds 0.04--0.05\,s per image to overall inference time relative to the fine reconstruction. The complete decoding procedure takes 0.23\,s per image, accounting for less than 5\% of overall inference time on both datasets. We therefore use the final decoding procedure to improve global image quality while maintaining comparable object-level generation quality and layout fidelity at a small additional inference cost.\looseness=-1

\GeoGrainDeferFloat{fixedimages}{\begin{table}[t]
\vskip6.9pt\relax 
\centering
\caption{\textbf{Ablation results on Trainability at fixed image count.} The best and second-best results are \textbf{bold} and \GeoGrainUnderline{underlined}, respectively.}
\label{tab:supp_synthetic_only}
\label{tab:supp_synthetic}
\GeoGrainTableFont{\fontsize{8.0}{9.5}\selectfont}
\setlength{\tabcolsep}{2pt}
\renewcommand{\arraystretch}{1.15}
\renewcommand{\tabularxcolumn}[1]{m{#1}}
\begin{tabularx}{\columnwidth}{l*{6}{Z}}
\toprule
\multirow{3}{*}{\textbf{Training data}} & \multicolumn{6}{c}{\textbf{Trainability} $\uparrow$} \\
\cmidrule(l){2-7}
 & \multicolumn{3}{c}{DCFL} & \multicolumn{3}{c}{Oriented R-CNN} \\
\cmidrule(lr){2-4}\cmidrule(l){5-7}
 & AP & AP$_{50}$ & AP$_{75}$ & AP & AP$_{50}$ & AP$_{75}$ \\
\midrule
\rowcolor{gray!5}\multicolumn{7}{l}{\textbf{SODA-A} (23,032 training images)} \\
Real 100\% & \textbf{29.0} & \textbf{67.4} & \textbf{18.3} & \textbf{33.8} & \textbf{70.0} & \textbf{27.3} \\
Synthetic 100\% & 25.7 & 61.6 & \GeoGrainUnderline{15.1} & 21.3 & 49.9 & 13.9 \\
Real 50\% + Synthetic 50\% & \GeoGrainUnderline{28.8} & \GeoGrainUnderline{66.8} & \textbf{18.3} & \GeoGrainUnderline{33.1} & \GeoGrainUnderline{69.4} & \GeoGrainUnderline{26.3} \\
\midrule
\rowcolor{gray!5}\multicolumn{7}{l}{\textbf{AI-TOD-R} (7,752 training images)} \\
Real 100\% & \textbf{9.1} & \textbf{29.9} & \textbf{2.4} & \textbf{10.5} & \textbf{33.0} & \textbf{3.0} \\
Synthetic 100\% & 7.8 & 25.7 & \GeoGrainUnderline{2.2} & 5.8 & 18.6 & \GeoGrainUnderline{1.7} \\
Real 50\% + Synthetic 50\% & \GeoGrainUnderline{8.7} & \GeoGrainUnderline{28.5} & \textbf{2.4} & \GeoGrainUnderline{10.0} & \GeoGrainUnderline{31.2} & \textbf{3.0} \\
\bottomrule
\end{tabularx}
\end{table}}
\GeoGrainDeferFloat{fixediterations}{\begin{table*}[t]
\centering
\caption{\textbf{Ablation results on Trainability at fixed iteration count.} The best and second-best results are \textbf{bold} and \GeoGrainUnderline{underlined}, \mbox{respectively.}}
\label{tab:supp_iter_detaug}
\label{tab:supp_updates}
\GeoGrainTableFont{\fontsize{8.0}{9.5}\selectfont}
\setlength{\tabcolsep}{2pt}
\renewcommand{\arraystretch}{1.15}
\renewcommand{\tabularxcolumn}[1]{m{#1}}
\begin{tabularx}{\textwidth}{l<{\hspace{6pt}}cc*{12}{Z}}
\toprule
\multirow{4}{*}{\textbf{Training data}} & \multirow{4}{*}{\textbf{\# Images}} & \multirow{4}{*}{\textbf{Epochs}} & \multicolumn{12}{c}{\textbf{Trainability} $\uparrow$} \\
\cmidrule(l){4-15}
 & & & \multicolumn{6}{c}{DCFL} & \multicolumn{6}{c}{Oriented R-CNN} \\
\cmidrule(lr){4-9}\cmidrule(l){10-15}
 & & & \multicolumn{3}{c}{w/ detector aug.} & \multicolumn{3}{c}{w/o detector aug.} & \multicolumn{3}{c}{w/ detector aug.} & \multicolumn{3}{c}{w/o detector aug.} \\
\cmidrule(lr){4-6}\cmidrule(lr){7-9}\cmidrule(lr){10-12}\cmidrule(l){13-15}
 & & & AP & AP$_{50}$ & AP$_{75}$ & AP & AP$_{50}$ & AP$_{75}$ & AP & AP$_{50}$ & AP$_{75}$ & AP & AP$_{50}$ & AP$_{75}$ \\
\midrule
\rowcolor{gray!5}\multicolumn{15}{l}{\textbf{SODA-A}} \\
Real 100\% & 23,032 & 12 & 29.0 & 67.4 & 18.3 & 28.6 & \GeoGrainUnderline{65.8} & 19.2 & 33.8 & 70.0 & \GeoGrainUnderline{27.3} & \GeoGrainUnderline{32.9} & \GeoGrainUnderline{67.1} & \GeoGrainUnderline{27.9} \\
Real 100\% & 23,032 & 24 & \GeoGrainUnderline{30.4} & \GeoGrainUnderline{69.0} & \GeoGrainUnderline{20.4} & \GeoGrainUnderline{29.2} & \GeoGrainUnderline{65.8} & \GeoGrainUnderline{20.4} & \GeoGrainUnderline{33.9} & \GeoGrainUnderline{70.4} & 27.1 & 32.0 & 65.6 & 27.0 \\
Real 100\% + Synthetic 100\% & 23,032 + 23,032 & 12 & \textbf{30.8} & \textbf{69.4} & \textbf{20.9} & \textbf{31.1} & \textbf{68.5} & \textbf{22.5} & \textbf{34.7} & \textbf{70.9} & \textbf{28.9} & \textbf{34.0} & \textbf{68.2} & \textbf{29.8} \\
\addlinespace[4.8pt]
Real 50\% & 11,516 & 12 & 26.1 & 63.0 & 14.9 & 25.8 & 61.8 & 15.4 & 32.2 & \GeoGrainUnderline{68.9} & 24.7 & \GeoGrainUnderline{31.3} & \GeoGrainUnderline{65.6} & \GeoGrainUnderline{25.4} \\
Real 50\% & 11,516 & 24 & \GeoGrainUnderline{28.4} & \GeoGrainUnderline{66.7} & \GeoGrainUnderline{17.8} & \GeoGrainUnderline{26.4} & \GeoGrainUnderline{62.2} & \GeoGrainUnderline{16.6} & \GeoGrainUnderline{32.8} & 68.8 & \GeoGrainUnderline{26.2} & 30.6 & 64.1 & 24.7 \\
Real 50\% + Synthetic 50\% & 11,516 + 11,516 & 12 & \textbf{28.8} & \textbf{66.8} & \textbf{18.3} & \textbf{28.0} & \textbf{64.8} & \textbf{18.4} & \textbf{33.1} & \textbf{69.4} & \textbf{26.3} & \textbf{32.1} & \textbf{66.4} & \textbf{26.4} \\
\midrule
\rowcolor{gray!5}\multicolumn{15}{l}{\textbf{AI-TOD-R}} \\
Real 100\% & 7,752 & 12 & 9.1 & 29.9 & 2.4 & 9.0 & 29.2 & 2.8 & 10.5 & \GeoGrainUnderline{33.0} & 3.0 & \GeoGrainUnderline{9.3} & \GeoGrainUnderline{28.9} & 2.8 \\
Real 100\% & 7,752 & 24 & \GeoGrainUnderline{10.9} & \GeoGrainUnderline{35.3} & \GeoGrainUnderline{3.2} & \GeoGrainUnderline{9.7} & \GeoGrainUnderline{32.3} & \GeoGrainUnderline{2.9} & \GeoGrainUnderline{10.9} & 32.9 & \GeoGrainUnderline{3.8} & 8.7 & 26.8 & \textbf{3.3} \\
Real 100\% + Synthetic 100\% & 7,752 + 7,752 & 12 & \textbf{11.2} & \textbf{35.8} & \textbf{3.3} & \textbf{10.7} & \textbf{34.4} & \textbf{3.2} & \textbf{11.0} & \textbf{33.2} & \textbf{3.9} & \textbf{9.7} & \textbf{30.5} & \GeoGrainUnderline{3.0} \\
\addlinespace[4.8pt]
Real 50\% & 3,876 & 12 & 5.7 & 19.5 & 1.3 & 6.0 & 20.5 & 1.6 & 9.1 & 28.9 & 2.2 & \GeoGrainUnderline{7.5} & \GeoGrainUnderline{24.3} & 2.0 \\
Real 50\% & 3,876 & 24 & \GeoGrainUnderline{8.3} & \GeoGrainUnderline{27.8} & \GeoGrainUnderline{2.2} & \GeoGrainUnderline{7.9} & \GeoGrainUnderline{26.3} & \GeoGrainUnderline{2.0} & \GeoGrainUnderline{9.5} & \GeoGrainUnderline{29.4} & \GeoGrainUnderline{2.7} & 7.1 & 22.5 & \GeoGrainUnderline{2.4} \\
Real 50\% + Synthetic 50\% & 3,876 + 3,876 & 12 & \textbf{8.7} & \textbf{28.5} & \textbf{2.4} & \textbf{8.7} & \textbf{29.3} & \textbf{2.1} & \textbf{10.0} & \textbf{31.2} & \textbf{3.0} & \textbf{8.2} & \textbf{25.9} & \textbf{2.5} \\
\bottomrule
\end{tabularx}
\par\vskip9.7pt\relax 
\end{table*}
}
\subsection{Trainability at Fixed Image Count}
\label{supp:synthetic}
To assess whether GeoGrain-generated images can replace part of the real training data while preserving detector performance, we compare real-only, synthetic-only, and 50:50 real--synthetic training on SODA-A and AI-TOD-R. For each dataset, all three training sets contain the same number of images as the full real training split.

\Cref{tab:supp_synthetic_only} shows that synthetic-only training yields lower AP than real-only training. In contrast, the 50:50 mixture approaches real-only performance, with an AP gap of at most 0.7 points across the evaluated datasets and detectors. These results support combining GeoGrain-generated images with real images for detector training, consistent with prior findings on synthetic data augmentation~\cite{geodiffusion2024}.

\GeoGrainShowFloat{fixedimages}

\subsection{Trainability at Fixed Iteration Count}
\label{supp:updates}

Prior L2I studies have evaluated trainability by augmenting real detector-training sets with generated images~\cite{geodiffusion2024,aerogen2025,ficgen2025,ofdiff2026,ccdiff2024,ccdiffpp2025} under the standard $1\times$ schedule~\cite{geodiffusion2024,ficgen2025}. Following this practice, we use the $1\times$ schedule (12 epochs) for our main trainability evaluation. We further examine whether the gains persist at a matched iteration budget by comparing mixed training under this schedule with real-only training for 24 epochs ($2\times$ schedule). We consider 50\% and 100\% of the real \mbox{training} split, adding an equal number of generated images to each mixed set.

We repeat these comparisons with and without detector augmentation. The former uses the standard MMRotate random image flips, while the latter disables them. As described in \cref{sec:setup}, synthetic images in both settings are generated from layouts taken from the real training set and transformed by simple flips and translations, following~\cite{geodiffusion2024}.

\GeoGrainShowFloat{fixediterations}

As shown in \Cref{tab:supp_iter_detaug}, mixed training improves AP over real-only training across the evaluated settings at a matched iteration budget. The gains are particularly pronounced without detector augmentation, reaching 0.8--2.0 AP points. This may partly reflect the overlap between standard detector augmentation and the simple geometric transformations applied to layouts in prior L2I studies~\cite{geodiffusion2024,aerogen2025,ficgen2025,ccdiff2024} and in our experiments. Future work could explore more diverse and realistic layout designs for generating synthetic images to augment small-object detection training data without relying solely on simple transformations of existing layouts.

\section{Additional Qualitative Results}
\label{supp:qualitative}

\begin{figure*}[t]
\centering
\iflatexml
\includegraphics[width=2\textwidth,alt={Qualitative comparison on SODA-A, examples 1 and 2. Input layouts, outputs from the compared methods and GeoGrain, and real images are shown with oriented bounding boxes and class legends.}]{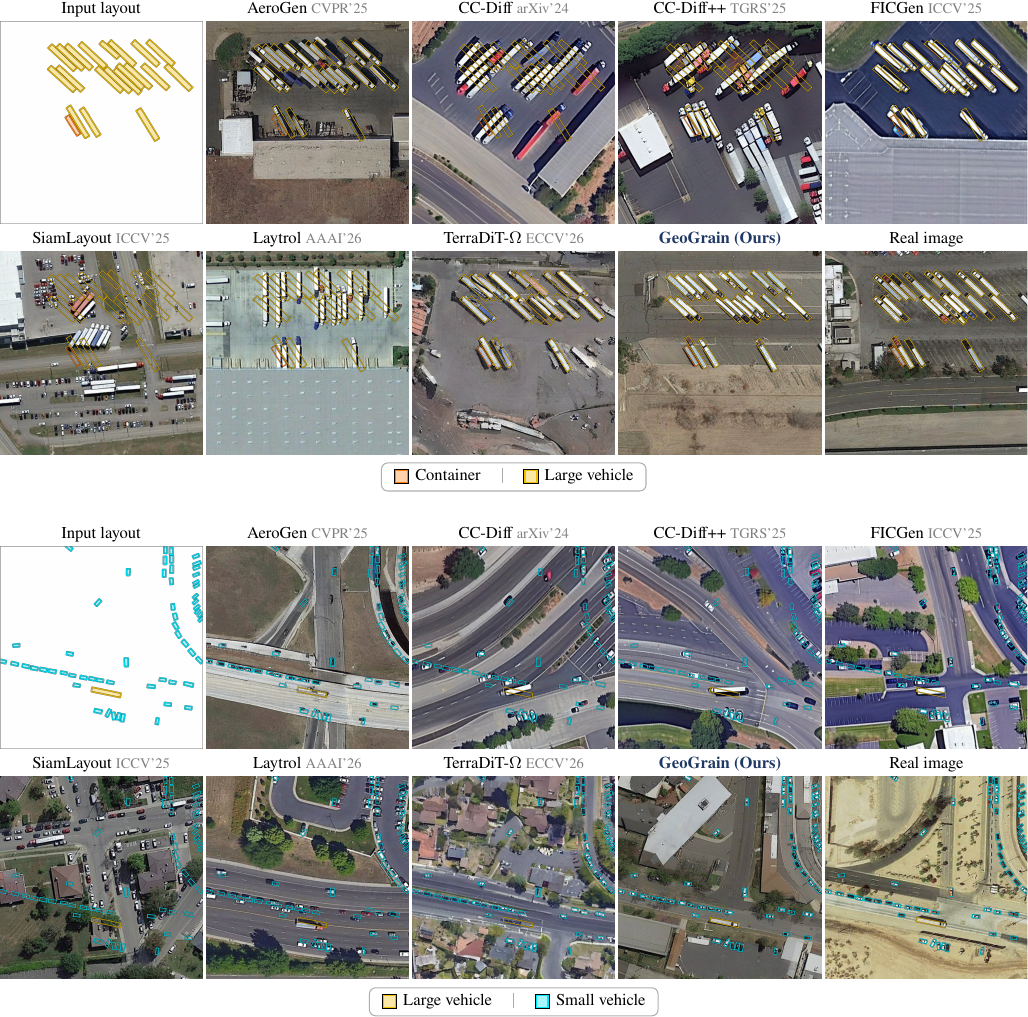}
\else
\includegraphics[width=\textwidth]{figures_supp/soda_2perpage_p1.pdf}
\fi
\caption{\textbf{Qualitative comparison on the SODA-A dataset (examples 1--2).} Input OBBs are overlaid on the generated images.}
\label{fig:supp_qual_soda}
\end{figure*}

\begin{figure*}[t]
\centering
\iflatexml
\includegraphics[width=2\textwidth,alt={Qualitative comparison on SODA-A, examples 3 and 4. Input layouts, outputs from the compared methods and GeoGrain, and real images are shown with oriented bounding boxes and class legends.}]{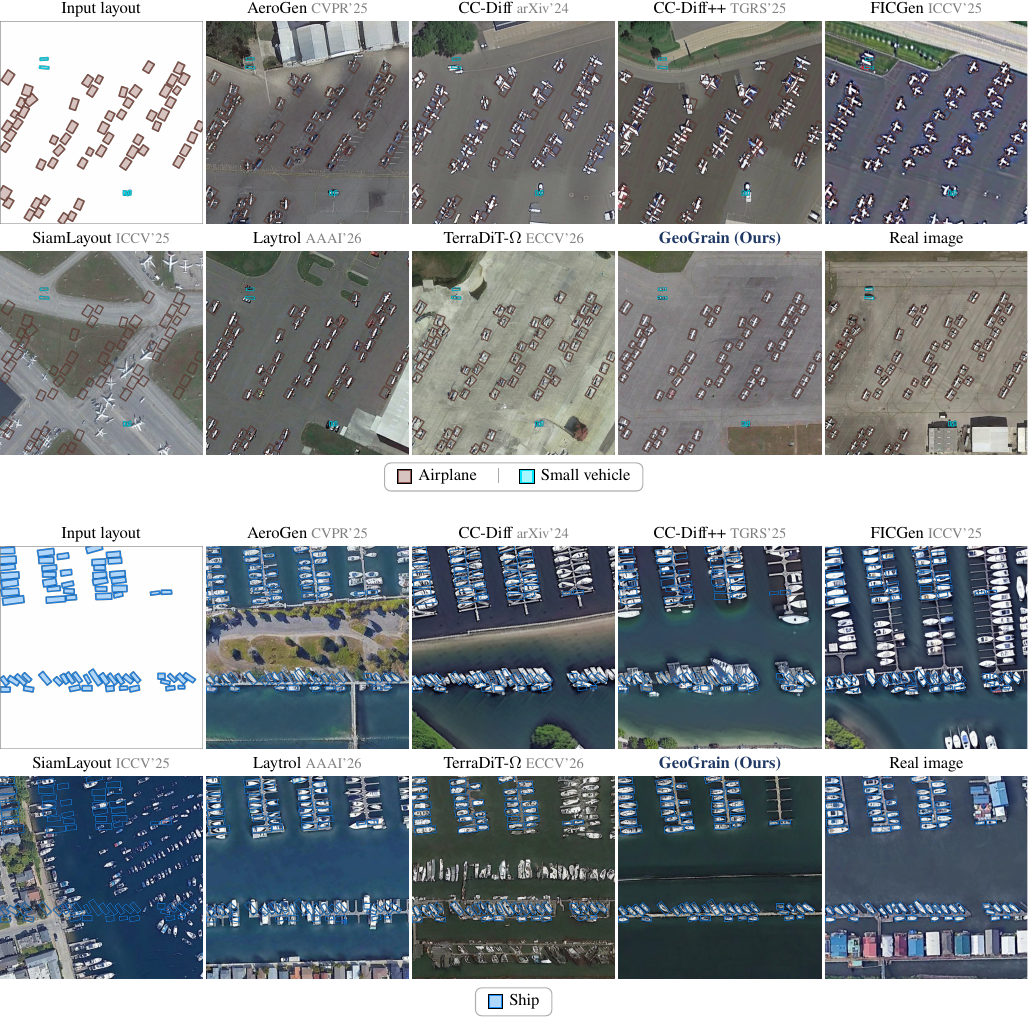}
\else
\includegraphics[width=\textwidth]{figures_supp/soda_2perpage_p2.pdf}
\fi
\caption{\textbf{Qualitative comparison on the SODA-A dataset (examples 3--4).} Input OBBs are overlaid on the generated images.}
\label{fig:supp_qual_soda2}
\end{figure*}

\begin{figure*}[t]
\centering
\iflatexml
\includegraphics[width=2\textwidth,alt={Qualitative comparison on SODA-A, examples 5 and 6. Input layouts, outputs from the compared methods and GeoGrain, and real images are shown with oriented bounding boxes and class legends.}]{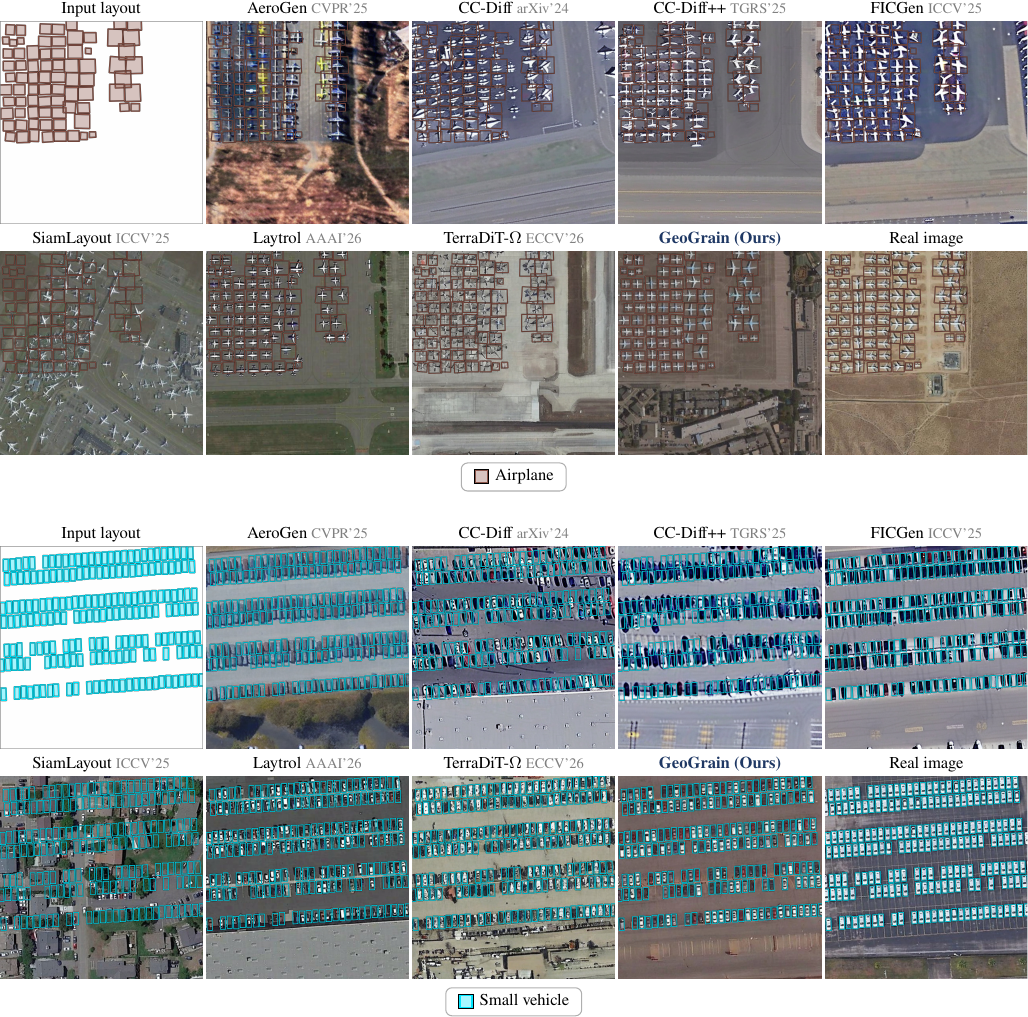}
\else
\includegraphics[width=\textwidth]{figures_supp/soda_2perpage_p3.pdf}
\fi
\caption{\textbf{Qualitative comparison on the SODA-A dataset (examples 5--6).} Input OBBs are overlaid on the generated images.}
\label{fig:supp_qual_soda3}
\end{figure*}

\begin{figure*}[t]
\centering
\iflatexml
\includegraphics[width=2\textwidth,alt={Qualitative comparison on AI-TOD-R, examples 1 and 2. Input layouts, outputs from the compared methods and GeoGrain, and real images are shown with oriented bounding boxes and class legends.}]{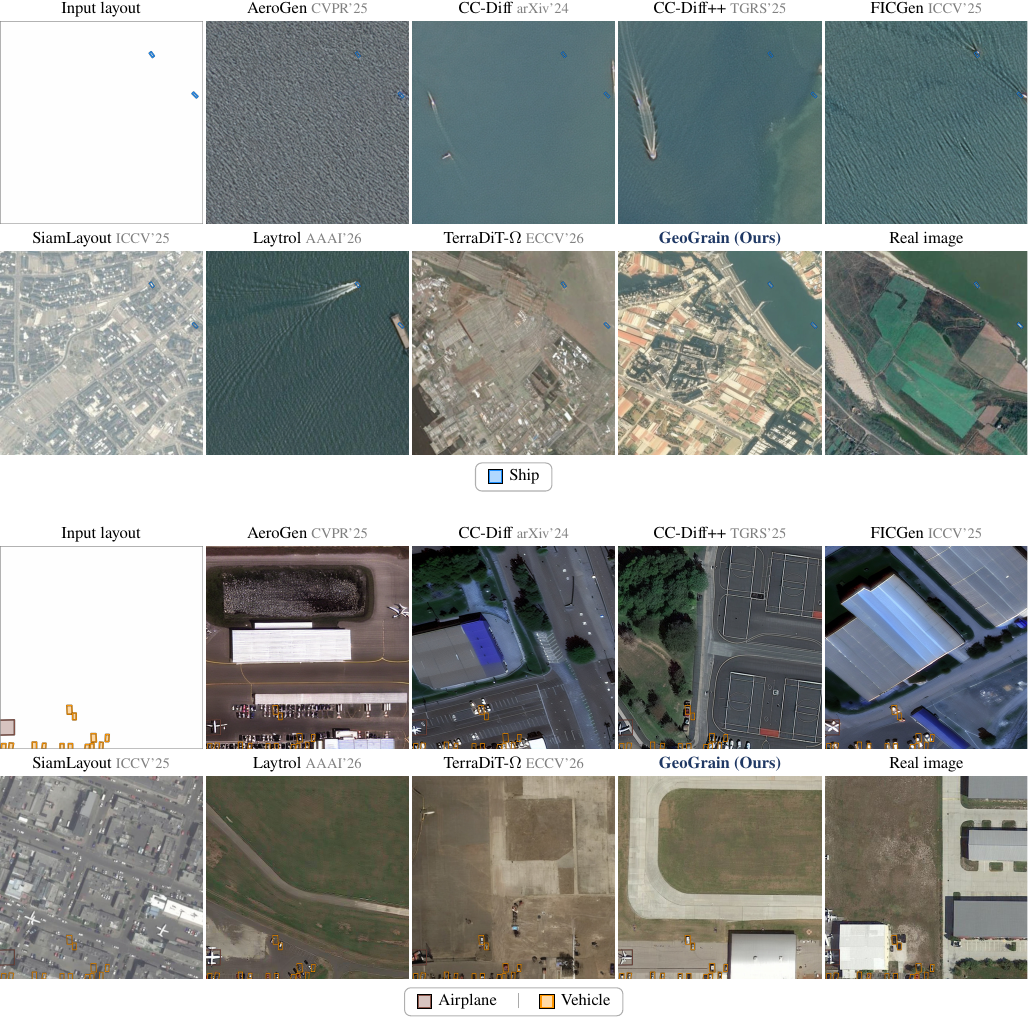}
\else
\includegraphics[width=\textwidth]{figures_supp/aitod_2perpage_p1.pdf}
\fi
\caption{\textbf{Qualitative comparison on the AI-TOD-R dataset (examples 1--2).} Input OBBs are overlaid on the generated images.}
\label{fig:supp_qual_aitod}
\end{figure*}

\begin{figure*}[t]
\centering
\iflatexml
\includegraphics[width=2\textwidth,alt={Qualitative comparison on AI-TOD-R, examples 3 and 4. Input layouts, outputs from the compared methods and GeoGrain, and real images are shown with oriented bounding boxes and class legends.}]{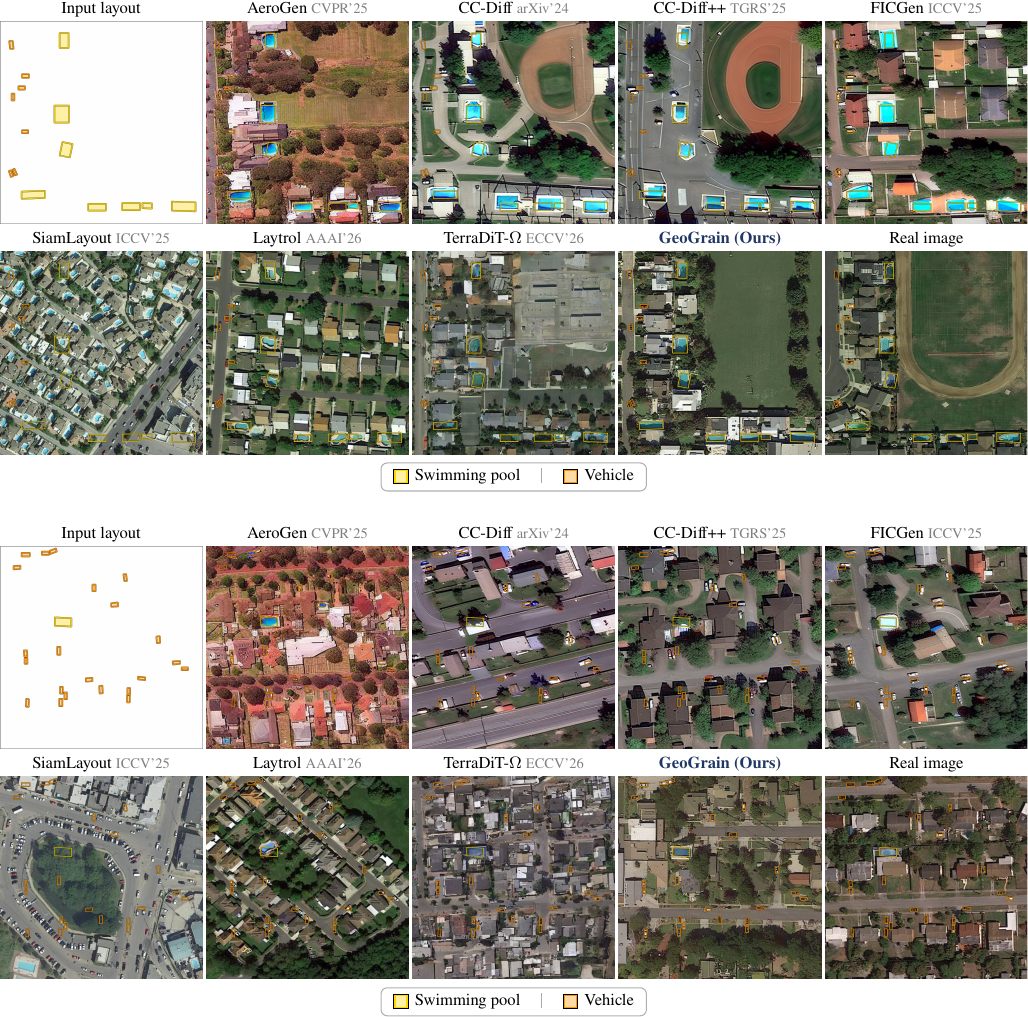}
\else
\includegraphics[width=\textwidth]{figures_supp/aitod_2perpage_p2.pdf}
\fi
\caption{\textbf{Qualitative comparison on the AI-TOD-R dataset (examples 3--4).} Input OBBs are overlaid on the generated images.}
\label{fig:supp_qual_aitod2}
\end{figure*}

\begin{figure*}[t]
\centering
\iflatexml
\includegraphics[width=2\textwidth,alt={Qualitative comparison on AI-TOD-R, examples 5 and 6. Input layouts, outputs from the compared methods and GeoGrain, and real images are shown with oriented bounding boxes and class legends.}]{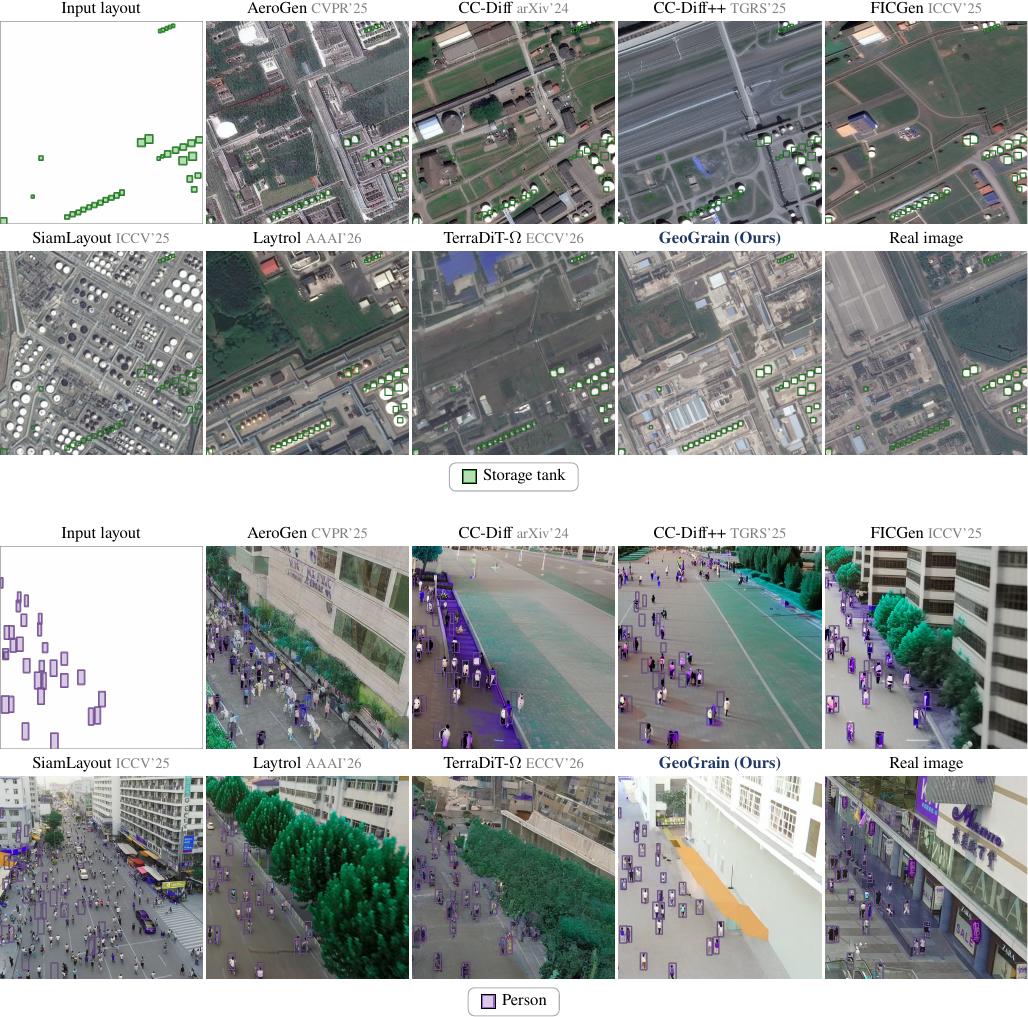}
\else
\includegraphics[width=\textwidth]{figures_supp/aitod_2perpage_p3.pdf}
\fi
\caption{\textbf{Qualitative comparison on the AI-TOD-R dataset (examples 5--6).} Input OBBs are overlaid on the generated images.}
\label{fig:supp_qual_aitod3}
\end{figure*}

\begin{figure*}[t]
\centering
\iflatexml
\includegraphics[width=2\textwidth,alt={Qualitative comparison on DIOR-R, examples 1 and 2. Input layouts, outputs from the compared methods and GeoGrain, and real images are shown with oriented bounding boxes and class legends.}]{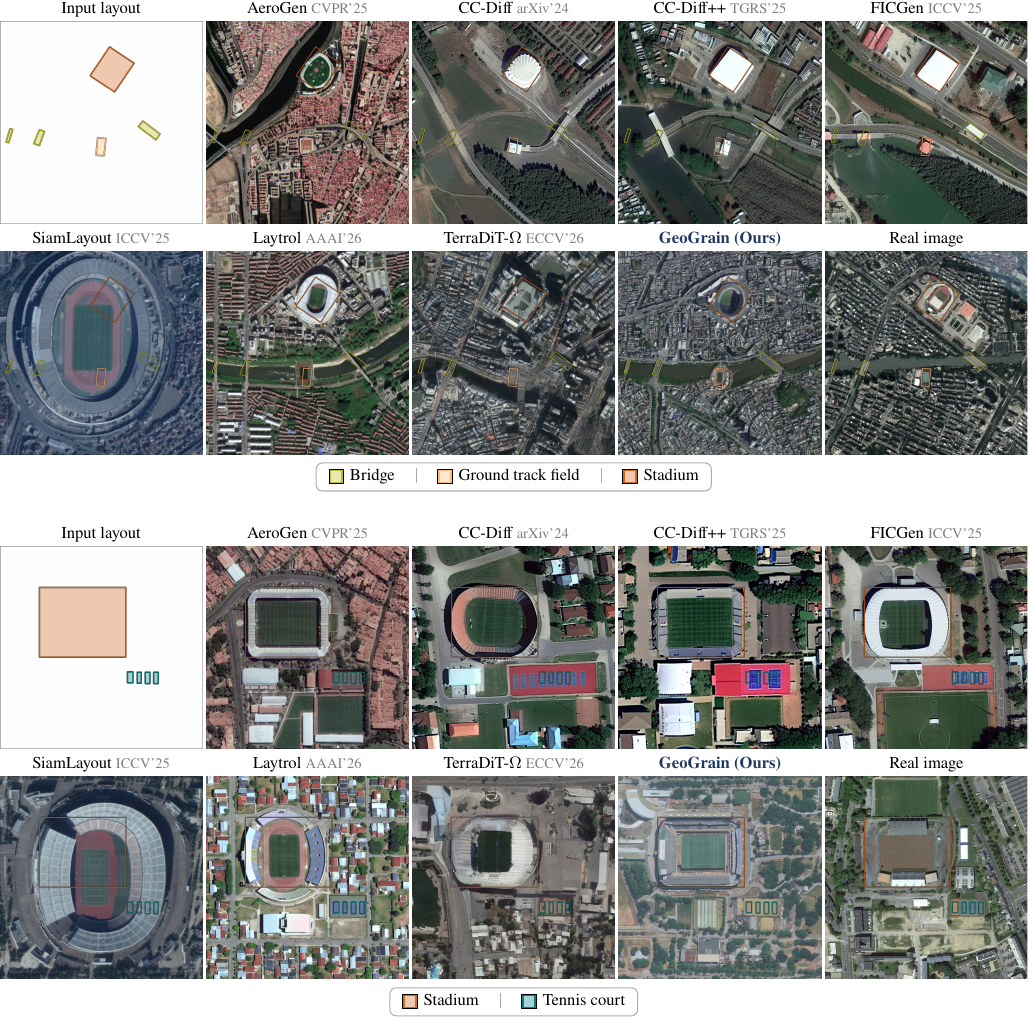}
\else
\includegraphics[width=\textwidth]{figures_supp/dior_2perpage_p1.pdf}
\fi
\caption{\textbf{Qualitative comparison on the DIOR-R dataset (examples 1--2).} Input OBBs are overlaid on the generated images.}
\label{fig:supp_qual_dior}
\end{figure*}

\begin{figure*}[t]
\centering
\iflatexml
\includegraphics[width=2\textwidth,alt={Qualitative comparison on DIOR-R, examples 3 and 4. Input layouts, outputs from the compared methods and GeoGrain, and real images are shown with oriented bounding boxes and class legends.}]{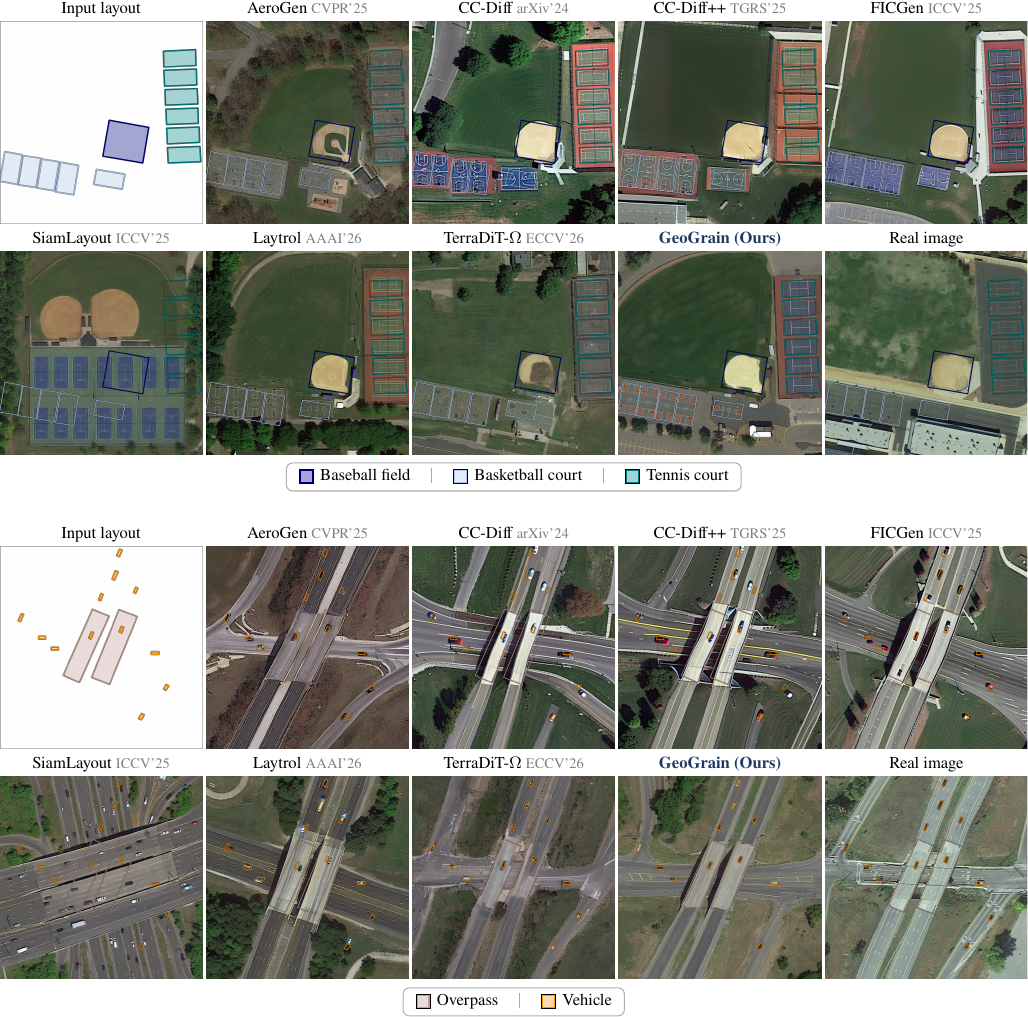}
\else
\includegraphics[width=\textwidth]{figures_supp/dior_2perpage_p2.pdf}
\fi
\caption{\textbf{Qualitative comparison on the DIOR-R dataset (examples 3--4).} Input OBBs are overlaid on the generated images.}
\label{fig:supp_qual_dior2}
\end{figure*}

\begin{figure*}[t]
\centering
\iflatexml
\includegraphics[width=2\textwidth,alt={Qualitative comparison on DIOR-R, examples 5 and 6. Input layouts, outputs from the compared methods and GeoGrain, and real images are shown with oriented bounding boxes and class legends.}]{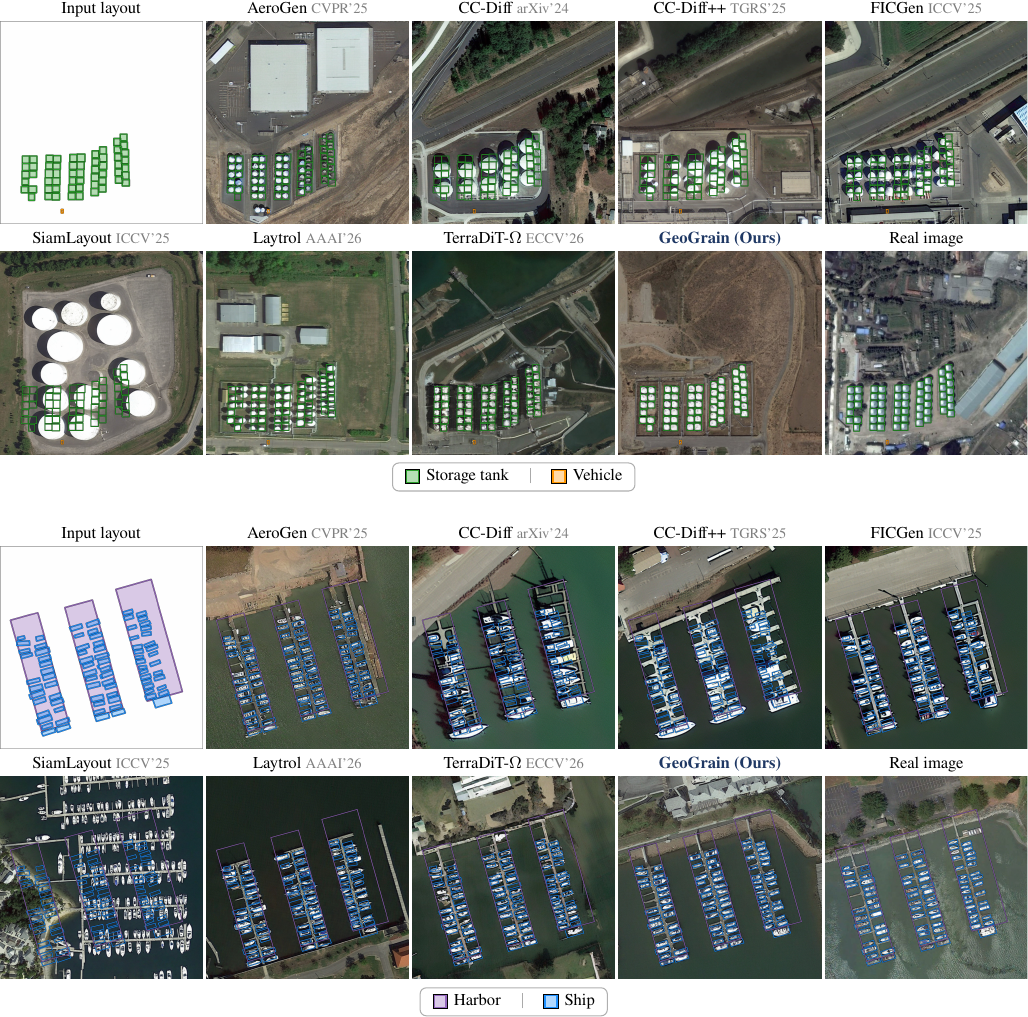}
\else
\includegraphics[width=\textwidth]{figures_supp/dior_2perpage_p3.pdf}
\fi
\caption{\textbf{Qualitative comparison on the DIOR-R dataset (examples 5--6).} Input OBBs are overlaid on the generated images.}
\label{fig:supp_qual_dior3}
\end{figure*}

We provide additional visual comparisons on SODA-A (\crefrange{fig:supp_qual_soda}{fig:supp_qual_soda3}), AI-TOD-R (\crefrange{fig:supp_qual_aitod}{fig:supp_qual_aitod3}), and DIOR-R (\crefrange{fig:supp_qual_dior}{fig:supp_qual_dior3}), with six examples per dataset. All methods use the same input OBB layout for each example, and real images are shown as scene references rather than pixel-wise reconstruction targets.

The SODA-A examples include densely arranged airplanes, small vehicles, and ships. In the crowded airplane and vehicle layouts (\cref{fig:supp_qual_soda3}), GeoGrain renders neighboring instances more distinctly and better follows their prescribed arrangement. Several baselines produce plausible surrounding scenes but leave some small objects less recognizable or less closely aligned with their input OBBs. These comparisons complement the object-level generation quality and layout fidelity results in \Cref{tab:main}.

On AI-TOD-R, the comparisons range from ships and vehicles (\cref{fig:supp_qual_aitod}) to swimming pools, storage tanks, and people (\cref{fig:supp_qual_aitod2,fig:supp_qual_aitod3}). Across these examples, GeoGrain renders recognizable small objects aligned with their prescribed OBBs, whereas several competing methods depict the specified objects less distinctly.

The DIOR-R examples (\crefrange{fig:supp_qual_dior}{fig:supp_qual_dior3}) extend the comparison to larger and more varied structures, including stadiums, sports courts, overpasses, storage tanks, and harbors. GeoGrain renders recognizable object shapes and preserves their spatial arrangements while maintaining plausible surrounding context. These examples complement the quantitative evaluation on DIOR-R in \cref{supp:dior}.
\GeoGrainDeferFloat{failurefigure}{%
\begin{figure}[t]
\centering
\iflatexml
\includegraphics[width=2\textwidth,alt={Masked FM loss limitation: red dashed boxes mark generated objects absent from the input layout, and magenta circles identify an unannotated object in the real image.}]{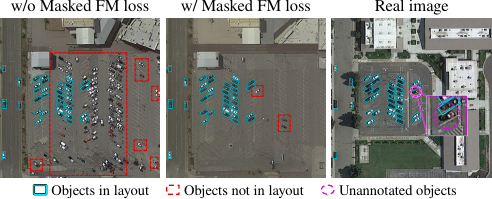}
\else
\includegraphics[width=\linewidth]{figures_supp/limitation.pdf}
\fi
\caption{\textbf{Limitation of Masked FM loss.} Red dashed boxes mark generated objects absent from the input layout; magenta circles highlight an unannotated object in the real image.}
\label{fig:supp_failure}
\par\vskip-2.09pt\relax 
\end{figure}
}

\section{Limitations and Future Work}
\label{supp:limitations}

Despite its strong overall performance in small-object L2I, GeoGrain remains computationally demanding, largely due to iterative denoising. Distilling the model to one or a few denoising steps~\cite{progressive_distillation2022,dmd2024} could reduce inference cost while maintaining performance. Our masked FM loss helps avoid generating objects absent from the input layout, although a few may still appear. Incomplete annotations may contribute to such cases, as illustrated in \cref{fig:supp_failure}, where an unannotated object appears in the real image. Automatically detecting and annotating unlabeled objects in training images could further reduce such errors. Moving beyond simple transformations of existing annotations such as flips and translations to generate diverse and realistic layouts for data augmentation, especially tailored for small objects, is also a promising avenue for \mbox{future work.}

\GeoGrainShowFloat{failurefigure}

\iflatexml
\section*{Additional Qualitative Results (continued)}
\fi

\clearpage

\clearpage

{
    \small
    \bibliographystyle{ieeenat_fullname}
    \iflatexml
\else\bibliography{main}\fi

\begin{thebibliography}{59}
\providecommand{\natexlab}[1]{#1}
\providecommand{\url}[1]{\texttt{#1}}
\expandafter\ifx\csname urlstyle\endcsname\relax
  \providecommand{\doi}[1]{doi: #1}\else
  \providecommand{\doi}{doi: \begingroup \urlstyle{rm}\Url}\fi

\bibitem{kid2018}
Miko{\l}aj Bi{\'n}kowski, Danica~J Sutherland, Michael Arbel, and Arthur
  Gretton.
\newblock Demystifying mmd gans.
\newblock \emph{arXiv preprint arXiv:1801.01401}, 2018.

\bibitem{laplacian1983}
Peter~J Burt and Edward~H Adelson.
\newblock A multiresolution spline with application to image mosaics.
\newblock \emph{ACM Transactions on Graphics (ToG)}, 2\penalty0 (4):\penalty0
  217--236, 1983.

\bibitem{foveated2026}
Brian Chao, Lior Yariv, Howard Xiao, and Gordon Wetzstein.
\newblock Foveated diffusion: Efficient spatially adaptive image and video
  generation.
\newblock \emph{arXiv preprint arXiv:2603.23491}, 2026.

\bibitem{geodiffusion2024}
Kai Chen, Enze Xie, Zhe Chen, Yibo Wang, Lanqing Hong, Zhenguo Li, and Dit-Yan
  Yeung.
\newblock Geodiffusion: Text-prompted geometric control for object detection
  data generation.
\newblock In \emph{International Conference on Learning Representations}, pages
  8979--9001, 2024.

\bibitem{soda2023}
Gong Cheng, Xiang Yuan, Xiwen Yao, Kebing Yan, Qinghua Zeng, Xingxing Xie, and
  Junwei Han.
\newblock Towards large-scale small object detection: Survey and benchmarks.
\newblock \emph{IEEE transactions on pattern analysis and machine
  intelligence}, 45\penalty0 (11):\penalty0 13467--13488, 2023.

\bibitem{sd32024}
Patrick Esser, Sumith Kulal, Andreas Blattmann, Rahim Entezari, Jonas
  M{\"u}ller, Harry Saini, Yam Levi, Dominik Lorenz, Axel Sauer, Frederic
  Boesel, et~al.
\newblock Scaling rectified flow transformers for high-resolution image
  synthesis.
\newblock In \emph{Forty-first international conference on machine learning},
  2024.

\bibitem{fid2017}
Martin Heusel, Hubert Ramsauer, Thomas Unterthiner, Bernhard Nessler, and Sepp
  Hochreiter.
\newblock Gans trained by a two time-scale update rule converge to a local nash
  equilibrium.
\newblock \emph{Advances in neural information processing systems}, 30, 2017.

\bibitem{cfg2022}
Jonathan Ho and Tim Salimans.
\newblock Classifier-free diffusion guidance.
\newblock \emph{arXiv preprint arXiv:2207.12598}, 2022.

\bibitem{laytrol2026}
Sida Huang, Siqi Huang, Ping Luo, and Hongyuan Zhang.
\newblock Laytrol: Preserving pretrained knowledge in layout control for
  multimodal diffusion transformers.
\newblock In \emph{Proceedings of the AAAI Conference on Artificial
  Intelligence}, pages 5113--5121, 2026.

\bibitem{vae2014}
Diederik~P Kingma and Max Welling.
\newblock Auto-encoding variational bayes.
\newblock \emph{arXiv preprint arXiv:1312.6114}, 2013.

\bibitem{flux2klein2026}
Black~Forest Labs.
\newblock {FLUX.2: Frontier Visual Intelligence}.
\newblock \url{https://bfl.ai/blog/flux-2}, 2025.

\bibitem{gligen2023}
Yuheng Li, Haotian Liu, Qingyang Wu, Fangzhou Mu, Jianwei Yang, Jianfeng Gao,
  Chunyuan Li, and Yong~Jae Lee.
\newblock Gligen: Open-set grounded text-to-image generation.
\newblock In \emph{2023 IEEE/CVF Conference on Computer Vision and Pattern
  Recognition (CVPR)}, pages 22511--22521. IEEE, 2023.

\bibitem{retinanet2017}
Tsung-Yi Lin, Priya Goyal, Ross Girshick, Kaiming He, and Piotr Doll{\'a}r.
\newblock Focal loss for dense object detection.
\newblock In \emph{Proceedings of the IEEE international conference on computer
  vision}, pages 2980--2988, 2017.

\bibitem{flowmatching2023}
Yaron Lipman, Ricky~TQ Chen, Heli Ben-Hamu, Maximilian Nickel, and Matt Le.
\newblock Flow matching for generative modeling.
\newblock \emph{arXiv preprint arXiv:2210.02747}, 2022.

\bibitem{rectifiedflow2023}
Xingchao Liu, Chengyue Gong, and Qiang Liu.
\newblock Flow straight and fast: Learning to generate and transfer data with
  rectified flow.
\newblock \emph{arXiv preprint arXiv:2209.03003}, 2022.

\bibitem{adamw2019}
Ilya Loshchilov and Frank Hutter.
\newblock Decoupled weight decay regularization.
\newblock In \emph{International Conference on Learning Representations}, 2019.

\bibitem{lot2026}
Kiyohiro Nakayama, Brian Chao, Jan Ackermann, Hansheng Chen, Federico Tombari,
  Leonidas Guibas, Lior Yariv, and Gordon Wetzstein.
\newblock Level-of-token diffusion.
\newblock \emph{arXiv preprint arXiv:2610.05816}, 2026.

\bibitem{dit2023}
William Peebles and Saining Xie.
\newblock Scalable diffusion models with transformers.
\newblock In \emph{2023 IEEE/CVF International Conference on Computer Vision
  (ICCV)}, pages 4172--4182. IEEE, 2023.

\bibitem{ldm2022}
Robin Rombach, Andreas Blattmann, Dominik Lorenz, Patrick Esser, and Bj{\"o}rn
  Ommer.
\newblock High-resolution image synthesis with latent diffusion models.
\newblock In \emph{2022 IEEE/CVF conference on computer vision and pattern
  recognition (CVPR)}, pages 10674--10685. ieee, 2022.

\bibitem{terradit2026}
Srikumar Sastry, Dan Cher, Brian Wei, Aayush Dhakal, Subash Khanal, Dev Gupta,
  and Nathan Jacobs.
\newblock Terradit: Point-conditioned diffusion transformer for satellite image
  synthesis.
\newblock \emph{arXiv preprint arXiv:2603.02172}, 2026.

\bibitem{rope2024}
Jianlin Su, Murtadha Ahmed, Yu Lu, Shengfeng Pan, Wen Bo, and Yunfeng Liu.
\newblock Roformer: Enhanced transformer with rotary position embedding.
\newblock \emph{Neurocomputing}, 568:\penalty0 127063, 2024.

\bibitem{scenefid2021}
Tristan Sylvain, Pengchuan Zhang, Yoshua Bengio, R~Devon Hjelm, and Shikhar
  Sharma.
\newblock Object-centric image generation from layouts.
\newblock In \emph{Proceedings of the AAAI Conference on Artificial
  Intelligence}, pages 2647--2655, 2021.

\bibitem{fourier2020}
Matthew Tancik, Pratul Srinivasan, Ben Mildenhall, Sara Fridovich-Keil, Nithin
  Raghavan, Utkarsh Singhal, Ravi Ramamoorthi, Jonathan Barron, and Ren Ng.
\newblock Fourier features let networks learn high frequency functions in low
  dimensional domains.
\newblock \emph{Advances in neural information processing systems},
  33:\penalty0 7537--7547, 2020.

\bibitem{aerogen2025}
Datao Tang, Xiangyong Cao, Xuan Wu, Jialin Li, Jing Yao, Xueru Bai, Dongsheng
  Jiang, Yin Li, and Deyu Meng.
\newblock Aerogen: Enhancing remote sensing object detection with
  diffusion-driven data generation.
\newblock In \emph{2025 IEEE/CVF Conference on Computer Vision and Pattern
  Recognition (CVPR)}, pages 3614--3624. IEEE, 2025.

\bibitem{ficgen2025}
Wenzhuang Wang, Yifan Zhao, Mingcan Ma, Ming Liu, Zhonglin Jiang, Yong Chen,
  and Jia Li.
\newblock Ficgen: Frequency-inspired contextual disentanglement for
  layout-driven degraded image generation.
\newblock In \emph{2025 IEEE/CVF International Conference on Computer Vision
  (ICCV)}, pages 19097--19107. IEEE, 2025.

\bibitem{instancediffusion2024}
Xudong Wang, Trevor Darrell, Sai~Saketh Rambhatla, Rohit Girdhar, and Ishan
  Misra.
\newblock Instancediffusion: Instance-level control for image generation.
\newblock In \emph{2024 IEEE/CVF Conference on Computer Vision and Pattern
  Recognition (CVPR)}, pages 6232--6242. IEEE, 2024.

\bibitem{terraditomega2026}
Brian Wei, Srikumar Sastry, Daniel Cher, Eric Xing, and Nathan Jacobs.
\newblock Terradit-{$\Omega$}: Unified spatial control for satellite image
  synthesis with any geospatial primitive.
\newblock In \emph{European Conference on Computer Vision}, pages 383--401.
  Springer, 2026.

\bibitem{pma2025}
Haoyu Wu, Jingyi Xu, Qiaomu Miao, Dimitris Samaras, and Hieu Le.
\newblock Phase-aligned rope for mixed-resolution diffusion transformer.
\newblock In \emph{European Conference on Computer Vision}, pages 272--293.
  Springer, 2026.

\bibitem{dota2018}
Gui-Song Xia, Xiang Bai, Jian Ding, Zhen Zhu, Serge Belongie, Jiebo Luo, Mihai
  Datcu, Marcello Pelillo, and Liangpei Zhang.
\newblock Dota: A large-scale dataset for object detection in aerial images.
\newblock In \emph{2018 IEEE/CVF Conference on Computer Vision and Pattern
  Recognition}, pages 3974--3983. IEEE, 2018.

\bibitem{dcfl2023}
Chang Xu, Jian Ding, Jinwang Wang, Wen Yang, Huai Yu, Lei Yu, and Gui-Song Xia.
\newblock Dynamic coarse-to-fine learning for oriented tiny object detection.
\newblock In \emph{2023 IEEE/CVF Conference on Computer Vision and Pattern
  Recognition (CVPR)}, pages 7318--7328. IEEE, 2023.

\bibitem{aitodr2026}
Chang Xu, Ruixiang Zhang, Wen Yang, Haoran Zhu, Fang Xu, Jian Ding, and
  Gui-Song Xia.
\newblock Oriented tiny object detection: A dataset, benchmark, and dynamic
  unbiased learning.
\newblock \emph{IEEE transactions on pattern analysis and machine
  intelligence}, 2025.

\bibitem{qwen32025}
An Yang, Anfeng Li, Baosong Yang, Beichen Zhang, Binyuan Hui, Bo Zheng, Bowen
  Yu, Chang Gao, Chengen Huang, Chenxu Lv, et~al.
\newblock Qwen3 technical report.
\newblock \emph{arXiv preprint arXiv:2505.09388}, 2025.

\bibitem{h2rbox2023}
Xue Yang, Gefan Zhang, Wentong Li, Xuehui Wang, Yue Zhou, and Junchi Yan.
\newblock H2rbox: Horizontal box annotation is all you need for oriented object
  detection.
\newblock \emph{arXiv preprint arXiv:2210.06742}, 2022.

\bibitem{ofdiff2026}
Ziqi Ye, Shuran Ma, Jie Yang, Xiaoyi Yang, Yi Yang, Ziyang Gong, Xue Yang, and
  Haipeng Wang.
\newblock Object fidelity diffusion for remote sensing image generation.
\newblock In \emph{International Conference on Learning Representations}, pages
  3258--3278, 2026.

\bibitem{point2rbox2024}
Yi Yu, Xue Yang, Qingyun Li, Feipeng Da, Jifeng Dai, Yu Qiao, and Junchi Yan.
\newblock Point2rbox: Combine knowledge from synthetic visual patterns for
  end-to-end oriented object detection with single point supervision.
\newblock In \emph{2024 IEEE/CVF Conference on Computer Vision and Pattern
  Recognition (CVPR)}, pages 16783--16793. IEEE, 2024.

\bibitem{creatilayout2025}
Hui Zhang, Dexiang Hong, Yitong Wang, Jie Shao, Xinglong Wu, Zuxuan Wu, and
  Yu-Gang Jiang.
\newblock Creatilayout: Siamese multimodal diffusion transformer for creative
  layout-to-image generation.
\newblock In \emph{2025 IEEE/CVF International Conference on Computer Vision
  (ICCV)}, pages 18487--18497. IEEE, 2025{\natexlab{a}}.

\bibitem{controlnet2023}
Lvmin Zhang, Anyi Rao, and Maneesh Agrawala.
\newblock Adding conditional control to text-to-image diffusion models.
\newblock In \emph{2023 IEEE/CVF International Conference on Computer Vision
  (ICCV)}, pages 3813--3824. IEEE, 2023.

\bibitem{ccdiff2024}
Mu Zhang, Yunfan Liu, Yue Liu, Yuzhong Zhao, and Qixiang Ye.
\newblock Cc-diff: enhancing contextual coherence in remote sensing image
  synthesis.
\newblock \emph{arXiv preprint arXiv:2412.08464}, 2024.

\bibitem{ccdiffpp2025}
Mu Zhang, Yunfan Liu, Yue Liu, Yuzhong Zhao, and Qixiang Ye.
\newblock Cc-diff++: Spatially controllable text-to-image synthesis for remote
  sensing with enhanced contextual coherence.
\newblock \emph{IEEE Transactions on Geoscience and Remote Sensing},
  2025{\natexlab{b}}.

\bibitem{layout2im2019}
Bo Zhao, Lili Meng, Weidong Yin, and Leonid Sigal.
\newblock Image generation from layout.
\newblock In \emph{2019 IEEE/CVF Conference on Computer Vision and Pattern
  Recognition (CVPR)}, pages 8576--8585. IEEE, 2019.

\bibitem{layoutdiffusion2023}
Guangcong Zheng, Xianpan Zhou, Xuewei Li, Zhongang Qi, Ying Shan, and Xi Li.
\newblock Layoutdiffusion: Controllable diffusion model for layout-to-image
  generation.
\newblock In \emph{2023 IEEE/CVF Conference on Computer Vision and Pattern
  Recognition (CVPR)}, pages 22490--22499. IEEE, 2023.

\bibitem{migc2024}
Dewei Zhou, You Li, Fan Ma, Xiaoting Zhang, and Yi Yang.
\newblock Migc: Multi-instance generation controller for text-to-image
  synthesis.
\newblock In \emph{2024 IEEE/CVF Conference on Computer Vision and Pattern
  Recognition (CVPR)}, pages 6818--6828. IEEE, 2024.

\bibitem{mmrotate2022}
Yue Zhou, Xue Yang, Gefan Zhang, Jiabao Wang, Yanyi Liu, Liping Hou, Xue Jiang,
  Xingzhao Liu, Junchi Yan, Chengqi Lyu, et~al.
\newblock Mmrotate: A rotated object detection benchmark using pytorch.
\newblock In \emph{Proceedings of the 30th ACM international conference on
  multimedia}, pages 7331--7334, 2022.

\bibitem{cheng2022diorr}
Gong Cheng, Jiabao Wang, Ke Li, Xingxing Xie, Chunbo Lang, Yanqing Yao, and
  Junwei Han.
\newblock Anchor-free oriented proposal generator for object detection.
\newblock \emph{IEEE Transactions on Geoscience and Remote Sensing},
  60:\penalty0 1--11, 2022.

\bibitem{duchon1979lanczos}
Claude~E Duchon.
\newblock Lanczos filtering in one and two dimensions.
\newblock \emph{Journal of Applied Meteorology (1962-1982)}, pages 1016--1022,
  1979.

\bibitem{everingham2010pascal}
Mark Everingham, Luc Van~Gool, Christopher~KI Williams, John Winn, and Andrew
  Zisserman.
\newblock The pascal visual object classes (voc) challenge.
\newblock \emph{International journal of computer vision}, 88\penalty0
  (2):\penalty0 303--338, 2010.

\bibitem{he2016deep}
Kaiming He, Xiangyu Zhang, Shaoqing Ren, and Jian Sun.
\newblock Deep residual learning for image recognition.
\newblock In \emph{Proceedings of the IEEE conference on computer vision and
  pattern recognition}, pages 770--778, 2016.

\bibitem{airbus2018}
inversion, Jeff Faudi, and Martin.
\newblock Airbus ship detection challenge.
\newblock \url{https://www.kaggle.com/competitions/airbus-ship-detection},
  2018.
\newblock Kaggle.

\bibitem{lam2018xview}
Darius Lam, Richard Kuzma, Kevin McGee, Samuel Dooley, Michael Laielli, Matthew
  Klaric, Yaroslav Bulatov, and Brendan McCord.
\newblock xview: Objects in context in overhead imagery.
\newblock \emph{arXiv preprint arXiv:1802.07856}, 2018.

\bibitem{dior2020}
Ke Li, Gang Wan, Gong Cheng, Liqiu Meng, and Junwei Han.
\newblock Object detection in optical remote sensing images: A survey and a new
  benchmark.
\newblock \emph{ISPRS journal of photogrammetry and remote sensing},
  159:\penalty0 296--307, 2020.

\bibitem{lin2014microsoft}
Tsung-Yi Lin, Michael Maire, Serge Belongie, James Hays, Pietro Perona, Deva
  Ramanan, Piotr Doll{\'a}r, and C~Lawrence Zitnick.
\newblock Microsoft coco: Common objects in context.
\newblock In \emph{European conference on computer vision}, pages 740--755.
  Springer, 2014.

\bibitem{liu2025text2earth}
Chenyang Liu, Keyan Chen, Rui Zhao, Zhengxia Zou, and Zhenwei Shi.
\newblock Text2earth: Unlocking text-driven remote sensing image generation
  with a global-scale dataset and a foundation model.
\newblock \emph{IEEE Geoscience and Remote Sensing Magazine}, 13\penalty0
  (3):\penalty0 238--259, 2025.

\bibitem{liu2021swin}
Ze Liu, Yutong Lin, Yue Cao, Han Hu, Yixuan Wei, Zheng Zhang, Stephen Lin, and
  Baining Guo.
\newblock Swin transformer: Hierarchical vision transformer using shifted
  windows.
\newblock In \emph{2021 IEEE/CVF international conference on computer vision
  (ICCV)}, pages 9992--10002. Ieee, 2021.

\bibitem{loshchilov2017sgdr}
Ilya Loshchilov and Frank Hutter.
\newblock {SGDR}: Stochastic gradient descent with warm restarts.
\newblock In \emph{International Conference on Learning Representations}, 2017.

\bibitem{progressive_distillation2022}
Tim Salimans and Jonathan Ho.
\newblock Progressive distillation for fast sampling of diffusion models.
\newblock \emph{arXiv preprint arXiv:2202.00512}, 2022.

\bibitem{wang2021aitod}
Jinwang Wang, Wen Yang, Haowen Guo, Ruixiang Zhang, and Gui-Song Xia.
\newblock Tiny object detection in aerial images.
\newblock In \emph{2020 25th international conference on pattern recognition
  (ICPR)}, pages 3791--3798. IEEE, 2021.

\bibitem{xie2021oriented}
Xingxing Xie, Gong Cheng, Jiabao Wang, Xiwen Yao, and Junwei Han.
\newblock Oriented r-cnn for object detection.
\newblock In \emph{2021 IEEE/CVF International Conference on Computer Vision
  (ICCV)}, pages 3500--3509. IEEE, 2021.

\bibitem{dmd2024}
Tianwei Yin, Micha{\"e}l Gharbi, Richard Zhang, Eli Shechtman, Fredo Durand,
  William~T Freeman, and Taesung Park.
\newblock One-step diffusion with distribution matching distillation.
\newblock In \emph{2024 IEEE/CVF Conference on Computer Vision and Pattern
  Recognition (CVPR)}, pages 6613--6623. IEEE, 2024.

\bibitem{zhu2022visdrone}
Pengfei Zhu, Longyin Wen, Dawei Du, Xiao Bian, Heng Fan, Qinghua Hu, and Haibin
  Ling.
\newblock Detection and tracking meet drones challenge.
\newblock \emph{IEEE Transactions on Pattern Analysis and Machine
  Intelligence}, 44\penalty0 (11):\penalty0 7380--7399, 2022.

\end{thebibliography}
}

\end{document}